\pdfoutput=1

\documentclass[11pt]{article}

\usepackage{times}
\usepackage{latexsym}
\usepackage{graphicx}
\usepackage{sidecap}
\usepackage[small]{caption}
\usepackage{subcaption}
\usepackage{amsmath}
\usepackage{amsfonts}     
\allowdisplaybreaks
\usepackage{amsthm}

\usepackage{thmtools}
\usepackage{thm-restate}

\usepackage{multicol}
\newcommand{\NC}{{NC}} 
\usepackage{pifont}
\usepackage{xspace}
\newcommand{\circone}{\ding{172}\xspace}
\newcommand{\circtwo}{\ding{173}\xspace}

\DeclareMathOperator{\argmin}{argmin}
\DeclareMathOperator{\argmax}{argmax}

\usepackage[round]{natbib}

\usepackage{appendix}

\newcommand{\rightcomment}[1]{\(\triangleright\) {\small \it #1}}
\newcommand{\eqcomment}[1]{\addtocounter{equation}{1}\tag*{\rightcomment{#1}\quad(\theequation)}}
\usepackage{suffix}
\WithSuffix\newcommand\eqcomment*[1]{\tag*{\rightcomment{#1}}}

\usepackage[hidelinks]{hyperref}

\usepackage[]{EMNLP2022}%

\usepackage[]{todonotes}

\newcommand{\cutforspace}[1]{}

\usepackage[T1]{fontenc}
\usepackage[utf8]{inputenc}
\usepackage{url}            %
\usepackage{booktabs}       %
\usepackage{amsfonts}       %
\usepackage{nicefrac}       %
\usepackage{microtype}      %
\usepackage{xcolor}         %

\usepackage{microtype}

\usepackage{inconsolata}

\usepackage{algorithm}
\usepackage[noend]{algpseudocode}

\renewcommand\algorithmicthen{:}
\algnewcommand{\IfThen}[2]{\State \algorithmicif\ #1\ \algorithmicthen\ #2}
\algnewcommand{\IfThenElse}[3]{\State \algorithmicif\ #1\ \algorithmicthen\ #2\ \algorithmicelse\ #3}
\algrenewcommand{\algorithmiccomment}[1]{\hfill \rightcomment{#1}}
\algnewcommand{\LineComment}[1]{\State \rightcomment{#1}}
\algnewcommand{\LinesComment}[1]{\State \rightcomment{\parbox[t]{\linewidth-\leftmargin-\widthof{\(\triangleright\) }}{#1}}\smallskip}
\algnewcommand\algorithmicinput{{\bfseries Input:}}
\algnewcommand\INPUT{\item[\algorithmicinput]}
\algnewcommand\algorithmicoutput{{\bfseries Output:}}
\algnewcommand\OUTPUT{\item[\algorithmicoutput]}
\makeatletter
\newcounter{algorithmicH}
\let\oldalgorithmic\algorithmic
\renewcommand{\algorithmic}{%
  \stepcounter{algorithmicH}
  \oldalgorithmic}
\renewcommand{\theHALG@line}{ALG@line.\thealgorithmicH.\arabic{ALG@line}}
\makeatother
\makeatletter
\newcommand{\algmargin}{\the\ALG@thistlm}
\makeatother
\algnewcommand{\Statepar}[1]{\State\parbox[t]{\dimexpr\linewidth-\algmargin}{\strut #1\strut}}

\usepackage{xpatch}
\usepackage[compact]{titlesec}
\titlespacing{\section}{0pt}{1ex}{0.5ex}
\titlespacing{\subsection}{0pt}{0.5ex}{0ex}
\titlespacing{\subsubsection}{0pt}{0.5ex}{0ex} 
\titlespacing{\paragraph}{0pt}{0ex}{1ex} 
\usepackage{setspace}
\AtBeginDocument{%
  \addtolength\abovedisplayskip{-0.25\baselineskip}%
  \addtolength\belowdisplayskip{-0.25\baselineskip}%
  \addtolength\abovedisplayshortskip{-0.25\baselineskip}%
  \addtolength\belowdisplayshortskip{-0.25\baselineskip}%
}

\usepackage{amsmath}
\usepackage{amssymb}
\usepackage{mathrsfs}
\usepackage{mathtools, cuted}
\usepackage{tabularx, booktabs}
\newcolumntype{C}{>{\centering\arraybackslash}X}
\newcolumntype{R}{>{\raggedleft\arraybackslash}X}
\newcolumntype{S}{>{\raggedleft\arraybackslash\hsize=.5\hsize}X}
\usepackage{latexsym}
\usepackage{url}
\usepackage{xspace}
\usepackage{bm,array}
\usepackage{amsfonts}
\usepackage{enumitem}
\usepackage{cases}
\usepackage{mathtools}
\usepackage{empheq}
\usepackage{bm}
\usepackage{esvect}
\usepackage[noabbrev,capitalize]{cleveref}
\crefname{equation}{equation}{equations}
\crefname{section}{section}{sections}
\crefname{footnote}{footnote}{footnotes}   
\crefname{line}{line}{lines}   
\crefname{assumption}{assumption}{assumptions}

\usepackage{bbm}

\let\frac=\tfrac  %

\usepackage[compact]{titlesec}  %

\renewcommand{\vec}[1]{{\boldsymbol{\mathbf{#1}}}}   %

\newcommand{\defeq}{\mathrel{\stackrel{\textnormal{\tiny def}}{=}}}

\renewcommand{\th}{\textsuperscript{th}\xspace}

\usepackage{tikz}
\usetikzlibrary{shapes.geometric}

\usetikzlibrary{arrows,decorations.markings}
\usetikzlibrary{arrows}

\usepackage{verbatim}

\title{Hidden State Variability of Pretrained Language Models\\Can Guide Computation Reduction for Transfer Learning}

\author{
 Shuo Xie\thanks{\ \ Work done during internship at TTI-Chicago.}$\,\,^{1,2}$\quad Jiahao Qiu$^{3}$\quad Ankita Pasad$^{2}$\quad Li Du$^{4}$\quad Qing Qu$^{3}$\quad Hongyuan Mei$^{2}$ \\
$^1$University of Chicago\quad
$^2$Toyota Technological Institute at Chicago\\
$^3$University of Michigan\quad 
$^4$Johns Hopkins University\\
\texttt{shuox@uchicago.edu,hongyuan@ttic.edu}
}

\begin{document}
\maketitle
\begin{abstract}
While transferring a pretrained language model, common approaches conventionally attach their task-specific classifiers to the top layer and adapt all the pretrained layers.
We investigate whether one could make a task-specific selection on which subset of the layers to adapt and where to place the classifier.%
The goal is to reduce the computation cost of transfer learning methods (e.g. fine-tuning or adapter-tuning) without sacrificing its performance.%

We propose to select layers based on the variability of their hidden states given a task-specific corpus. 
We say a layer is already ``well-specialized'' in a task if the within-class variability of its hidden states is low relative to the between-class variability.
Our variability metric is cheap to compute and doesn't need any training or hyperparameter tuning. It is robust to data imbalance and data scarcity.
Extensive experiments on the GLUE benchmark demonstrate that selecting layers based on our metric can yield significantly stronger performance than using the same number of top layers and often match the performance of fine-tuning or adapter-tuning the entire language model.%

\end{abstract}

\section{Introduction}\label{sec:intro}\label{sec:problem}

Transfer learning from a pretrained language model (PLM) is now the de-facto paradigm in natural language processing (NLP). 
The conventional approaches of leveraging PLMs include fine-tuning all the parameters in the language model (LM) and some lightweight alternatives that can decrease the number of tuning parameters such as adapter-tuning~\citep{houlsby2019parameter,hu2021lora,he2021towards} and prefix-tuning~\citep{li2021prefix}. 
These methods have one thing in common: they all involve the entire PLM and attach a classifier to its top layer. 
However, PLMs were optimized via the language modeling objective and thus their top layers have been \emph{specialized} in producing representations which facilitate optimizing that objective. 
Such mismatch between the pretraining and fine-tuning objectives poses the following questions: 
\begin{description}[leftmargin=*]
    \item \circone Given a pretrained language model and a downstream task, can we measure how ``\emph{well-specialized}'' each layer has already been in that task, without any task-specific tuning?
    \item \circtwo If the answer to \circone is yes, can we use the layer-wise ``\emph{task-specialty}'' as a guide in improving the computation efficiency of the transfer learning methods such as fine-tuning and adapter-tuning?
\end{description}

In this paper, we take a {technically principled} approach to investigate the research questions \circone and \circtwo. 
First, we define a metric in \cref{sec:var} to measure the ``task-specialty'' of each layer in a given PLM. 
Our task-speciality score is inspired by the neural collapse (\NC) phenomenon which has been widely observed in the computer vision community~\citep{papyan2020prevalence}: as training converges, the top-layer representations of the images with the same label form an extremely tight cluster.
In our setting, we examine the variability of the representations of the linguistic sequences given by each layer of the PLM, and define our layer-wise task-specialty to be the within-class variability normalized by the between-class variability. 
Computing our metric does not require any training or hyperparameter tuning.
Experiments on the GLUE benchmark demonstrate that it is highly correlated with layer-wise probing performance, thus giving a clear ``yes'' to the question \circone above. 

We propose several layer-selecting strategies in \cref{sec:strategy} based on our proposed task-specialty metric.
Our strategies are complementary to all the major paradigms of transfer learning (such as fine-tuning and adapter-tuning) and thus can take advantages of the state-of-the-art at the time: only the selected layers will be tuned (e.g., via fine-tuning or using adapters) such that the computation cost of the tuning methods can be further reduced. 
Experiments on the GLUE benchmark demonstrate that our proposed strategies are highly effective: under comparable computation budget, fine-tuning or adapter-tuning the layers selected by our strategies can achieve significantly higher performance than using the layers selected by the widely adopted baseline strategies; it can even often match the performance of fine-tuning or adapter-tuning the entire PLM which takes 500\% more computation cost.%

Through extensive ablation studies, we demonstrate the comparable advantages of our proposed task-specialty metric over potential alternatives (such as CCA and mutual information) as well as its robustness to data scarcity and data imbalance. 

\section{Technical Background}\label{sec:tech}%
In this paper, we focus on classification tasks. %
Technically, each classification task has a corpus of training data $\{(\vec{x}_n,y_n)\}_{n=1}^{N}$ where each $\vec{x}_n=(x_{n,1}, \ldots, x_{n,T})$ is a sequence of linguistic tokens and each $y_n \in \mathcal{Y}$ is a discrete class label. Such tasks include
\begin{itemize}[leftmargin=*,noitemsep,topsep=0pt]
    \item {\em Sentiment analysis.} Each $\vec{x}$ is a single sequence of words such as ``This movie is fantastic'' and $y$ is a sentiment label from $\{ \text{positive}, \text{negative} \}$. Thus, the sentiment analysis can be cast as a binary-class classification problem.
    \item {\em Natural language inference.} Each $\vec{x}$ is of the form ``{premise} [SEP] {hypothesis}'' such as ``Fun for adults and children. [SEP] Fun for only children.'' where ``[SEP]'' is a special separator token. The label $y \in \{\text{yes}, \text{neutral}, \text{no}\}$ indicates whether the premise entails the hypothesis. It is a three-class classification problem.
\end{itemize}
A PLM performs a classification task as follows: 
\begin{enumerate}[leftmargin=*,noitemsep,topsep=0pt]
    \item It reads each given sequence $\vec{x}_n$ and embed it into a series of hidden state vectors
    \begin{align*}
    &\text{layer }L\quad
    &&\vec{h}_{n,0}^{(L)}\quad \vec{h}_{n,1}^{(L)}\, \ldots\, \vec{h}_{n,t}^{(L)}\, \ldots\, \vec{h}_{n,T}^{(L)}\\
    &&\ldots\\
    &\text{layer }\ell\quad
    &&\vec{h}_{n,0}^{(\ell)}\quad \vec{h}_{n,1}^{(\ell)}\, \ldots\, \vec{h}_{n,t}^{(\ell)}\, \ldots\, \vec{h}_{n,T}^{(\ell)}\\
    &&\ldots\\
    &\text{layer }1\quad
    &&\vec{h}_{n,0}^{(1)}\quad \vec{h}_{n,1}^{(1)}\, \ldots\, \vec{h}_{n,t}^{(1)}\, \ldots\, \vec{h}_{n,T}^{(1)}
    \vspace{-10pt}
    \end{align*}
    where $\vec{h}_{n,t}^{(\ell)}$ denotes the hidden state of token $\vec{x}_{n,t}$ given by layer $\ell$ and $x_{n,0}=\text{CLS}$ is a special classification (CLS) token.%
    \item 
    The top-layer hidden state $\vec{h}_{n,0}^{(L)}$ of the CLS token is read by a neural network $f$ followed by a softmax layer, which gives the probability distribution over the target label $y\in\mathcal{Y}$:%
    \begin{align}
        p(y \mid \vec{x}_n) = \textrm{softmax}_{y}(f(\vec{h}_{n,0}^{(L)})) \label{eq:loglike}
    \end{align}
    The net $f$ is also called ``classification head''.
\end{enumerate}

Transfer learning is to maximize the log probability of the ground-truth label $y_n$---i.e., \mbox{$\log p(y_n \mid \vec{x}_n)$}---by learning the parameters of the classification head $f$ as well as certain \emph{method-specific} parameters: 
\begin{itemize}[leftmargin=*,noitemsep,topsep=0pt]%
    \item \textit{Fine-tuning} updates all the PLM parameters~\citep{peters-18-elmo,devlin-18-bert}.
    \item \textit{Adapter-tuning} inserts adapters (i.e., small neural networks) into the LM layers and updates the new adapter parameters~\citep{houlsby2019parameter,hu2021lora,he2021towards}. %
    \item \textit{Prefix-tuning} augments trainable tokens to the input $\vec{x}$ and tunes the new token embeddings~\citep{li2021prefix,qin-eisner-2021-learning,hambardzumyan-etal-2021-warp}. 
\end{itemize}

\section{The Method}\label{sec:method}
Our goal is to answer the research questions \circone and \circtwo introduced in \cref{sec:problem}. 
That involves finding a layer-specific metric $\nu^{(1)}, \ldots, \nu^{(\ell)}, \ldots, \nu^{(L)}$ where each $\nu^{(\ell)}$ measures the task-specialty of layer $\ell$. 
Suppose that we use $s^{(\ell)}$ to denote the task score that we can achieve by letting the classification head read the layer $\ell$ hidden state $\vec{h}_{n,0}^{(\ell)}$ of the CLS token. 
If $\nu^{(\ell)}$ is highly (positively or negatively) correlated with $s^{(\ell)}$, then the answer to question \circone is yes. 
To answer question \circtwo involves designing $\nu$-based strategies that select a subset of layers to use in transfer learning approaches. 

In this section, we introduce our task-specialty metric $\nu^{(\ell)}$ (\cref{sec:var}) along with a few strategies for selecting layers (\cref{sec:strategy}). 
In \cref{sec:exp}, we will empirically demonstrate the effectiveness of our proposed metric and strategies. 
\begin{figure}[t]
	\begin{center}
	    \begin{subfigure}[t]{0.48\linewidth}
			\includegraphics[width=0.95\linewidth]{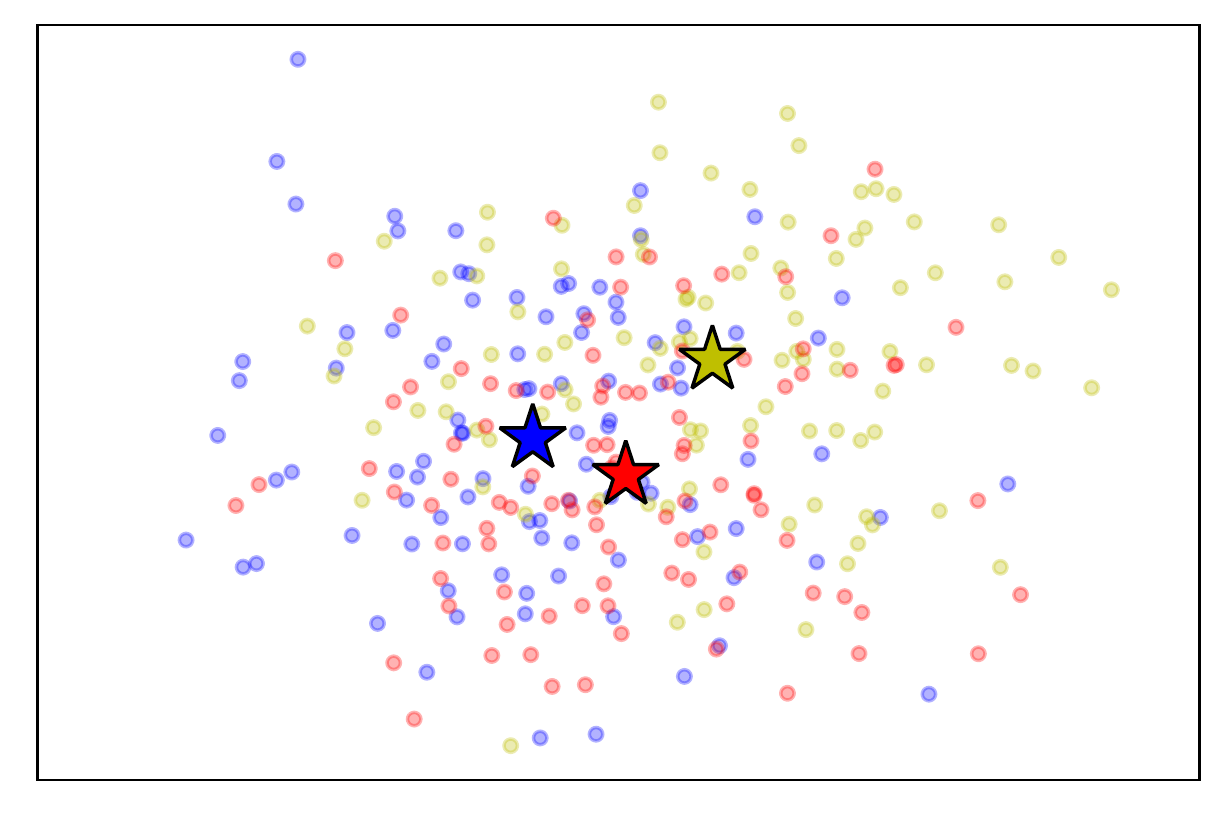}
			\vspace{-4pt}
			\caption{High within-class variability and low between-class variability.}\label{fig:nc_bad}
		\end{subfigure}
		\hfill
		\begin{subfigure}[t]{0.48\linewidth}
			\includegraphics[width=0.95\linewidth]{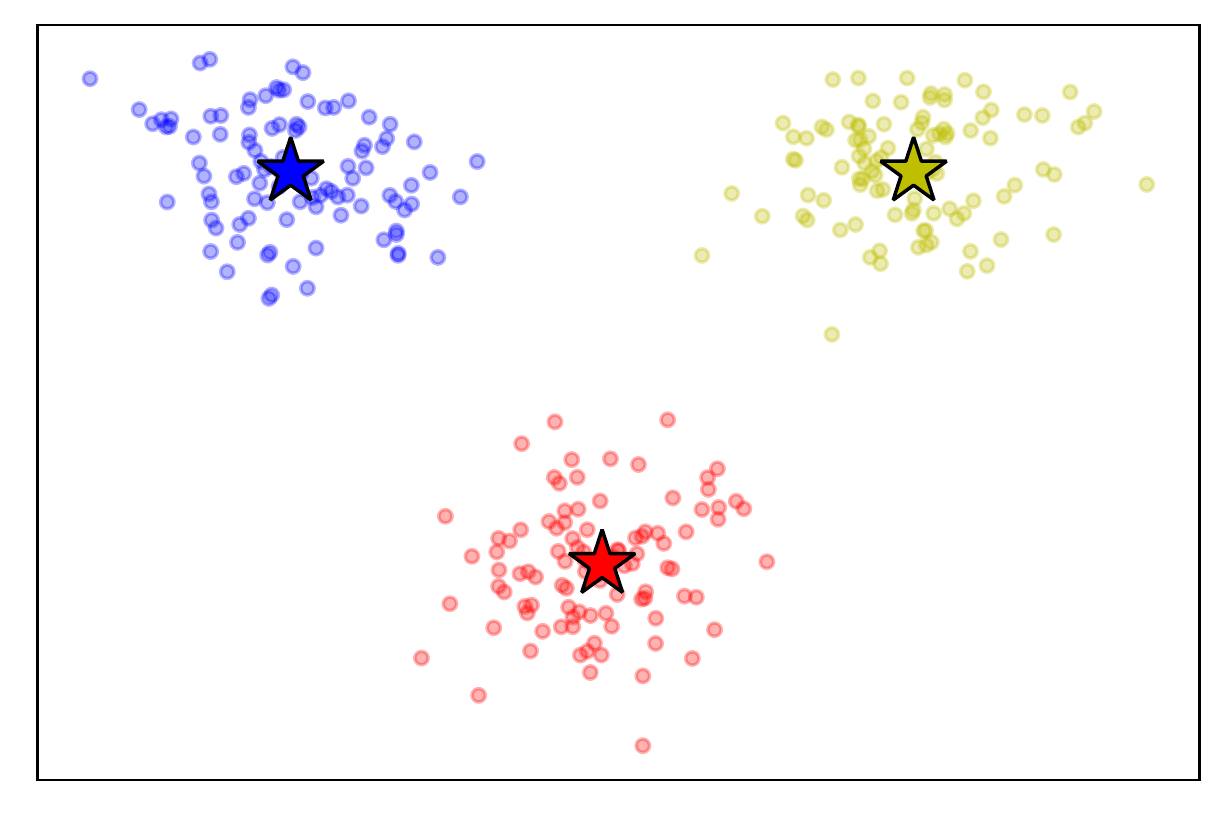}
			\vspace{-4pt}
			\caption{Low within-class variability and high between-class variability.}\label{fig:nc_good}
		\end{subfigure}
	\vspace{-8pt}
	\caption{An illustration of our variability-based task-specialty metric with hypothetical data. Each dot denotes a two-dimensional hidden state vector and its color denotes its target label. Each colored star denotes the mean vector of its class.}\label{fig:nc}
	\end{center}
	\vspace{-10pt}
\end{figure}

\subsection{Hidden State Variability Ratio}\label{sec:var}
For a given task, we define our task-specialty metric $\nu^{(1)},\ldots,\nu^{(L)}$ based on the \emph{variability} of the hidden state vectors 
that the PLM produces by embedding the training input sequences $\{ \vec{x}_n \}_{n=1}^{N}$.
We use hypothetical data to illustrate our intuition in \cref{fig:nc}: after grouped based on the target labels $y_n$, the variability of the hidden states within the same group
(dots of same color) measures the difficulty of separating them, while the variability of the mean states (stars of different colors) quantifies how easy it is to tell the different groups apart.%

Technically, for each layer $\ell$, we first define the sequence-level hidden state $\vec{h}_{n}^{(\ell)}$ for each input $\vec{x}_n$ to be the average of the hidden states of all the (non-CLS) tokens $\vec{h}_{n}^{(\ell)} \defeq \frac{1}{T} \sum_{t=1}^{T}\vec{h}_{n,t}^{(\ell)}$. %
These sequence-level states correspond to the dots in \cref{fig:nc}. 
Then we group all the $\vec{h}_{n}^{(\ell)}$ based on the target labels $y_n$: $\mathcal{G}_{y}^{(\ell)} \defeq \{ \vec{h}_{n}^{(\ell)} : y_n = y \}$. The mean vector of each group is defined as $\bar{\vec{h}}_{y}^{(\ell)} \defeq \frac{1}{|\mathcal{G}_{y}^{(\ell)}|} \sum_{\vec{h} \in \mathcal{G}_{y}^{(\ell)}} \vec{h}$ and they correspond to the stars in \cref{fig:nc}. %
The mean vector between classes is defined as $\bar{\vec{h}}^{(\ell)} \defeq \frac{1}{|\mathcal{Y}|} \sum_{y\in\mathcal{Y}} \bar{\vec{h}}_{y}^{(\ell)}$.
Then the within-group variability $\vec{\Sigma}_{\text{w}}^{(\ell)}$ and between-group variability $\vec{\Sigma}_{\text{b}}^{(\ell)}$ are defined using the sequence-level states and mean states: 
\begin{subequations}
\begin{align*}
    \vec{\Sigma}_{\text{w}}^{(\ell)}
    &\defeq 
    \frac{1}{|\mathcal{Y}|} \sum_{y\in\mathcal{Y}} \frac{1}{|\mathcal{G}_{y}^{(\ell)}|} \sum_{\vec{h}\in\mathcal{G}_{y}^{(\ell)}} (\vec{h} -
    \bar{\vec{h}}_{y}^{(\ell)}) (\vec{h} - 
    \bar{\vec{h}}_{y}^{(\ell)})^{\top} \\
    \vec{\Sigma}_{\text{b}}^{(\ell)}
    &\defeq 
    \frac{1}{|\mathcal{Y}|} \sum_{y\in\mathcal{Y}} (\bar{\vec{h}}_{y}^{(\ell)} - \bar{\vec{h}}^{(\ell)}) (\bar{\vec{h}}_{y}^{(\ell)} - \bar{\vec{h}}^{(\ell)})^{\top} 
\end{align*}
\end{subequations}
Both $\vec{\Sigma}_{\text{w}}^{(\ell)}$ and $\vec{\Sigma}_{\text{b}}^{(\ell)}$ are a lot like covariance matrices since they measure the deviation from the mean vectors. %
Finally, we define our task-specialty metric to be the within-group variability $\vec{\Sigma}_{\text{w}}^{(\ell)}$ scaled and rotated by the pseudo-inverse of between-class variability $\vec{\Sigma}_{\text{b}}^{(\ell)}$
\begin{align}
    \nu^{(\ell)} \defeq \frac{1}{|\mathcal{Y}|} \mathrm{trace}\left(\vec{\Sigma}_{\text{w}}^{(\ell)}\vec{\Sigma}_{\text{b}}^{^{(\ell)}\dagger}\right) \label{eq:ratio}
\end{align}
The pseudo-inverse in \cref{eq:ratio} is why we use the average state as our sequence-level representation: averaging reduces the noise in the state vectors and thus leads to stable computation of $\nu^{(\ell)}$.%

We believe that the layers with small $\nu^{(\ell)}$ are likely to do better than those with large $\nu^{(\ell)}$ when transferred to the downstream task. 
Our belief stems from two key insights. 

\paragraph{Remark-I: neural collapse.}
Our proposed metric is mainly inspired by the neural collapse (NC) phenomenon: when training a deep neural model in classifying images, one can see that the top-layer representations of the images with the same label form an extremely tight cluster as training converges. 
Extensive theoretical and empirical studies show that a lower within-class variability can indicate a better generalization~\citep{papyan2020prevalence,hui2022limitations,galanti2022on}.
Thus we examine the variability of the layer-wise representations of the linguistic sequences and hope that it can measure the task-specialty of each layer of the given PLM. 
Our metric is slightly different from the widely accepted neural collapse metric; please see \cref{app:metric_diff} for a detailed discussion.

\paragraph{Remark-II: signal-to-noise ratio.}
In multivariate statistics~\citep{anderson1973asymptotically}, $\mathrm{trace}\left(\vec{\Sigma}_{\text{w}} \vec{\Sigma}_{\text{b}}^{\dagger}\right)$ is able to measure the {inverse} signal-to-noise ratio for classification problems and thus a lower value indicates a lower chance of misclassification. 
Intuitively, the between-class variability $\vec{\Sigma}_{\text{b}}$ is the signal which one can use to tell different clusters apart while the within-class variability $\vec{\Sigma}_{\text{w}}$ is the noise that makes the clusters overlapped and thus the separation difficult; see \cref{fig:nc} for examples. 

\paragraph{Remark-III: linear discriminant analysis.}
A low $\nu$ implies that it is easy to correctly classify the data with linear discriminant analysis (LDA)~\citep{hastie2009elements}. 
Technically, LDA assumes that the data of each class is Gaussian-distributed and it classifies a new data point $\vec{h}$ by checking how close it is to each mean vector $\bar{\vec{h}}_{y}$ scaled by the covariance matrix $\vec{\Sigma}$ which is typically shared across classes. 
Though our metric does not make the Gaussian assumption, a low $\nu$ suggests that the class means $\bar{\vec{h}}_{y}$ are far from each other relative to the within-class variations $\vec{\Sigma}_{\text{w}}$, meaning that the decision boundary of LDA would tend to be sharp. 
Actually, our $\vec{\Sigma}_{\text{w}}$ is an estimate to the Gaussian covariance matrix $\vec{\Sigma}$ of LDA. 

\subsection{Layer-Selecting Strategies}\label{sec:strategy}
\begin{figure*}[t]
	\begin{center}
    	\begin{subfigure}[t]{0.23\linewidth}
    			\includegraphics[width=0.95\linewidth]{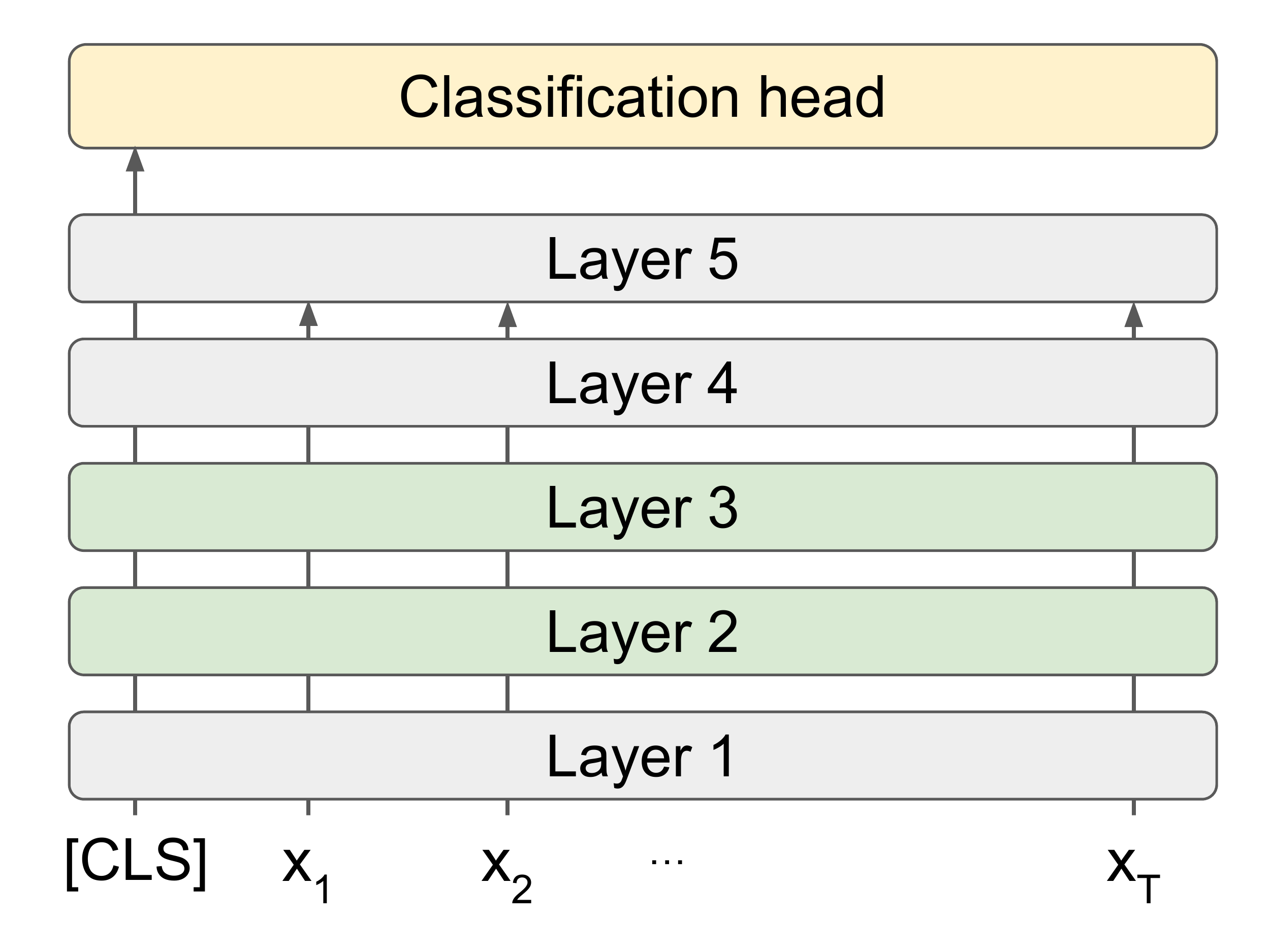}
    			\vspace{-5pt}
    			\caption{$(2,3,5)$}
    			\label{fig:partial-1}
        \end{subfigure}
        ~
	    \begin{subfigure}[t]{0.23\linewidth}
			\includegraphics[width=0.95\linewidth]{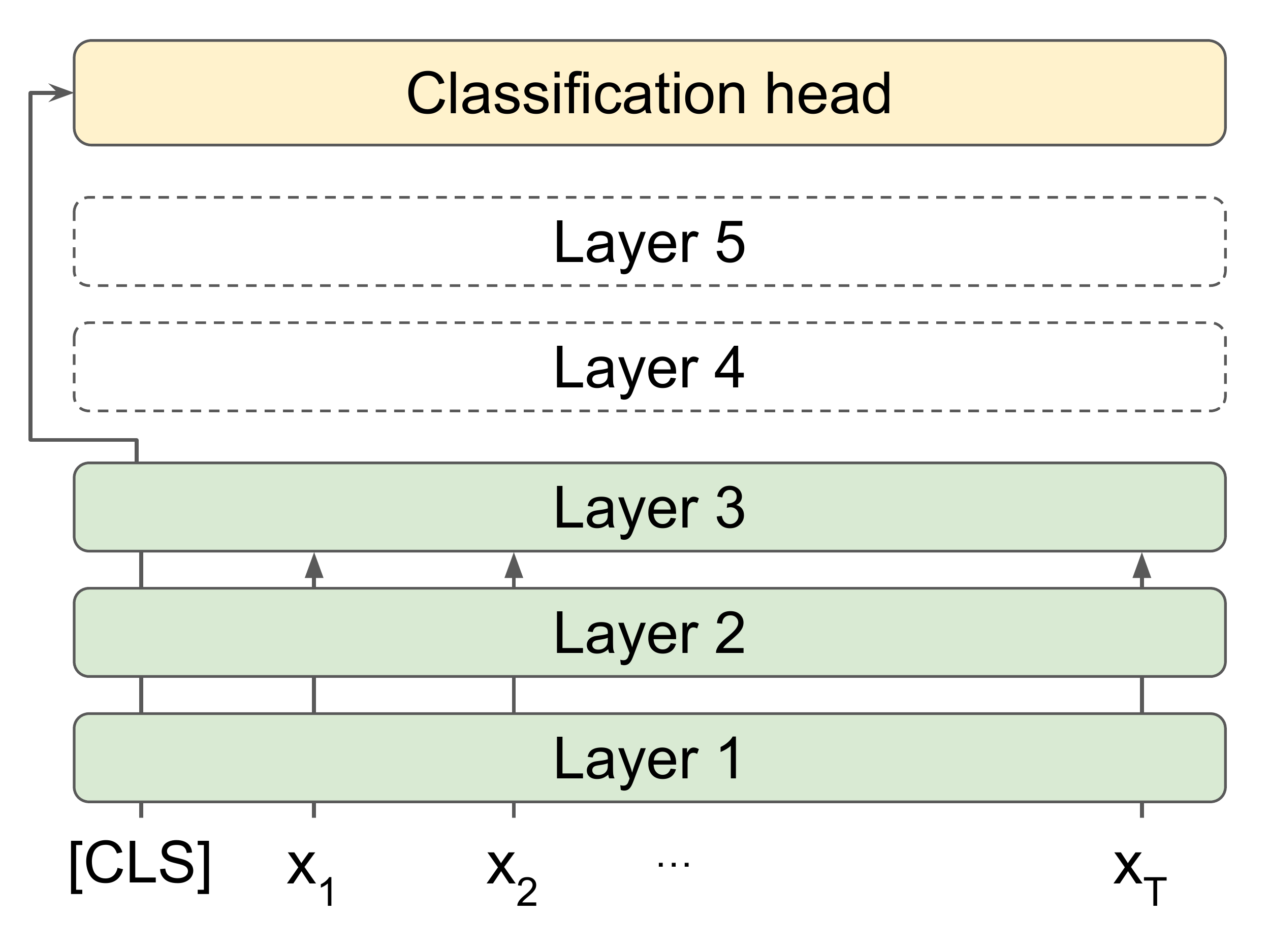}
			\vspace{-5pt}
 			\caption{$(1,3,3)$}
 			\label{fig:partial-2}
		\end{subfigure}
		~
		\begin{subfigure}[t]{0.23\linewidth}
			\includegraphics[width=0.95\linewidth]{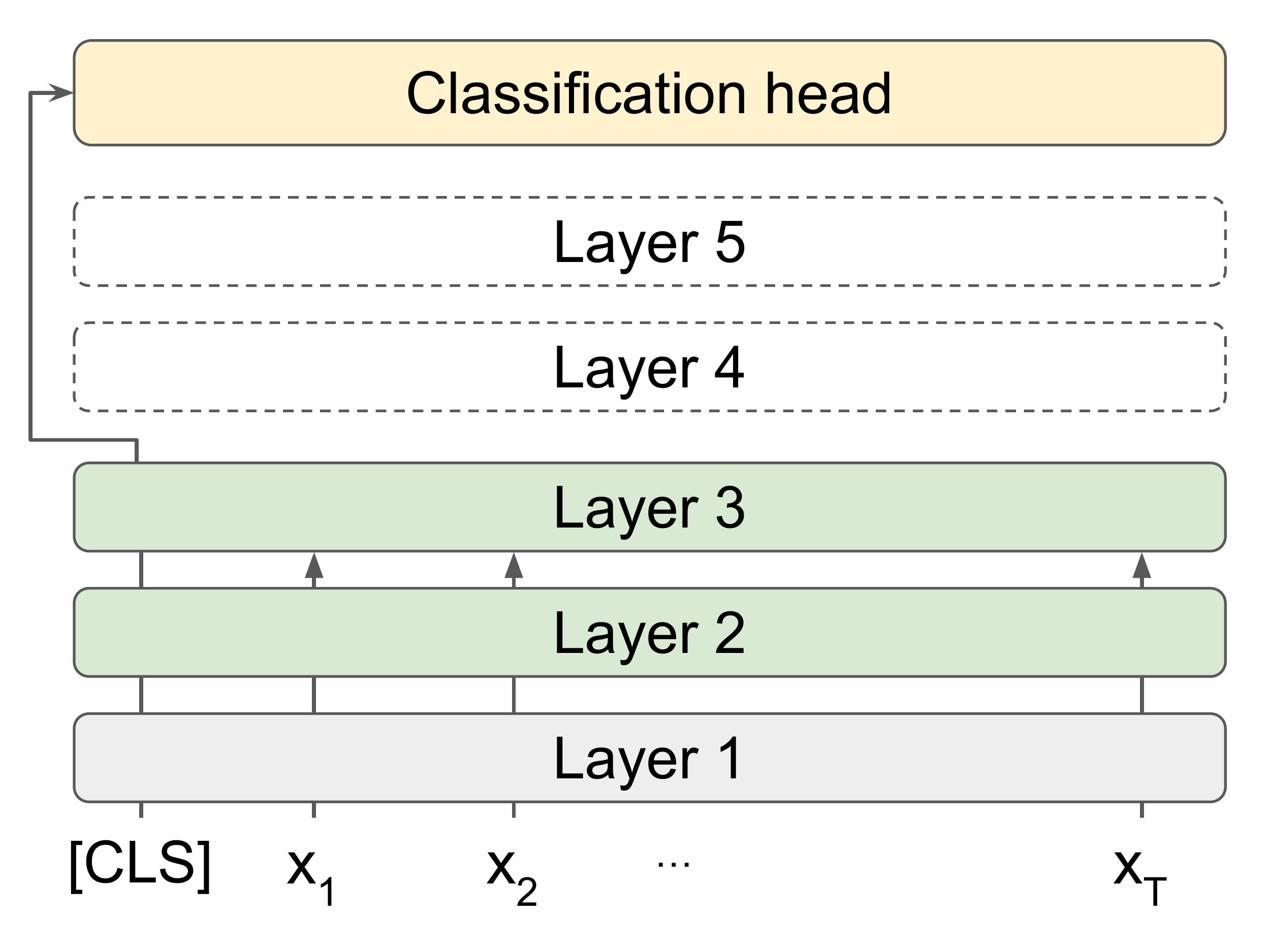}
			\vspace{-5pt}
			\caption{$(2,3,3)$}
			\label{fig:partial-3}
		\end{subfigure}
		~
		\begin{subfigure}[t]{0.23\linewidth}
			\includegraphics[width=0.95\linewidth]{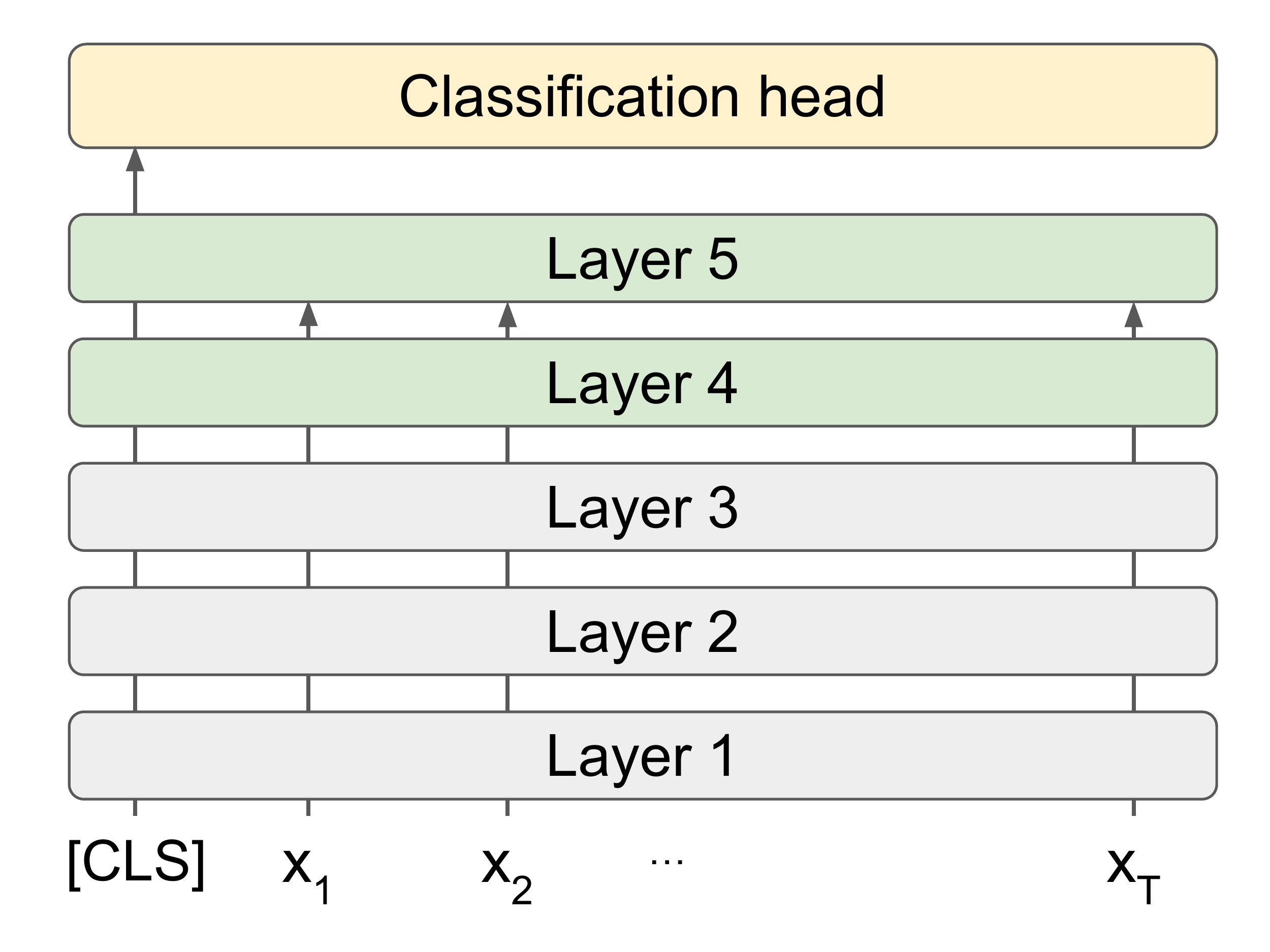}
			\vspace{-5pt}
			\caption{$(4,5,5)$}
			\label{fig:partial-4}
        \end{subfigure}
        \vspace{-6pt}
		\caption{%
		We present our strategies with a toy model of $L=5$ layers and $\ell^* = 3$. 
		The green layers will be tuned (e.g. fine-tuned or adapter-tuned) during the task-specific training while the grey layers are not. The white layers are dropped from the tuning and inference procedures, thus further reducing the computation and memory cost.
	    }\label{fig:partial}
	\end{center}
	\vspace{-16pt}
\end{figure*}

Suppose that our metric $\nu$ can indeed measure the task-specialty of each layer.
Then it is natural to investigate how the knowledge of layer-wise task-specialty can be leveraged to improve the transfer learning methods; that is what the question \circtwo in \cref{sec:intro} is concerned with. 
Recall from \cref{sec:tech} that the major paradigms of transfer learning use all the layers of the given PLM by default. 
We propose to select a subset of the layers based on the their task-specialty which will benefit all the paradigms of transfer learning: they may be able to only use the selected layers yet still achieve strong performance. 
Only using the selected layers will result in significant reduction of computation cost:%
\begin{itemize}[leftmargin=*,noitemsep,topsep=0pt]
    \item In fine-tuning, only the parameters of the selected layers will be updated. 
    \item In adapter-tuning, adapters are only added to the selected layers but not all the layers. 
    \item In prefix-tuning, we ``deep-perturb'' fewer layers.
\end{itemize}
A smaller number of task-specific parameters means not only less training cost but also less storage cost and less inference cost. 

\paragraph{Strategy-I: $\ell^*$-down.}
We use $\ell^*$ to denote the layer which achieves the best task-specialty: i.e., $\ell^* \defeq \argmin_{\ell} \nu^{(\ell)}$.
Our first strategy is motivated by the following intuition: if layer $\ell^*$ has already been well-specialized in the given task, then it may suffice to just mildly tune it along with a few layers below it. 
Meanwhile, we may keep the classification head on the top layer $L$ or move it to the best-specialized layer $\ell^*$: the former still utilizes the higher layers in training and inference; the latter does not and thus will result in even less computation and memory cost. 

Technically, we use $(\ell_{\text{bottom}}, \ell_{\text{top}}, \ell_{\text{head}})$ to denote the strategy of selecting the layers $\ell_{\text{bottom}}, \ell_{\text{bottom}}+1, \ldots, \ell_{\text{top}}$ and connect the classification head to the layer $\ell_{\text{head}}$. 
Then all the instances of our first strategy can be denoted as $(\ell_{\text{bottom}}, \ell^*, L)$ or $(\ell_{\text{bottom}}, \ell^*, \ell^*)$ with appropriately chosen $\ell_{\text{bottom}}$. 
\cref{fig:partial-1,fig:partial-2,fig:partial-3} illustrate a few specific instances of our $\ell^*$-down strategy.

\paragraph{Strategy-II: $\ell^*$-up.}
Alternative to the $\ell^*$-down strategy, our second strategy is to select the layers {above} the best-specialized layer $\ell^*$ and we call it $\ell^*$-up strategy.
Intuitively, if layer $\ell^*$ is already well-specialized in the given task, then what we need is perhaps just a powerful classification head. 
That is, we can regard the higher layers $\ell^*+1, \ldots, L$ along with the original classification head $f$ as a new ``deep'' classification head and then tune it to better utilize the layer $\ell^*$ representations. 

In principle, all the instances of our second strategy can be denoted as $(\ell^*+1, \ell_{\text{top}}, \ell_{\text{top}})$ or $(\ell^*+1, \ell_{\text{top}}, L)$ since we may select the layers up through $\ell_{\text{top}} \leq L$ and move the classification head $f$ to layer $\ell_{\text{top}}$. 
\cref{fig:partial-4} shows an instance of our $\ell^*$-up strategy.

Note that our $(\ell_{\text{bottom}}, \ell_{\text{top}}, \ell_{\text{head}})$ notation can apply to the conventional layer-selecting strategies as well. 
For example, $(1,L,L)$ denotes the naive option of tuning all the layers of the given PLM; $(L-2,L,L)$ denotes a baseline method of only selecting the top three layers.%

\section{Experiments}\label{sec:exp}
We evaluated the effectiveness of our task-specialty metric along with our layer-selecting strategies through extensive experiments on the six classification tasks of the GLUE benchmark~\citep{wang-etal-2018-glue}. 
The tasks are: CoLA, MNLI, MRPC, QNLI, QQP, and SST-2.%
All of them are sequence-level classification tasks related to natural language understanding, thus being very different from how language models are pretrained. 

We chose the widely accepted RoBERTa model~\citep{liu2019roberta} to be our PLM and used the pretrained roberta-large instance (355M parameters) downloaded from HuggingFace~\citep{wolf2020huggingface}. 
Our experiments are mainly conducted with this model. 
We also experimented with DeBERTa~\citep{he2020deberta} to investigate whether our methods generalize across models:\footnote{\citet{bowman2022dangers} advocate that it is important to experiment with more than one pretrained models before drawing any general conclusions about ``pretrained language models''.}
those results are in \cref{app:deberta} and are similar to the RoBERTa results.
Prior work~\citep{mosbach2020stability} found that fine-tuning RoBERTa on GLUE could be unstable, so we ran each of our experiments with five random seeds and reported the means and standard errors. 
Experiment details (e.g., hyperparameters) can be found in \cref{app:exp}.

Our code is implemented in PyTorch~\citep{paszke-17-pytorch} and heavily relies on HuggingFace.
It will be released after the paper is published.
Implementation details can be found in \cref{app:code}.

\subsection{Answer to Question \circone: Hidden State Variability Ratio Measures Task-Specialty}\label{sec:nc_pretrain}\label{sec:finetune}
\begin{figure}[t]
	\begin{center}
	    \begin{subfigure}[t]{0.48\linewidth}
			\includegraphics[width=0.99\linewidth]{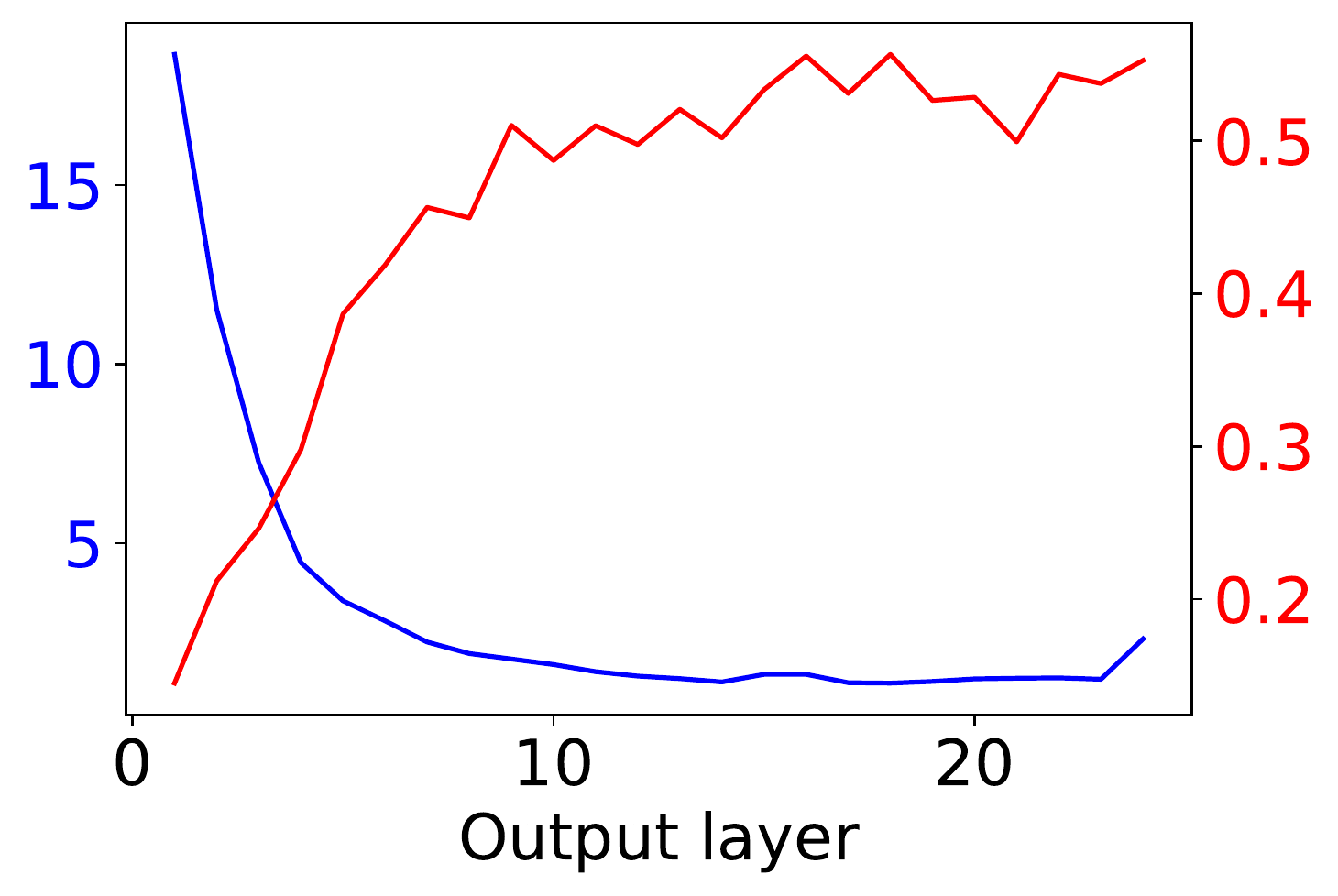}
			\caption{CoLA}
		\end{subfigure}
		~
		\begin{subfigure}[t]{0.48\linewidth}
			\includegraphics[width=0.99\linewidth]{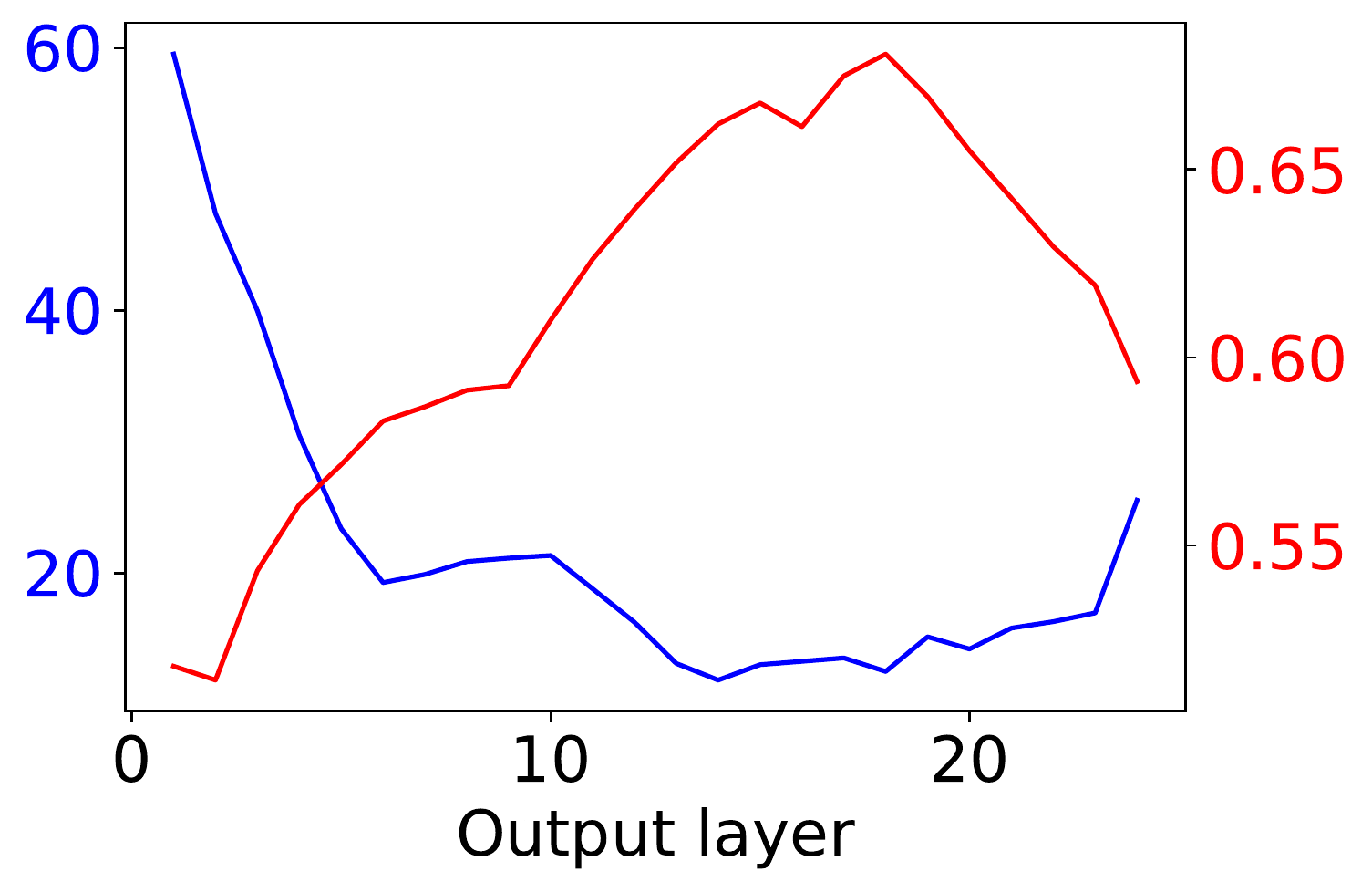}
			\caption{MNLI}
		\end{subfigure}
		
		\begin{subfigure}[t]{0.48\linewidth}
			\includegraphics[width=0.99\linewidth]{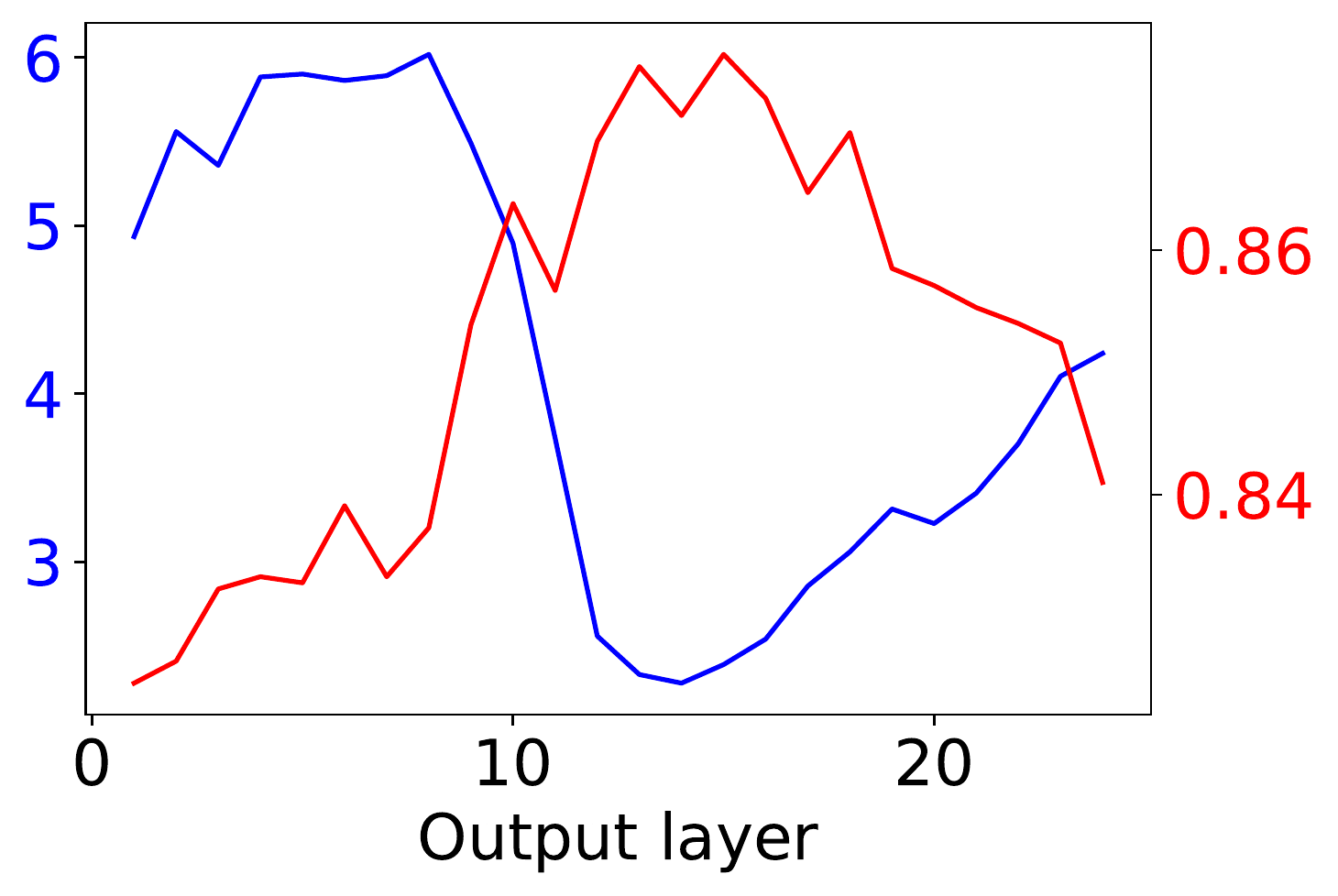}
			\caption{MRPC}
		\end{subfigure}
		~
		\begin{subfigure}[t]{0.48\linewidth}
			\includegraphics[width=0.99\linewidth]{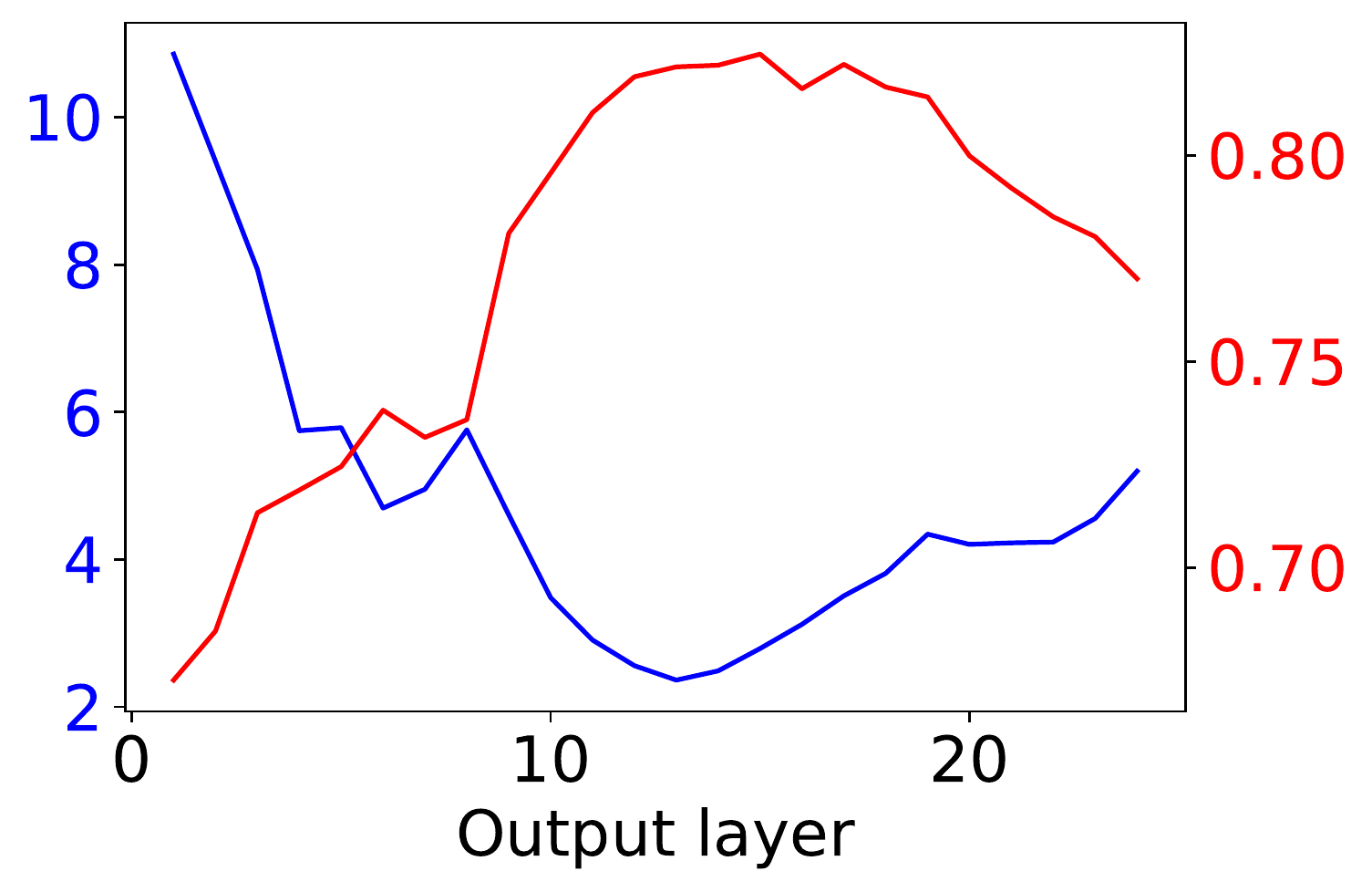}
			\caption{QNLI}
		\end{subfigure}
		
		\begin{subfigure}[t]{0.48\linewidth}
			\includegraphics[width=0.99\linewidth]{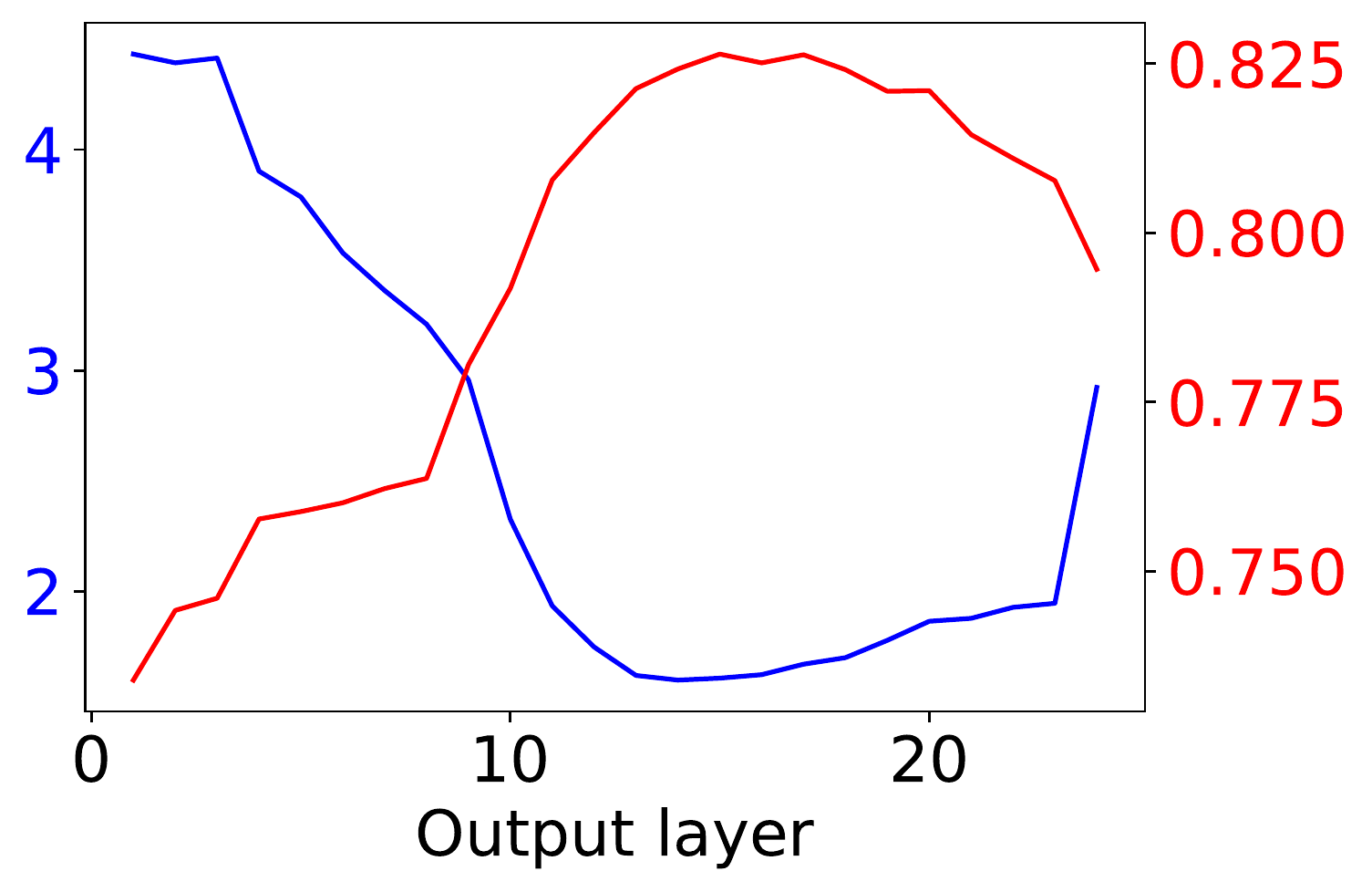}
			\caption{QQP}
		\end{subfigure}
		~
		\begin{subfigure}[t]{0.48\linewidth}
			\includegraphics[width=0.99\linewidth]{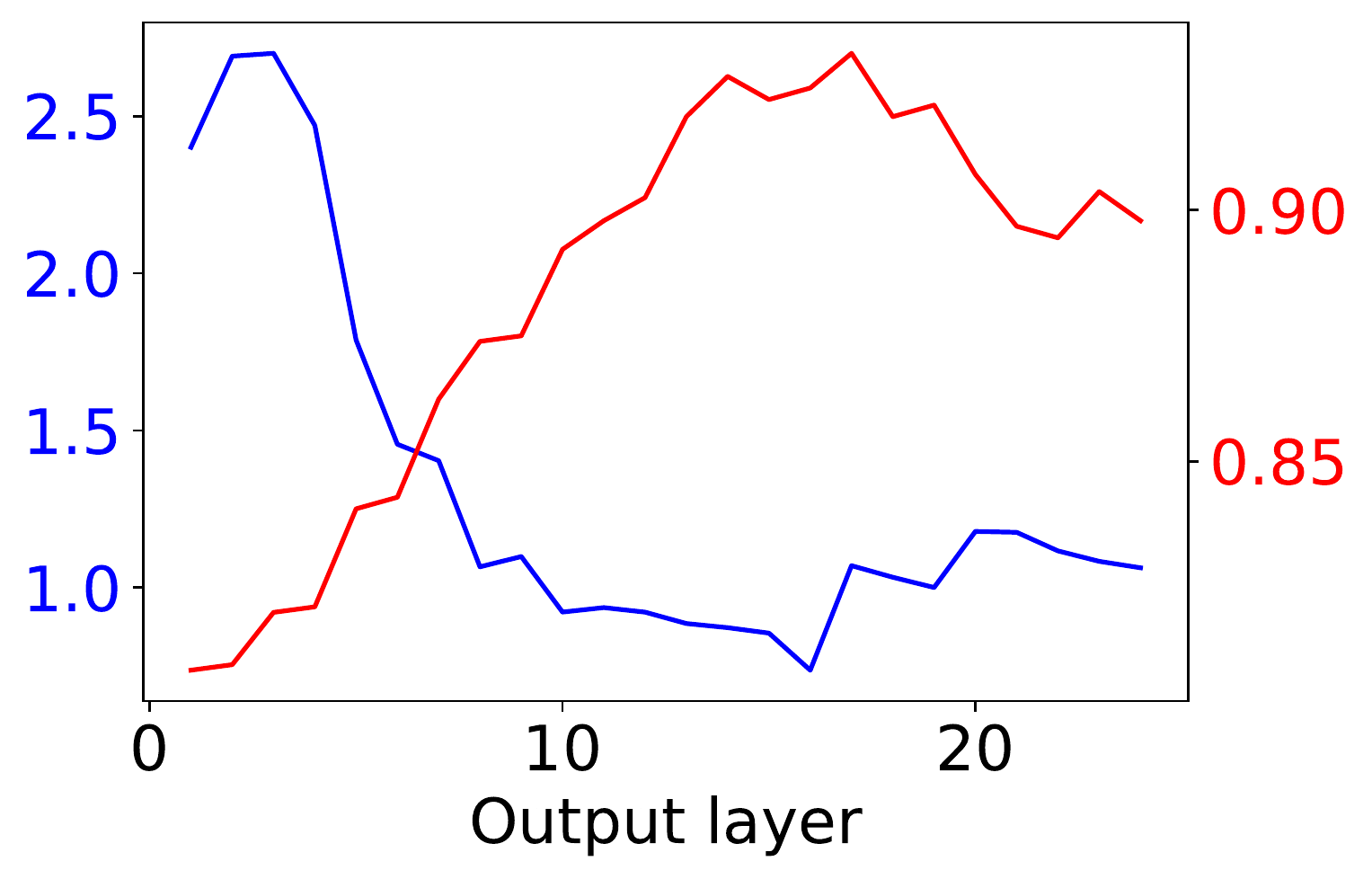}
			\caption{SST-2}
		\end{subfigure}
		\vspace{-8pt}
		\caption{The task-specialty metric (blue) and probing performance (red) of each layer of a pretrained RoBERTa model. Each figure is a GLUE task.}\label{fig:nc_pretrain}
	\end{center}
	\vspace{-6pt}
\end{figure}

For each task, we computed the task-specialty $\nu^{(1)}, \ldots, \nu^{(24)}$ (defined in \cref{sec:var}) for all the 24 layers. 
They are plotted as blue curves in \cref{fig:nc_pretrain}: as we can see, the middle layers ($10\leq\ell\leq15$) tend to have the lowest $\nu$ on most tasks. 

Then we probed the pretrained RoBERTa: for each layer $\ell$, we trained a classification head (see \cref{sec:tech}) that reads the hidden state $\vec{h}_{n}^{(\ell)}$ and evaluated it on the held-out development set. 
The probing performance is plotted as red curves in \cref{fig:nc_pretrain}: as we can see, the middle layers ($10\leq\ell\leq15$) tend to achieve the highest scores.%

How much is the probing performance correlated with the task-specialty metric $\nu$? 
To answer that question, we regressed the probing performance on $\nu$ and found that they are highly correlated. 
All the slope coefficients are negative, meaning that a low $\nu$ predicts a high score. 
All the $R^2$ are high, meaning that a large fraction of the probing performance variation can be explained by the variation in the task-specialty score. %
Remarkably, $R^2=0.97$ on QQP. 
Detailed results (e.g., fitted lines) are in \cref{fig:nc_regress_pretrain} of \cref{app:nc_pretrain}.

This set of experiments answers our question \circone (\cref{sec:intro}): yes, we can measure the task-specialty of each layer of a given PLM on a given task and our metric $\nu$ doesn't require task-specific tuning.%

Furthermore, we fully fine-tuned a RoBERTa on each task and obtained the $\nu$ and probing performance of the fine-tuned models. 
The results are presented in \cref{fig:nc_finetune,fig:nc_regress_finetune} of \cref{app:finetune}. 
After full fine-tuning, strong correlation between the probing performance and the task-specialty $\nu$ is still observed, but now higher layers are observed to have lower $\nu$ and stronger probing performance. 
That is because the parameters of higher layers have received stronger training signals back-propagated from the classification heads, which aim to specialize the full model on the tasks.

\subsection{Answer to Question \circtwo: Task-Specialty Helps Layer Selection}\label{sec:progressive}
As discussed in \cref{sec:strategy}, our task-specialty-based layer-selecting strategies are supposed to help improve the computation efficiency of transfer learning and they are compatible with all the major paradigms of transfer learning. 
We evaluated our strategies by pairing them with the widely adopted fine-tuning and adapter-tuning methods. 
In this section, we show and discuss our empirical results. 

\paragraph{Layer selection for fine-tuning.}\label{sec:ls_finetune}
For each task, we experimented with our $\ell^*$-down and $\ell^*$-up strategies.
The best-specialized layers $\ell^*$ are those with the lowest task-specialty $\nu^{(\ell)}$.
They are 
\begin{table}[ht]\centering
\small
\begin{tabular}{lcccccc}\hline
 &CoLA&\!MNLI&\!MRPC&\!QNLI&\!QQP&\!SST-2\\
$\ell^*$& 18 & 14 & 14 & 13 & 14 & 16\\
\hline
\end{tabular}
\end{table}

\begin{figure}[t]
	\begin{center}
	    \begin{subfigure}[t]{0.48\linewidth}
		\includegraphics[width=0.95\linewidth]{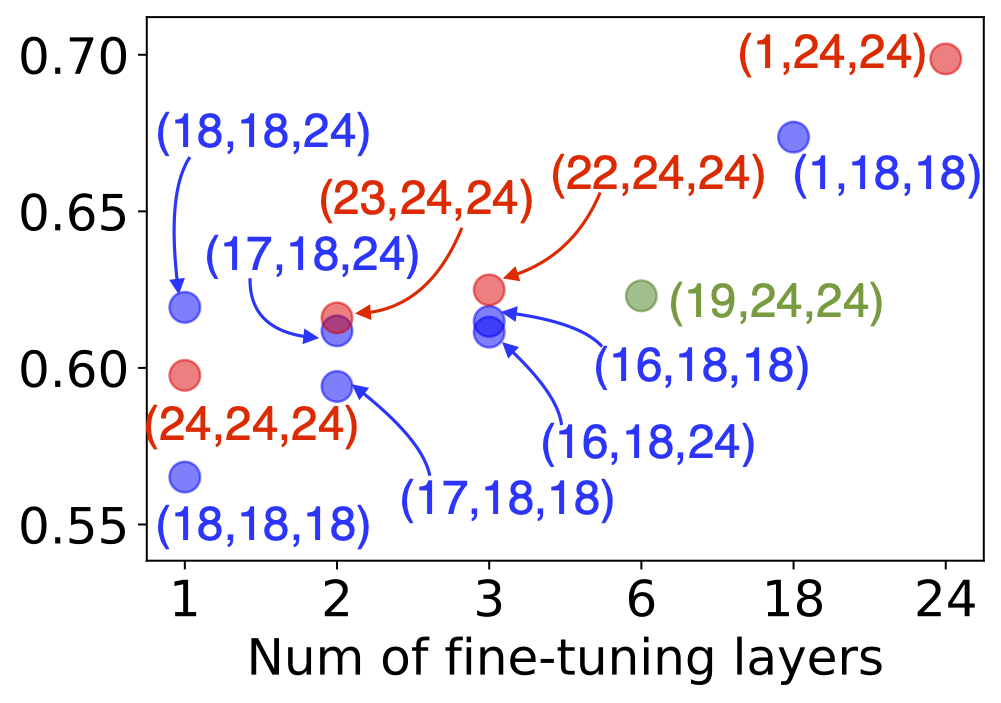}
			\vspace{-4pt}
			\caption{CoLA}
		\end{subfigure}
		~
		\begin{subfigure}[t]{0.48\linewidth}
			\includegraphics[width=0.95\linewidth]{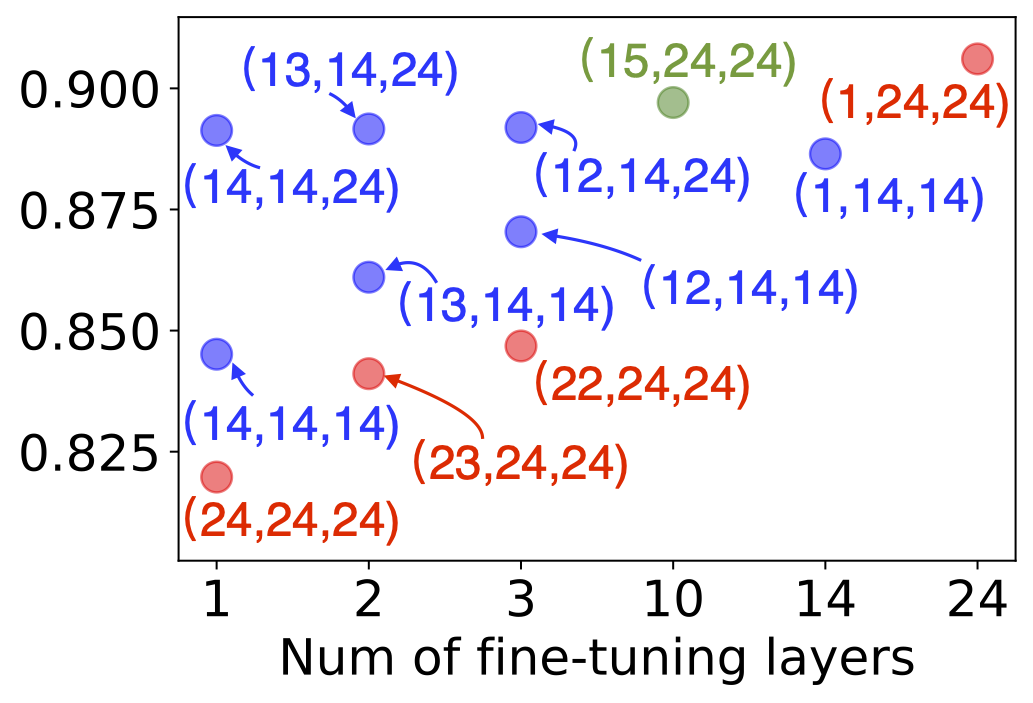}
			\vspace{-4pt}
			\caption{MNLI}
		\end{subfigure}
		\begin{subfigure}[t]{0.48\linewidth}
			\includegraphics[width=0.95\linewidth]{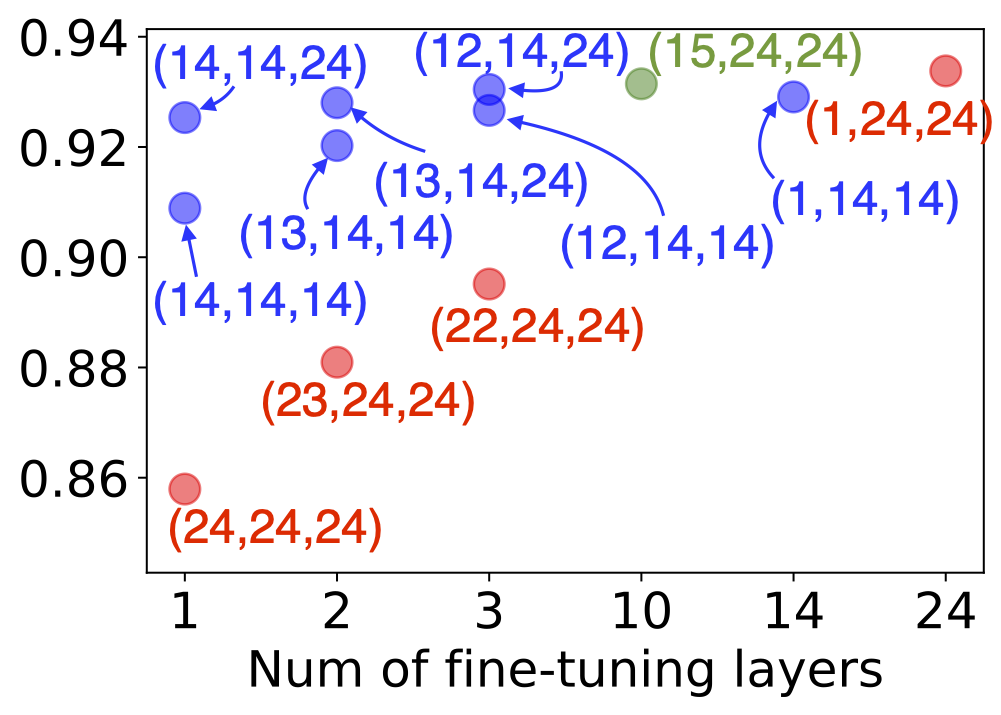}
			\vspace{-4pt}
			\caption{MRPC}
		\end{subfigure}
		~
		\begin{subfigure}[t]{0.48\linewidth}
			\includegraphics[width=0.95\linewidth]{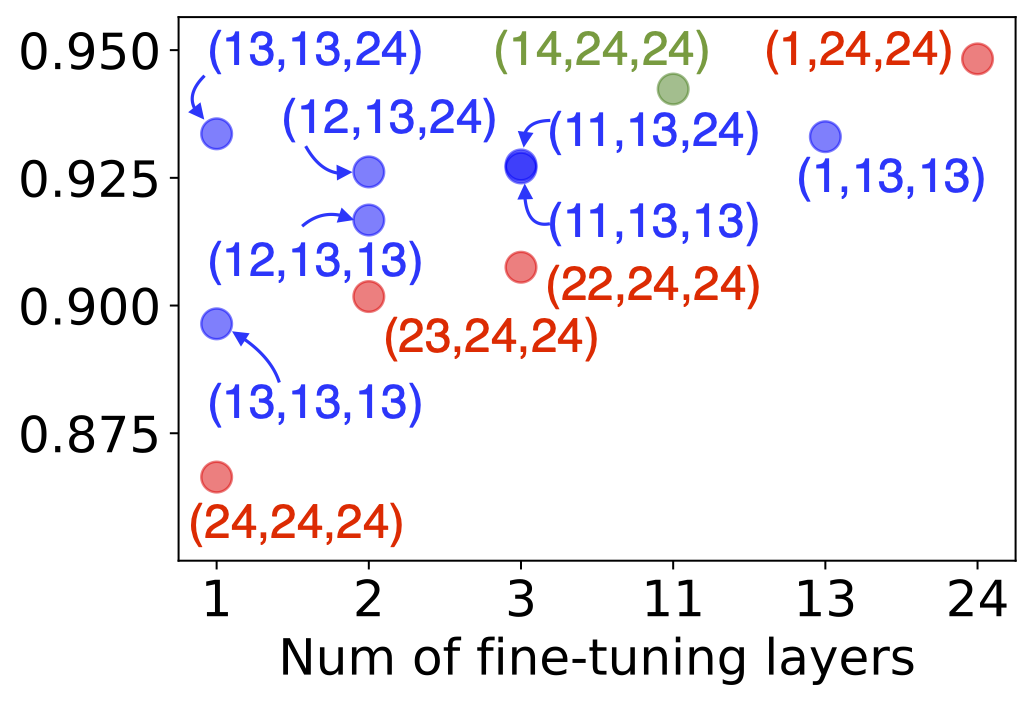}
			\vspace{-4pt}
			\caption{QNLI}
		\end{subfigure}
		\begin{subfigure}[t]{0.48\linewidth}
			\includegraphics[width=0.95\linewidth]{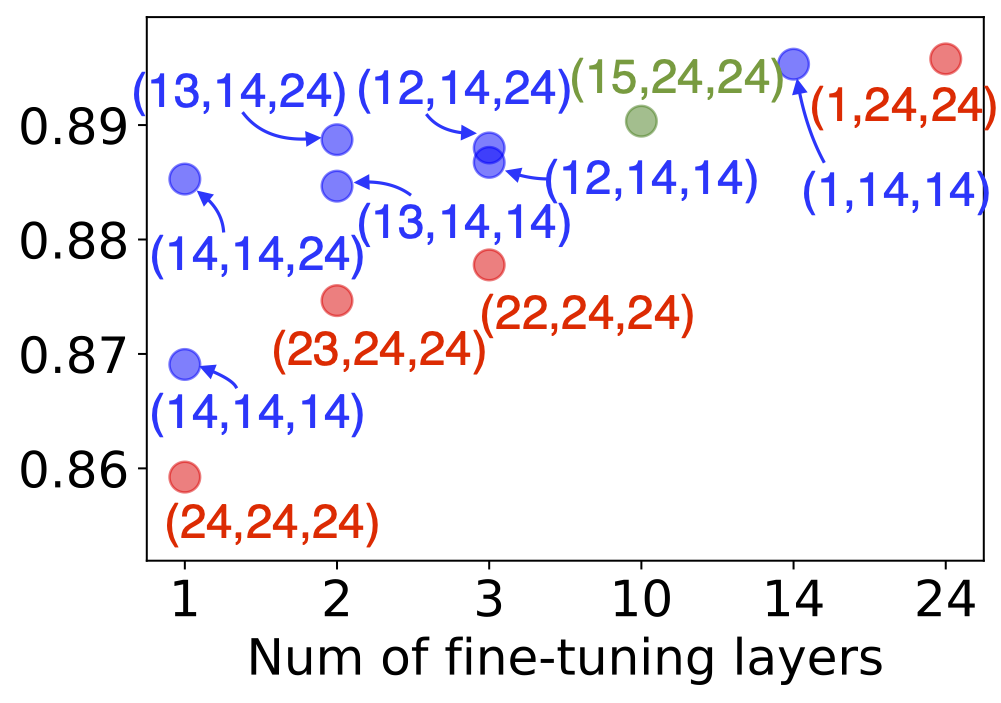}
			\vspace{-4pt}
			\caption{QQP}
		\end{subfigure}
		~
		\begin{subfigure}[t]{0.48\linewidth}
			\includegraphics[width=0.95\linewidth]{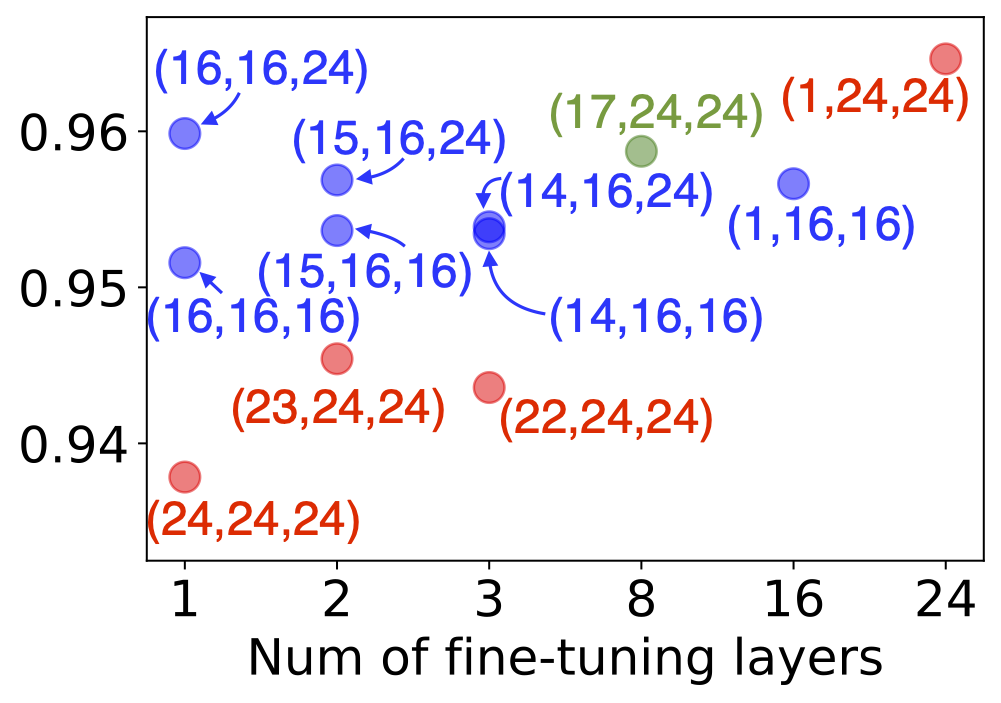}
			\vspace{-4pt}
			\caption{SST-2}
		\end{subfigure}
		\vspace{-6pt}
		\caption{%
		Task performance vs.\@ the number of selected layers for fine-tuning. 
		The annotation of each dot is its strategy identifier $(\ell_{\text{bottom}},\ell_{\text{top}},\ell_{\text{head}})$.
		}\label{fig:progressive_finetune}
	\end{center}
	\vspace{-6pt}
\end{figure}

Our experiment results are shown in \cref{fig:progressive_finetune}.
For the $\ell^*$-down strategy, we experimented with $(\ell_{\text{bottom}},\ell^*,\ell_{\text{head}})$ where $\ell_{\text{bottom}} \in \{1, \ell^*-2, \ell^*-1, \ell^*\}$ and $\ell_{\text{head}} \in \{ \ell^*, L \}$.
They are plotted as blue dots. 
For the $\ell^*$-up strategy, we experimented with $(\ell^*+1,L,L)$ and they are shown as green dots. 
For a comparison, we also experimented with the conventionally adopted baseline strategies $(1,L,L)$ and $(\ell_{\text{bottom}},L,L)$ with $\ell_{\text{bottom}} \in \{L-2, L-1, L\}$. 
They are red dots. 
The actual performance scores of the dots along with standard errors are in \cref{tab:roberta_finetune} of \cref{app:roberta_numbers}. 

As shown in \cref{fig:progressive_finetune}, when we are constrained by a budget of tuning only $\leq 3$ layers, 
our strategies can almost always lead to significantly higher performances than the baseline strategies.  
Moreover, the $(\ell_{\text{bottom}},\ell^*,L)$ strategies consistently outperform the $(\ell_{\text{bottom}},\ell^*,\ell^*)$ strategies across all the tasks. 
It means that the higher layers $\ell^*+1,\ldots,L$ are still very useful for performing well on the given task even if their parameters are not updated: they are perhaps able to help shape the training signals back-propagated through the tuned layers. 

Furthermore, the performance of only tuning the selected layers is often close to that of full fine-tuning. 
Remarkably, on QQP, our $(1,\ell^*,\ell^*)$ strategy even matches the performance of full fine-tuning yet only uses the bottom 14 layers; that is, it reduces the computation and storage cost by more than 40\%. 
Detailed discussion about wall-clock time saving can be found in \cref{app:roberta_cost}. 

Because most $\ell^*$ are in the middle of the PLM, we experimented another baseline of directly using the middle layer $\ell_{\text{mid}} = 13$ for the $24$-layer RoBERTa-large. This baseline is only implemented on CoLA and SST-2 because $\ell^*$ of other tasks is already the same as or very close to $\ell_{\text{mid}}$. 
We found that tuning around layer $\ell^*$ always outperforms tuning around $\ell_{\text{mid}}$ although the improvement is not large.
Result details are in \cref{tab:middle_finetune_roberta} of \cref{app:roberta_middle}.

\begin{figure}[t]
	\begin{center}
	    \begin{subfigure}[t]{0.48\linewidth}
			\includegraphics[width=0.95\linewidth]{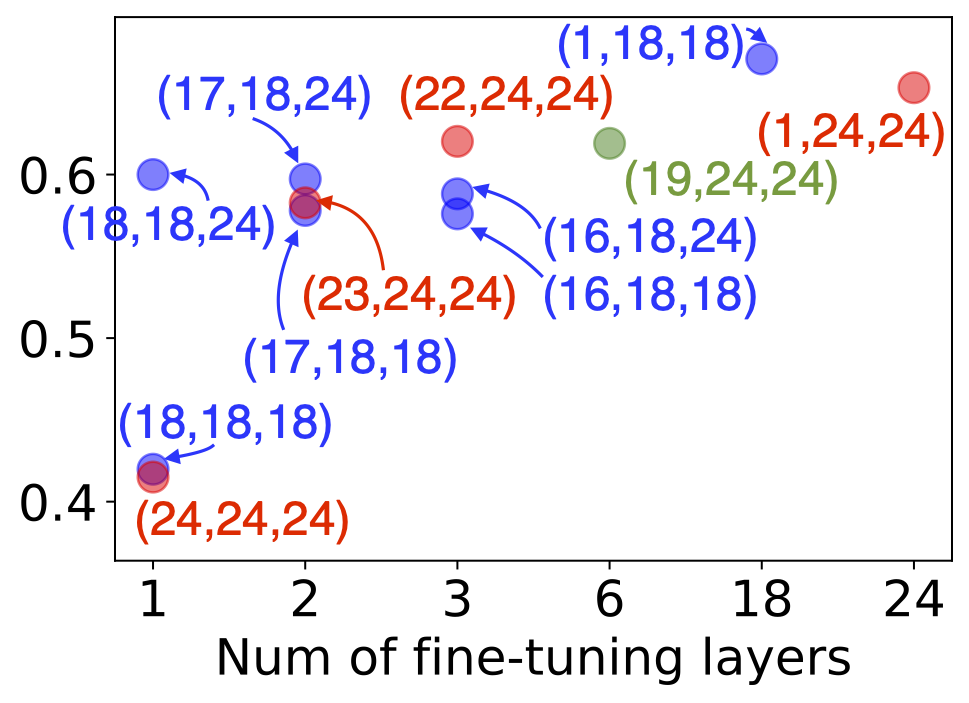}
			\vspace{-4pt}
			\caption{CoLA}
		\end{subfigure}
		~
		\begin{subfigure}[t]{0.48\linewidth}
            \includegraphics[width=0.95\linewidth]{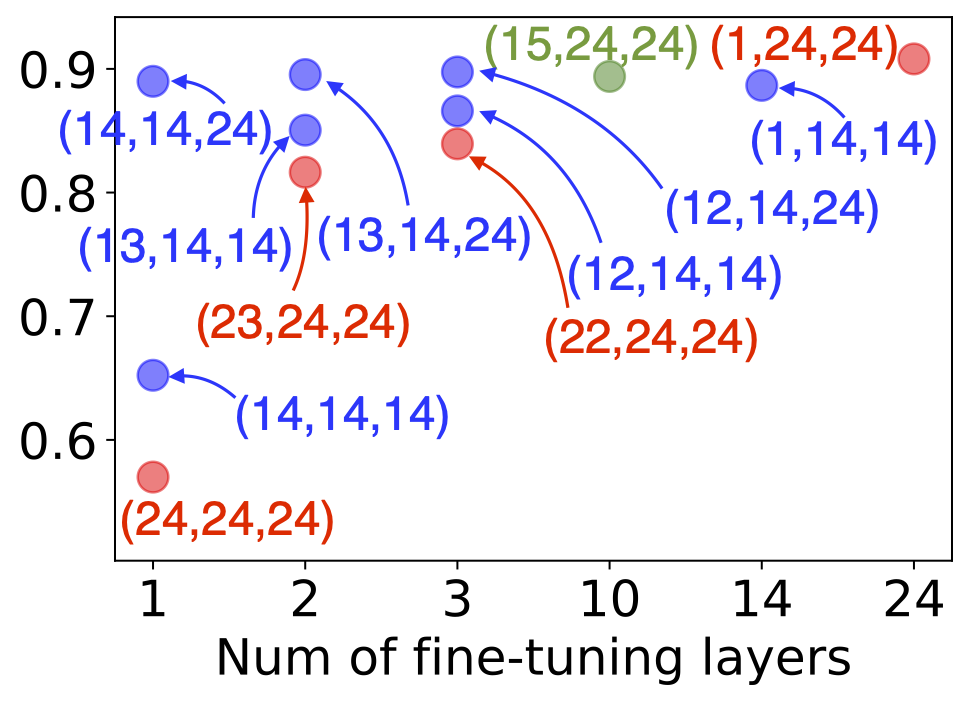}
			\vspace{-4pt}
			\caption{MNLI}
		\end{subfigure}
		\begin{subfigure}[t]{0.48\linewidth}
			\includegraphics[width=0.95\linewidth]{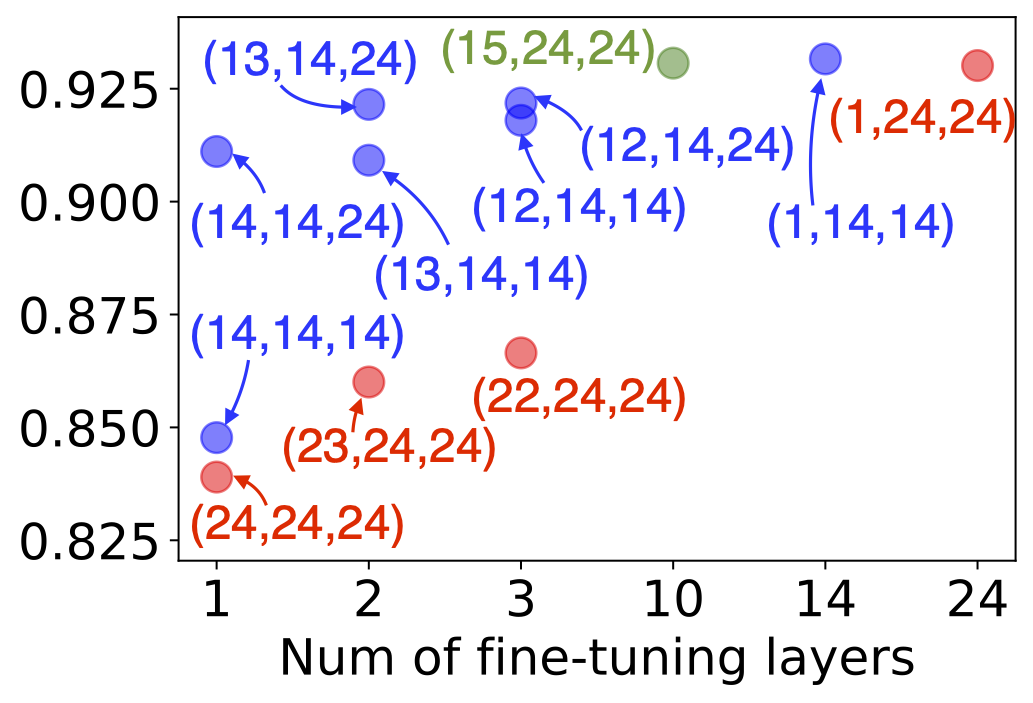}
			\vspace{-4pt}
			\caption{MRPC}
		\end{subfigure}
		~
		\begin{subfigure}[t]{0.48\linewidth}
			\includegraphics[width=0.95\linewidth]{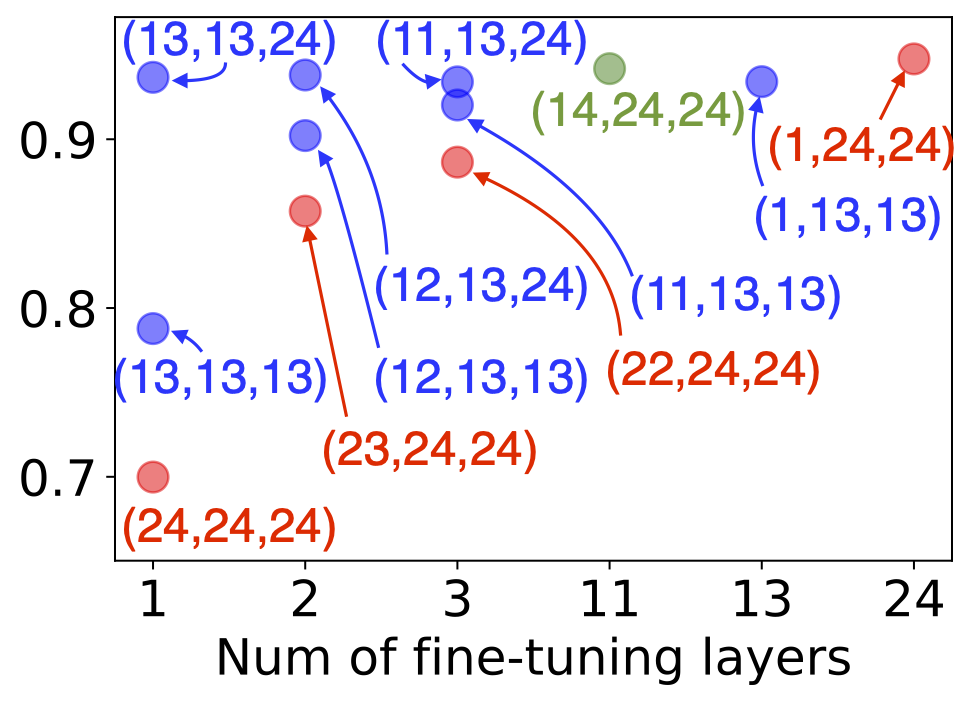}
			\vspace{-4pt}
			\caption{QNLI}
		\end{subfigure}
		\begin{subfigure}[t]{0.48\linewidth}
			\includegraphics[width=0.95\linewidth]{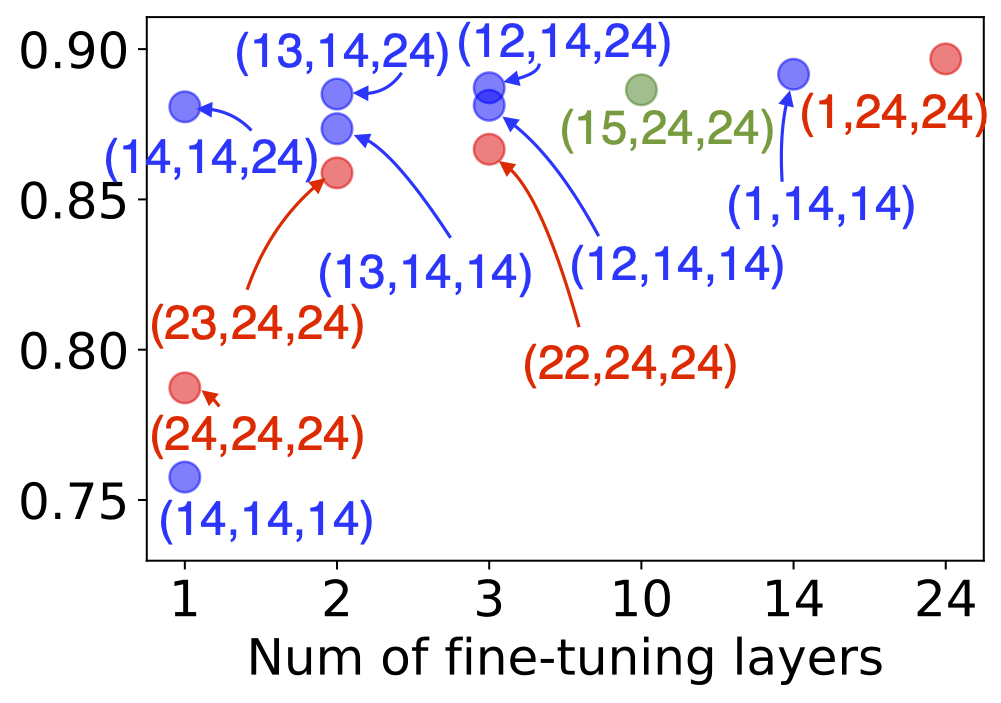}
			\vspace{-4pt}
			\caption{QQP}
		\end{subfigure}
		~
		\begin{subfigure}[t]{0.48\linewidth}
			\includegraphics[width=0.95\linewidth]{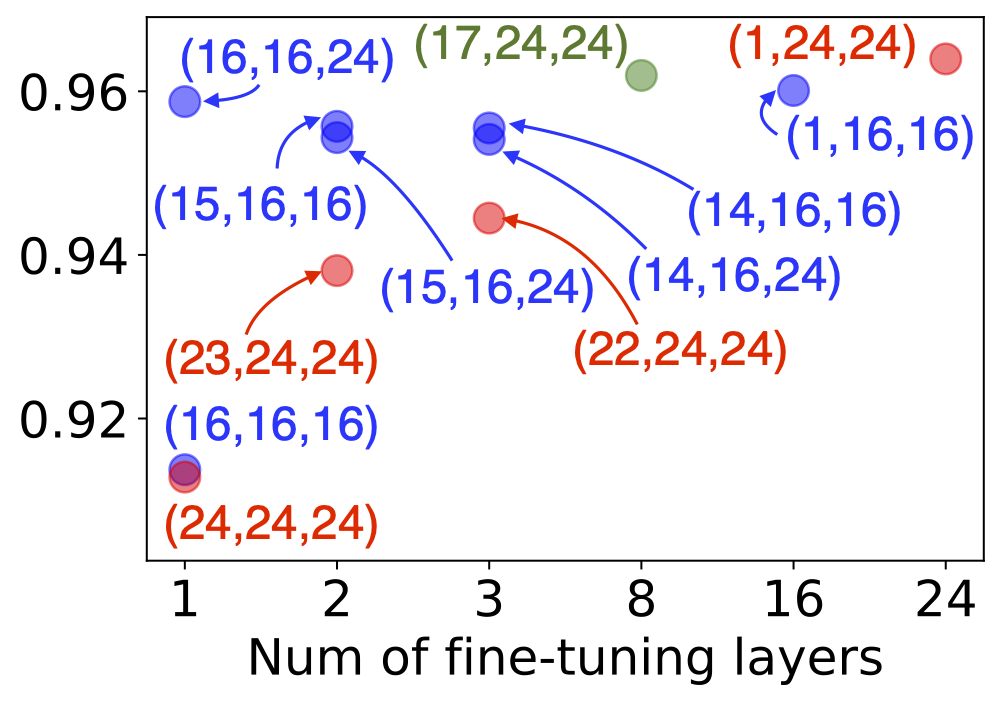}
			\vspace{-4pt}
			\caption{SST-2}
		\end{subfigure}
		\vspace{-6pt}
		\caption{
		Task performance vs.\@ the number of selected layers for adapter-tuning. 
		The annotation of each dot is its strategy identifier $(\ell_{\text{bottom}},\ell_{\text{top}},\ell_{\text{head}})$.
		}\label{fig:progressive_adapter}
	\end{center}
	\vspace{-6pt}
\end{figure}

\paragraph{Layer selection for adapter-tuning.}\label{sec:adapter}
For adapter-tuning, we implemented the earliest adapter architecture designed by \citet{houlsby2019parameter}.
We experimented with the same set of our strategies and baseline strategies as in the fine-tuning experiments. 
The results are plotted in \cref{fig:progressive_adapter}. 
Detailed numbers including standard errors are listed in \cref{tab:roberta_adapter} in \cref{app:roberta_numbers}. 

Like for fine-tuning, in most cases, our strategies are significantly better than the baseline strategies under the budget of tuning only $\leq 3$ layers. 
The $(\ell_{\text{bottom}},\ell^*,L)$ strategies also tend to outperform the $(\ell_{\text{bottom}},\ell^*,\ell^*)$ strategies.  

What's impressive is that the performance of only adapter-tuning the selected layers is often comparable to that of full adapter-tuning. 
Surprisingly, on MNLI and QNLI, our $(\ell^*-2,\ell^*,L)$ and $(\ell^*-1,\ell^*,L)$ strategies match the full adapter-tuning but only need 12\% and 8\% of the trainable parameters as full adapter-tuning respectively. 

 We also implemented the middle layer baseline for adapter-tuning on CoLA and SST-2. 
The results are listed in \cref{tab:middle_adapter_roberta} of \cref{app:roberta_middle}. 
In most cases, $\ell^*$ outperforms $\ell_{\text{mid}}$ with the same number of adapted layers.%

Interestingly, \citet{houlsby2019parameter} also found that not every layer in the PLM is equally important for adapter-tuning. 
For example, they experimented with a 12-layer BERT model and found that ablating the layer-7 adapters will lead to a much larger performance drop than ablating those of the top three layers on the CoLA task. 

\subsection{Is Our Metric the Only Option?}\label{sec:only}
In this section, we will discuss a few potential alternatives to our proposed task-specialty metric $\nu$. 
The introduction of the alternatives will be brief and only focused on their definitions and intuitions; more details can be found in \cref{app:only}. 
\begin{table}[ht]\centering
\small
\begin{tabular}{lccc}\hline
Task & $\rho(\nu, s)$ & $\rho(\text{CCA}, s)$ & $\rho(\varrho, s)$\\\hline
CoLA & -0.901 & 0.977 & -0.914\\
MNLI & -0.885 & 0.977 & -0.493\\
MRPC & -0.874 & 0.934 & -0.593\\
QNLI & -0.915 & 0.988 & -0.514\\
QQP & -0.984 & 0.989 & -0.723\\
SST-2 & -0.913 & 0.986 & -0.798\\
\hline
\end{tabular}
\vspace{-4pt}
\caption{Correlations between different metric (our $\nu$, CCA, effective rank) and probing performance $s$}\label{tab:correlation}
\vspace{-6pt}
\end{table}

\paragraph{Canonical correlation analysis.}\label{sec:cca}
A potential alternative to our proposed metric is the canonical correlation between the hidden states $\vec{h}_n^{(\ell)}$ and the class labels $y_n$. 
Canonical correlation analysis (CCA) involves learning a series of projection parameters $(\vec{v}_1,\vec{w}_1), \ldots, (\vec{v}_J,\vec{w}_J)$ such that each $(\vec{v}_j,\vec{w}_j)$ maximizes the correlation between $\vec{v}_j^{\top} \vec{h}_n^{(\ell)}$ and $\vec{w}_j^\top \vec{y}_n$ under certain constraints: $\vec{y}_n$ is a one-hot vector with its $y_n$\th entry being one.

\paragraph{Numerical Rank.}\label{sec:rank}%
Another potential alternative is the rank-based metric proposed by \citet{zhou2022optimization}. 
It is also inspired by the neural collapse phenomenon. 
The intuition is: if the sequence-level representations $\vec{h}_{n}^{(\ell)}$ of the same $\mathcal{G}_y^{(\ell)}$ exhibit a low variability, then the matrix $\vec{H}_{y}^{(\ell)} = [\ldots \vec{h}_{n}^{(\ell)} \ldots]$ formed by the vectors in $\mathcal{G}_y^{(\ell)}$ will have a low rank since its columns will be similar. 
The rank of $\vec{H}_{y}^{(\ell)}$ can be estimated by $\varrho_y^{(\ell)} \defeq\frac{\|\vec{H}_y^{(\ell)} \|_{*}^2}{\| \vec{H}_y^{(\ell)}\|_{F}^2}$ where $\| \|_*$ is the nuclear norm (i.e., the sum of singular values) and $\| \|_F$ is the Frobenius norm. 
The rank-based metric is the average over $y$: $\varrho^{(\ell)} \defeq \frac{1}{|\mathcal{Y}|} \sum_{y\in\mathcal{Y}} \varrho_y^{(\ell)}$. 
The correlation between $\varrho^{(\ell)}$ and the probing performance is presented in \cref{tab:correlation}.%

\paragraph{Analysis.}
As we can see in \cref{tab:correlation}, our variability-based metric $\nu$ and CCA score both exhibit a high correlation (above 0.85) with the probing performance. 
The correlation of CCA is moderately better than ours and it is higher than 0.9 on all the tasks. 
However, CCA optimization is dependent on the regularization terms (see \cref{app:cca}) and thus requires cross-validation to choose the right set of hyperparameters. This makes it more involved to compute than the proposed variability metric. 
Technically, CCA requires a ``training procedure'' to fit a set of parameters ($\vec{v}_j$ and $\vec{w}_j$) though the computation of that part is as light as calculating our metric. 
Both CCA and our metric are scale invariant. 
Overall, the rank-based metric $\varrho$ has a lower correlation. 
That is because it only considers the within-group properties but ignores the between-group properties. 

\subsection{Ablation Studies and Analysis}\label{sec:ablation}\label{sec:analysis}

\paragraph{Averaged state vs.\@ CLS state.}\label{sec:cls}\label{sec:othernu}

The $\nu$ that we have reported so far are all computed using the average hidden state of each $\vec{x}_n$ as defined in \cref{sec:var}. 
We also experimented with the CLS token hidden states---i.e., $\vec{h}_{n}^{(\ell)} \defeq \vec{h}_{n,0}^{(\ell)}$---and found that the computation of $\nu$ became much less stable: e.g., on SST-2, the metrics $\nu^{(\ell)}$ of the pretrained RoBERTa are below 50 for most of the layers but can jump above 500 for a couple of layers.  
Intuitively, one would like to trust a $\nu^{(\ell)}$ curve that is at least \emph{locally smooth}.
Technically, local smoothness means a small local second-order derivative, which can be estimated by finite differences.
Therefore, we measure the local smoothness of a $\nu^{(\ell)}$ curve by the following quantity $\zeta$
\vspace{-3pt}
\begin{align*}
    \zeta \defeq \sum_{\ell=1}^{L-2} ({\nu}^{(\ell+2)} - 2 {\nu}^{(\ell+1)} + {\nu}^{(\ell)})^2
\end{align*}
The results\footnote{We normalized the $\nu^{(\ell)}$ values before computing $\zeta$ so that the $\nu^{(\ell)}$ and $\zeta$ are more comparable across different choices of the sequence-level representation. Our normalization is: $\nu^{(\ell)} \gets \frac{\nu^{(\ell)} - \bar{\nu}}{\sigma}$ where $\bar{\nu}$ and $\sigma$ is the mean and standard deviation of the original $\nu^{(\ell)}$. 
} are presented in \cref{tab:nc_stable}. 
\begin{table}[ht]\centering
\small
\begin{tabular}{lrr}\hline
Task & $\zeta$ with averaged state & $\zeta$ with CLS state\\\hline
CoLA & 0.950 &0.949\\
MNLI & 1.347 &88.162\\
MRPC & 1.994 &33.038\\
QNLI & 3.256 &13.566\\
QQP & 1.640 &7.629\\
SST-2 & 3.604 &58.040\\
\hline
\end{tabular}
\caption{Overall local smoothness $\zeta$.}\label{tab:nc_stable}
\end{table}
As we can see, the $\zeta$ computed using the CLS token states is dramatically larger than that of using the averaged states on all the tasks except CoLA. 
It means that using the averaged states will give us a more trustable $\nu^{(\ell)}$ curve.%
Thus, we used the average hidden states throughout our experiments; we made this choice before seeing any task performance.

The actual $\nu^{(\ell)}$ curves computed using the CLS states are presented in \cref{fig:nc_pretrain_cls} of \cref{app:othernu}.

\paragraph{Robustness to data imbalance.}\label{sec:imbalance}
The datasets of GLUE tasks are almost perfectly balanced. 
So it remains a question whether our proposed task-specialty $\nu$ is still effective when the data is not imbalanced. 
To answer this question, we synthesized a series of data-imbalanced experiments on SST-2. 
In each experiment, we randomly sampled $N=20000$ training examples with a portion $p$ from the negative class where $p \in \{ 0.5, 0.25, 0.1, 0.05 \}$. 
We did the same analysis on MNLI; see results in \cref{app:mnli_sensitivity}.
\begin{figure}[t]
	\begin{center}
		\begin{subfigure}[t]{0.48\linewidth}
			\includegraphics[width=0.95\linewidth]{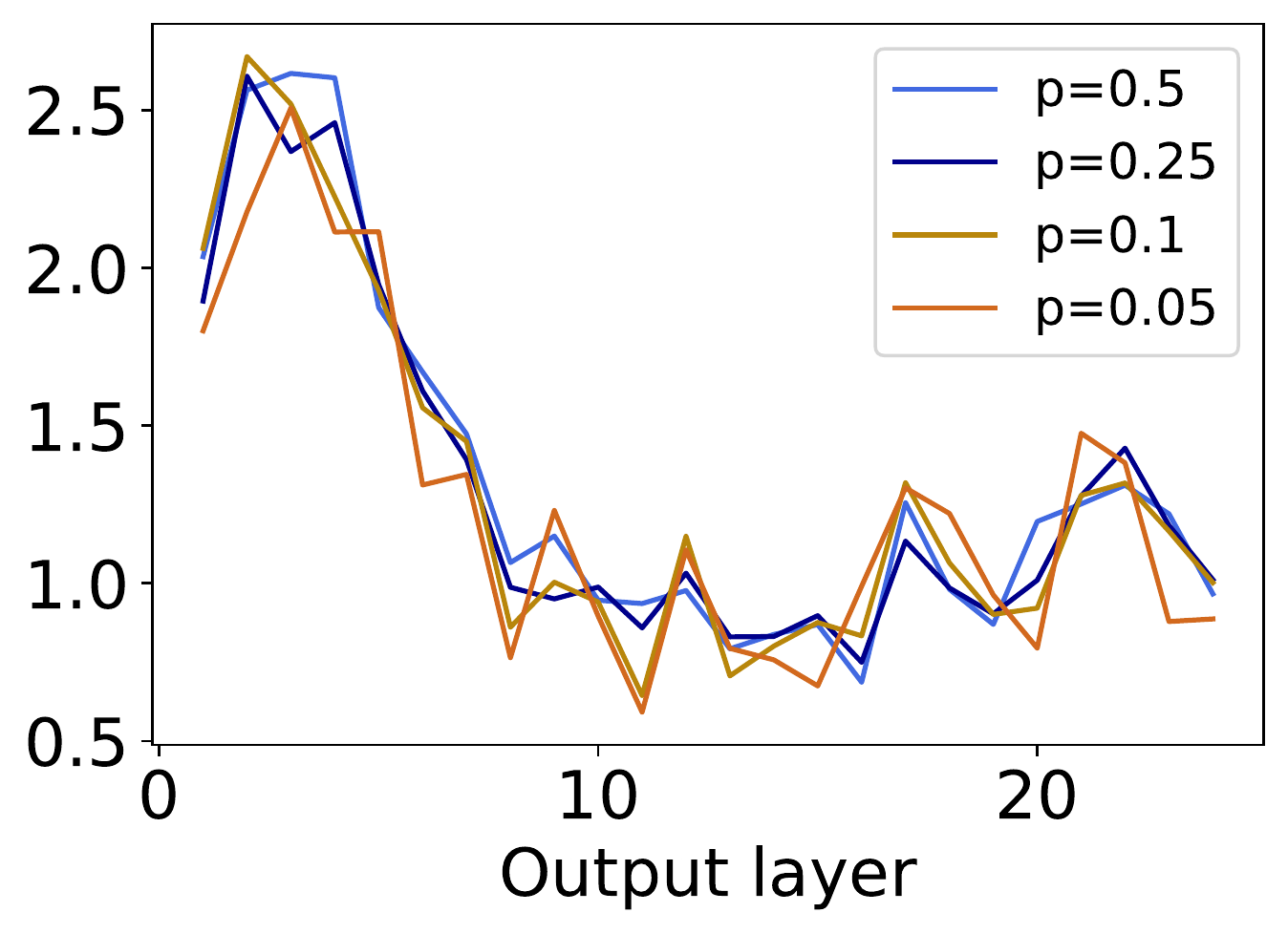}
			\vspace{-4pt}
			\caption{$\nu$ curves}
			\label{fig:layerwise_imbalance}
		\end{subfigure}
		\hfill
		\begin{subfigure}[t]{0.48\linewidth}
			\includegraphics[width=0.95\linewidth]{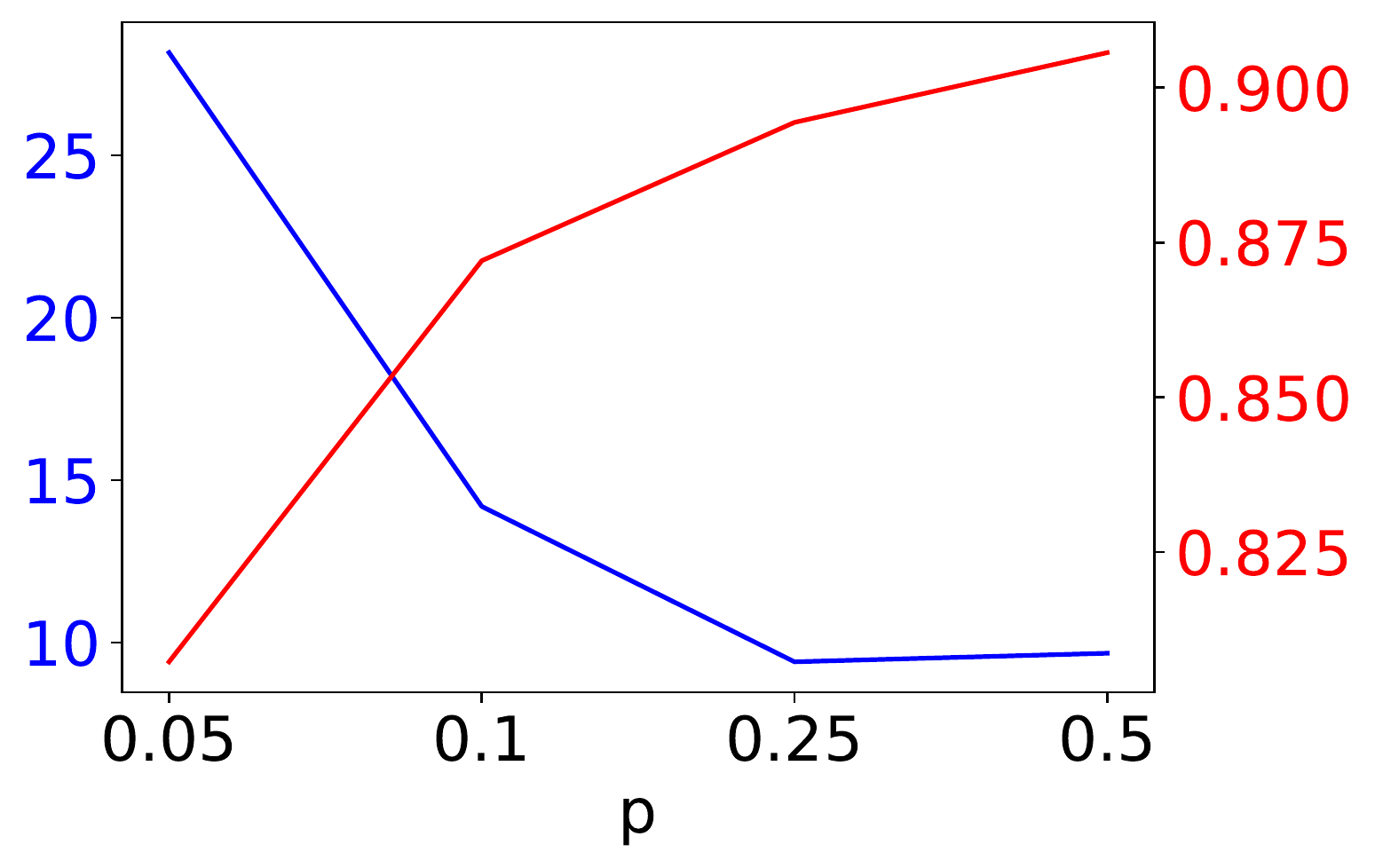}
			\vspace{-4pt}
            \caption{$\zeta$ (\textcolor{blue}{blue}) and $|\rho|$ (\textcolor{red}{red})}
			\label{fig:metric_imbalance}
		\end{subfigure}
	\vspace{-8pt}
	\caption{Results of data-imbalanced experiments on SST-2.}
	\end{center}
	\vspace{-4pt}
\end{figure}

For each experiment, we plot the $\nu^{(\ell)}$ curve in \cref{fig:layerwise_imbalance}. 
We also computed the local smoothness score $\zeta$ and the absolute value of the correlation $\rho(\nu,s)$ (defined in \cref{sec:only}) and plot them in \cref{fig:metric_imbalance}. 
Interestingly, the $\zeta$ in the case of $p=0.25$ is as low as that of $p=0.5$, though it becomes much larger when the data is extremely imbalanced (e.g., $p < 0.1$). 
Moreover, the $\rho(\nu,s)$ is still above 0.85 even when $p=0.1$. 
Those findings mean that our task-specialty metric is reasonably robust to data imbalance. 
More details are in \cref{app:nu_imbalance}.

\paragraph{Robustness to data scarcity.}\label{sec:scarcity}
We also examined the robustness of our metric to data scarcity. 
On SST-2, we conducted a series of experiments with varying number of training samples: $N \in {40000, 20000, 5000, 200}$. For each experiment, we made the sampled dataset label-balanced. 
See the results on MNLI in \cref{app:mnli_sensitivity}.
\begin{figure}[t]
	\begin{center}
		\begin{subfigure}[t]{0.48\linewidth}
			\includegraphics[width=0.95\linewidth]{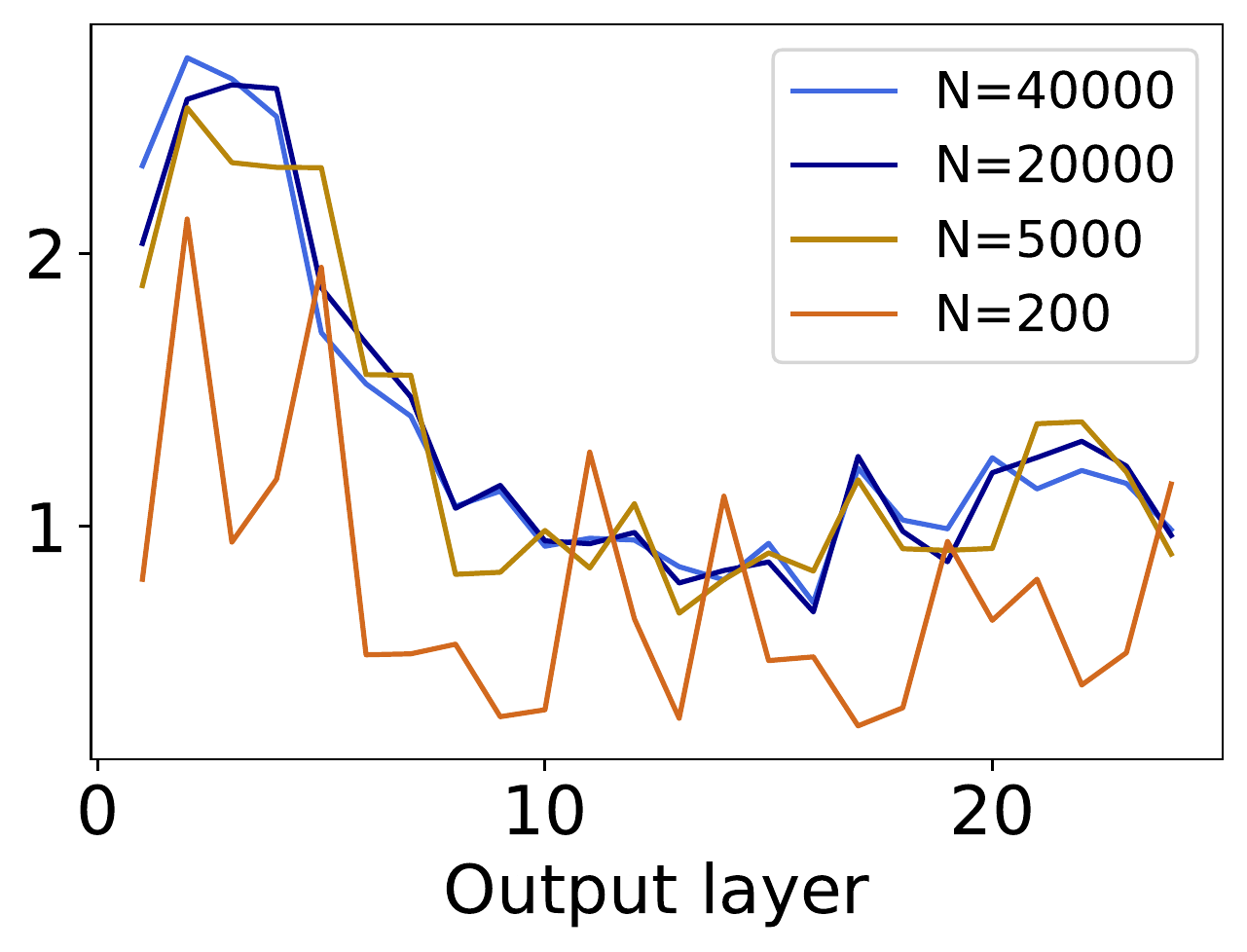}
			\vspace{-4pt}
            \caption{$\nu$ curves}
			\label{fig:layerwise_scarcity}
		\end{subfigure}
		\hfill
		\begin{subfigure}[t]{0.48\linewidth}
			\includegraphics[width=0.95\linewidth]{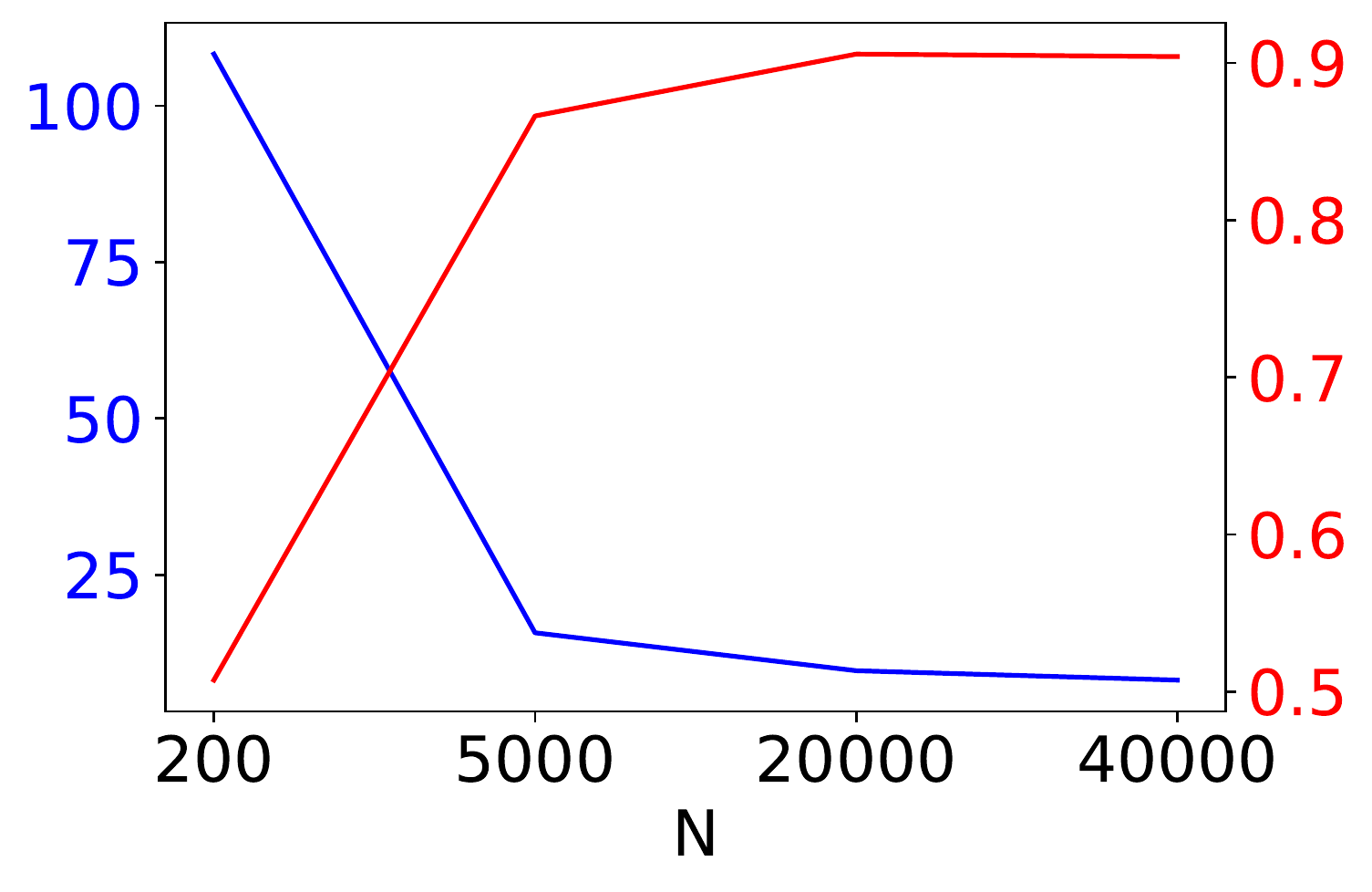}
			\vspace{-4pt}
			\caption{$\zeta$ (\textcolor{blue}{blue}) and $|\rho|$ (\textcolor{red}{red})}
			\label{fig:metric_scarcity}
		\end{subfigure}
	\vspace{-4pt}
	\caption{Results of data-scarce experiments on SST-2.}\label{fig:sensitivity_scarcity}
	\end{center}
	\vspace{-2pt}
\end{figure}

Like in the data-imbalanced experiments, we plot $\nu$ and $\zeta$ and $|\rho|$ in \cref{fig:sensitivity_scarcity}.
As we can see, for $N \geq 5000$, the $\nu$ curves almost perfectly overlap and the $\zeta$ and $|\rho|$ only slightly change with $N$. 
It means that the $\nu$ values are trustable as long as they are computed using thousands of examples. %
In the extremely data-scarce case of $N=200$, the $\nu$ curve becomes not trustable: after all, it will be extremely difficult to estimate $\vec{\Sigma}_{\text{w}}$ and $\vec{\Sigma}_{\text{b}}$ with so few samples.
However, even in the case of $N=200$, the middle layers tend to have the lowest $\nu^{(\ell)}$ which agree with the data-adequate cases. 
More details are in \cref{app:nu_sensitivity}.

\section{Related Work}\label{sec:related}
\paragraph{Analysis of PLMs.} Representations from PLMs have been widely studied using probing methods.\footnote{In this paper, we focus on the models that are pretrained only with the language modeling objective. Other pretraining objectives such as those of~\citet{sun2019ernie,wang2021kepler,raffel2020exploring} are out of our current scope.}
There is evidence showing that pre-trained features from intermediate layers are more transferable~\citep{tenney2019bert, rogers2020primer}. 
Additionally, \citet{voita2019bottom} show that masked language modelling objective introduces an auto-encoder like structure in the PLM, meaning the behaviour of the top most layers is similar to the bottom layers. Studies have also shown that the effects of fine-tuning are non-uniformly spread across different layers~\citep{peters2019tune, merchant2020happens, liu2019linguistic, phang2021fine}. 
Those findings challenge the default choice of tuning the entire PLM for adapting it to downstream tasks. While probing tools have been used to study task-specific layer importance~\citep{tamkin2020investigating, mosbach2020interplay}, the probing paradigm is parametric, hard to interpret, and is known to be unreliable~\citep{ravichander2020probing, belinkov2022probing, hewitt2019designing, voita2020information, pimentel2020information}. Instead, we propose to measure the layer-wise task-specialty of PLMs using a non-parametric tool that quantifies the task-specific variability of the hidden features.%

\paragraph{PLM-based transfer learning.} PLMs are widely used in the transfer learning setup to improve performance on a variety of downstream tasks. %
Typically, a task-specific classification layer is added on the top of the network and the entire network is trained to minimize the supervised task loss. Recently there has been growing interest in  parameter-efficient alternatives to this approach. A subset of these methods add a few new trainable parameters to the PLM while the pre-trained weights are frozen and are thus kept consistent across tasks~\citep{houlsby2019parameter, guo2020parameter, li2021prefix}.
Another set of methods either reparameterizes PLMs~\citep{hu2021lora} or chooses only a subset of the PLM parameters~\citep{voita2019analyzing, sajjad2020effect, gordon2020compressing, zaken2021bitfit}, thus reducing the number of trainable parameters for transfer learning.
In particular, the ``early exit'' methods~\citep{xin2020deebert,xin2021berxit,zhou2020bert} allow samples to pass through part of PLM if the prediction from a middle layer is trusted by the off-ramp following that layer.
This method can reduce inference cost but increase training cost because it adds a classification head to each hidden layer. 
Our technique can reduce both training and inference cost by tuning fewer layers and moving classification head to an intermediate layer. 

In this work we leverage our proposed metric of layer-wise task-specificity to make an informed decision to retain/drop and tune/freeze layers on the downstream task. There has been work studying the effect of dropping PLM layers~\citep{sajjad2020effect, phang2021fine, tamkin2020investigating} but their decision is driven by the performance on the downstream task itself and thus every new task will require a slew of ablation studies to find the applicability of each layer. Whereas, our task-specificity measure is completely parameter-free and is also agnostic to the specific transfer learning approach (fine-tuning, adapter-tuning, prefix-tuning) and thus complements the existing methods on parameter-efficient approaches~\citep{he2021towards}.

\paragraph{Neural collapse.} Our task-specialty metric $\nu$ is inspired by the neural collapse phenomenon. 
A surging line of work has been demystifying the training, generalization, and transferability of deep networks through \NC~\citep{kothapalli2022neural}. 
For training, recent works showed that \NC\ happens for a variety of loss functions such as cross-entropy~\citep{papyan2020prevalence,zhu2021geometric,fang2021exploring,ji2022an}, mean-squared error~\citep{mixon2020neural,han2022neural,zhou2022optimization,tirer2022extended}, and supervised contrastive loss \citep{graf2021dissecting}. 
For generalization, \citet{galanti2022on,galanti2022note} show that \NC\ also happens on test data drawn from the same distribution asymptotically, but not for finite samples \citep{hui2022limitations};
\citet{hui2022limitations,papyan2020traces} showed that the variability collapse is happening progressively from shallow to deep layers;
\citet{ben2022nearest} showed that test performance can be improved when enforcing variability collapse on features of intermediate layers; 
\citet{xie2022neural,yang2022we} showed that fixing the classifier as simplex ETFs improves test performance on imbalanced training data and long-tailed classification problems.
For transferability, \citet{kornblith2021why} showed that there is an inherent tradeoff between variability collapse and transfer accuracy.

\section{Conclusion}
In this paper, we present a comprehensive study on how to measure the task-specialty of each layer of a pretrained language model as well as how to leverage that knowledge to improve transfer learning. 
Our proposed layer-wise task-specialty metric is based on the variability of the hidden states of each layer given a task-specific corpus. 
Our metric is highly correlated with the layer-wise probing performance, though it is cheap to compute and does not require any training or hyperparameter tuning. 
We propose a couple of strategies based on the metric for selecting a subset of the layers to use in the PLM-based transfer learning methods. 
Extensive experiments demonstrate that our strategies can help fine-tuning and adapter-tuning achieve strong performance under a greatly reduced computation budget. 
Our strategies are complementary to all the major paradigms of PLM-based transfer learning and thus they will also benefit other methods.

\section*{Acknowledgements}
This work was supported by a research gift to the last author by Adobe Research. 
We thank the anonymous EMNLP reviewers and meta-reviewer for their constructive feedback. 
We also thank our colleagues at Toyota Technological Institute at Chicago as well as Dongji Gao (JHU), Yiyuan Li (UNC), Hao Tan (Adobe Research), and Zhihui Zhu (OSU) for helpful discussion. 

\section*{Limitations}\label{sec:limit}
Our main technical limitation is that the proposed metric only measures the task specificity from the perspective of variability.
Thus, it might underestimate the task specificity if the features has other kinds of good spacial structures with large within-class variability.
For example, concentric rings are separable but not linearly separable; see Fig-3 in \citet{hofmann2006support}. 
Although we have seen that our proposed $\nu$ is a good predictor for the final performance in all our experiments (\cref{sec:exp}), it is still possible that, for some tasks and some models, the layers with high $\nu$ can actually achieve good performance. 
Fortunately, such clustering structure is rare in the hidden space of deep neural networks.

Another technical limitation is that our proposed hidden state variability ratio only works for classification tasks. 
An open research question is how to generalize it to regression or generation tasks.

\section*{Ethics Statement}
In this work, we introduce a simple yet effective approach for substantially reducing the computation for transferring PLMs to downstream tasks.
Our proposed strategies obviate the need for tuning the entire model, which can significantly reduce the cost of computation and memory.
Therefore, they can help reduce greenhouse gas emissions and combat climate change.

However, our technical approaches involve pretrained language models for which a range of ethical concerns exist including privacy leakage, data bias, and vulnerablility to adversarial attacks.

\bibliography{nc-nlp}
\bibliographystyle{acl_natbib}

\clearpage
\newpage
\appendix

\section{Metric Details}\label{app:nu}

\subsection{Compared to the Neural Collapse Metric}\label{app:metric_diff}

Our task-specialty metric $\nu$ defined in \cref{sec:var} is a lot like the neural collapse metric proposed by~\citet{papyan2020prevalence} except that they assume a balanced dataset. 
First, they define the $\bar{\vec{h}}^{(\ell)}$ to be global mean vector: i.e., $\bar{\vec{h}}^{(\ell)} \defeq \frac{1}{N} \sum_{n=1}^{N} \vec{h}_{n}^{(\ell)}$, but we define it to be the mean of the within-group mean vectors. 
In data-balanced cases, those two definitions are equivalent. 
But when data is imbalanced, our version is better since it prevents the $\bar{\vec{h}}^{(\ell)}$ from being dominated by the group that has the largest number of samples. 

If we strictly follow \citet{papyan2020prevalence}, then our within-class variability $\vec{\Sigma}_{\text{w}}^{(\ell)}$ and between-class variability $\vec{\Sigma}_{\text{b}}^{(\ell)}$ will be defined to be
\begin{subequations}
\begin{align*}
    \vec{\Sigma}_{\text{w}}^{(\ell)}
    &\defeq 
    \frac{1}{N} 
    \sum_{y\in\mathcal{Y}}\sum_{\vec{h}\in\mathcal{G}_{y}^{(\ell)}} (\vec{h} -
    \bar{\vec{h}}_{y}^{(\ell)}) (\vec{h} - 
    \bar{\vec{h}}_{y}^{(\ell)})^{\top} \\
    \vec{\Sigma}_{\text{b}}^{(\ell)}
    &\defeq 
    \frac{1}{|\mathcal{Y}|} \sum_{y\in\mathcal{Y}}(\bar{\vec{h}}_{y}^{(\ell)} - \bar{\vec{h}}^{(\ell)}) (\bar{\vec{h}}_{y}^{(\ell)} - \bar{\vec{h}}^{(\ell)})^{\top} 
\end{align*}
\end{subequations}
Then the within-class variability $\vec{\Sigma}_{\text{w}}^{(\ell)}$ will also be dominated by the group that has the largest number of samples. 
However, our current definition in \cref{sec:var} will scale the $\sum_{\vec{h}} (\vec{h}-\bar{\vec{h}}_y)(\vec{h}-\bar{\vec{h}}_y))^{\top}$ term by $|\mathcal{G}_y^{(\ell)}|$ before taking the outer sum $\sum_{y}$, thus being more robust to data imbalance.

\begin{figure}[t]
	\begin{center}
	    \begin{subfigure}[t]{0.96\linewidth}
			\includegraphics[width=0.95\linewidth]{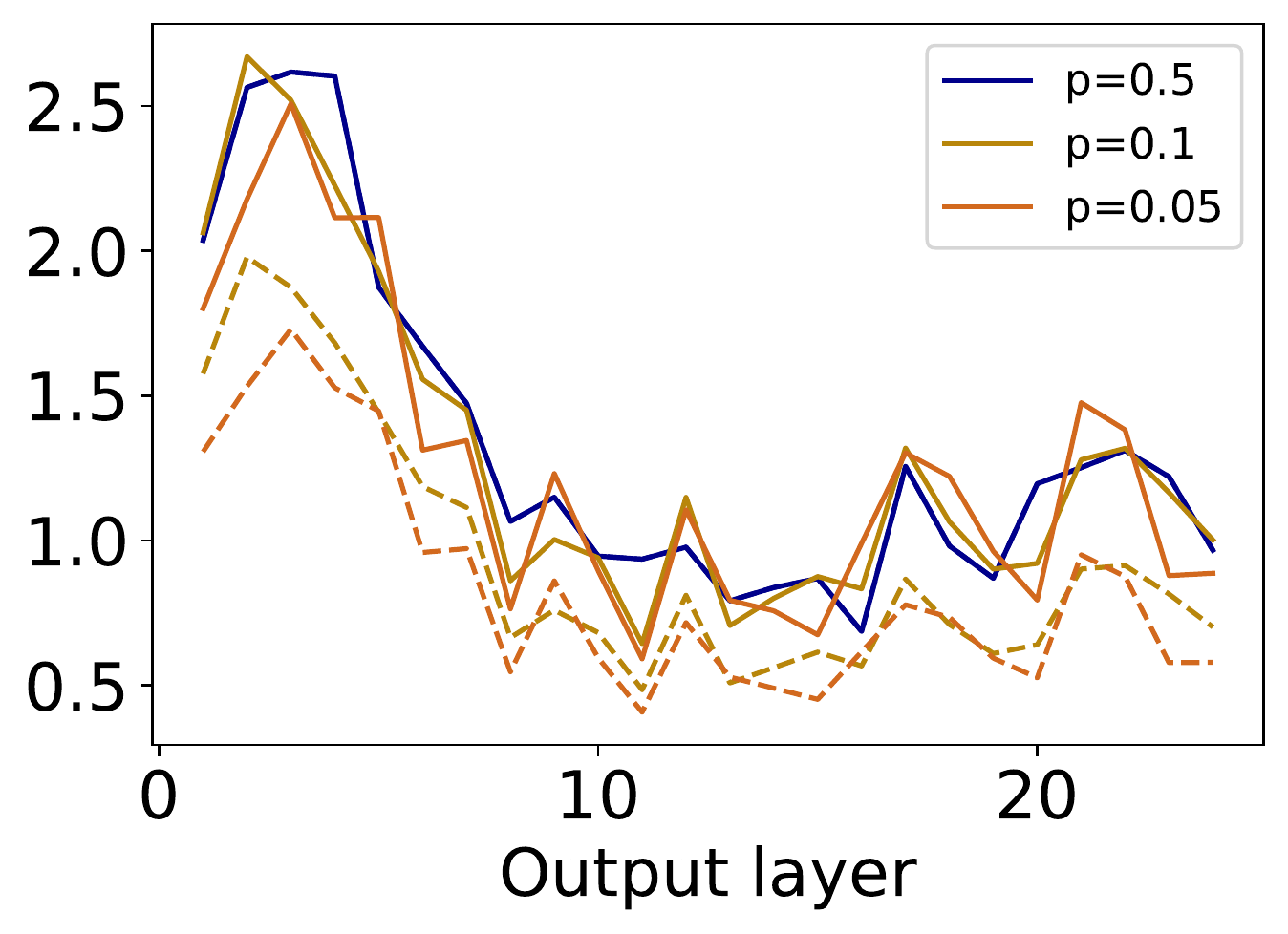}
		\end{subfigure}
	\caption{The ${\nu}^{(\ell)}$ metric computed using our formulas (solid lines) and the formulas of \citet{papyan2020prevalence} (dashed lines) in data-imbalanced experiments.}\label{fig:sensitivity_NC}
	\end{center}
\end{figure}

To verify our intuition, we conducted a series of data-imbalanced experiments on SST-2 like we did in \cref{sec:ablation}.
In each experiment, we randomly sampled $N=20000$ training examples with a portion $p$ from the negative class where $p \in \{ 0.5, 0.1, 0.05 \}$. 
Then we constructed the $\nu$ curves and plot them in \cref{fig:sensitivity_NC}: solid curves use our math formulas in \cref{sec:var} while dashed lines use the formulas that strictly follow \citet{papyan2020prevalence}. 
As we can see, when data is balanced, the solid and dashed lines are exactly the same. 
When data is imbalanced, the solid lines still stay close while the dashed lines move apart. 
This figure illustrates that our formulas are more robust to data imbalance. 

\section{Experiment Details}\label{app:exp}
For the pretrained language model, we use the implementation and pretrained weights (roberta-large with 24 layers and 355M parameters) of Huggingface~\citep{wolf2020huggingface}.%
We follow previous work~\citep{wang-etal-2018-glue} and use different evaluation methods for different tasks:
\begin{itemize}[leftmargin=*]
    \item On CoLA, we use Matthews correlation coefficient. 
    \item On MRPC and QQP, we use F1 score. 
    \item On the others, we use classification accuracy.
\end{itemize}

\subsection{Training Details}\label{app:training}

We only tune learning rate for each task and strategy and specific tuning methods.
The number of epochs is 10 for full-fine-tuning on MNLI and QQP and 20 for all the other experiments.
All the other hyperparameters are the same for all the experiments.
We use the AdamW optimizer with a linear learning rate scheduler with 6\% warm-up steps. 
We set the dropout rate to be 0.1. %
The weight decay is 0. %
The batch size is 8. %
We evaluate on the validation set per epoch and report the best result. 
We run the experiments on Nvidia RTX A4000 and GeForce RTX 2080 Ti.
We use the standard splits and the datasets can be downloaded at \url{https://huggingface.co/datasets/glue}.

\subsection{Implementation Details}\label{app:code}

Our code is implemented in PyTorch~\citep{paszke-17-pytorch} and heavily relies on HuggingFace.
It will be released after the paper is published.

For all the experiments that requires a new task-specific classification head, we relied on the original implementation in RoBERTa of Huggingface~\citep{wolf2020huggingface}\footnote{\url{https://github.com/huggingface/transformers/blob/main/src/transformers/models/roberta/modeling_roberta.py}}.

For adapter-tuning, we implemented the earliest adapter architecture designed by \citet{houlsby2019parameter} and relied on the public implementation\footnote{\url{https://github.com/jxhe/unify-parameter-efficient-tuning}} provided by \citet{he2021towards}.
The bottleneck dimension is 256 and the adapter uses the same initialization as BERT~\citep{devlin-18-bert}.
The trainable parameters to update are the parameters of the inserted adapters, the layer normalization parameters in the selected layers, and the parameters of the classification head.

\subsection{Probing Experiments}\label{app:probing}
In \cref{sec:exp}, we probed both the pretrained and fine-tuned models. 
On each GLUE task, we used a fixed learning rate in \cref{tab:lr_probing} to train a classification head for each layer. 
\begin{table}[ht]\centering
\begin{tabular}{cc}\hline
Task & Learning rate \\\hline
CoLA &1e-3 \\
MNLI &1e-3 \\
MRPC &1e-3 \\
QNLI &1e-3 \\
QQP &3e-4 \\
SST-2 &3e-3 \\
\hline
\end{tabular}
\caption{Learning rate for probing experiments.}\label{tab:lr_probing}
\end{table}

\subsection{Fine-Tuning Experiments}\label{app:finetune}
For the fine-tuning experiments in \cref{sec:progressive}, we only tuned the learning rate. 
Ideally, we should have swept a large range of learning rates for all the strategies and found the best learning rate for each strategy; but that would require too much computation cost that we couldn't afford. 
Our preliminary experiments showed that small learning rates tend to work better when the number of trainable parameters is large and that large learning rates tend to work better when the number of trainable parameters is small. 
Therefore, we set a different range of learning rates for each different strategy based on their numbers of trainable parameters. 
The ranges that we used are in \cref{tab:lr_finetune}. 
For each strategy, on each task, we chose the best learning rate based on the performance on the held-out validation set. 
The $(\ell_{\text{bottom}}, \ell^*, \ell^*)$ and $(\ell_{\text{bottom}}, \ell^*, L)$ strategies use the same learning rate as the conventional $(\ell_{\text{bottom}}, L, L)$ strategy.%
\begin{table}[ht]\centering
\begin{tabular}{cc}\hline
Strategy & Learning rate set\\\hline
full fine-tuning &1e-6, 5e-6, 8e-6, 1e-5, 2e-5  \\
$(\ell_{\text{bottom}}, L, L)$ & 8e-6, 1e-5, 2e-5, 3e-5, 5e-5 \\
$(1,\ell^*, \ell^*)$ & 5e-6, 1e-5, 5e-5, 1e-4 \\
$(\ell^*+1, L, L)$ & 1e-6, 3e-6, 1e-5, 3e-5 \\
\hline
\end{tabular}
\caption{Learning rate for fine-tuning experiments.}\label{tab:lr_finetune}
\end{table}

\subsection{Adapter-Tuning Experiments}\label{app:adapter}
For the adapter-tuning experiments in \cref{sec:progressive}, we only tuned the learning rate. 
For the same reason as we discussed in \cref{app:finetune}, we set a different range of learning rates for each different strategy based on their numbers of trainable parameters. 
The ranges that we used are in \cref{tab:lr_adapter}.
Again, the $(\ell_{\text{bottom}}, \ell^*, \ell^*)$ and $(\ell_{\text{bottom}}, \ell^*, L)$ strategies use the same learning rate as the conventional $(\ell_{\text{bottom}}, L, L)$ strategy.
\begin{table}[ht]\centering
\begin{tabular}{cc}\hline
Strategy & Learning rate set\\\hline
full adapter & 1e-5, 3e-5, 1e-4  \\
$(\ell_{\text{bottom}}, L, L)$ & 3e-5, 1e-4, 3e-4 \\
$(1,\ell^*, \ell^*)$ & 1e-5, 3e-5, 1e-4 \\
$(\ell^*+1, L, L)$ & 1e-5, 3e-5, 1e-4 \\
\hline
\end{tabular}
\caption{Learning rate for adapter-tuning experiments.}\label{tab:lr_adapter}
\end{table}

\section{More Results} \label{app:results}

\subsection{Detailed numbers of RoBERTa experiments}\label{app:roberta_numbers} \label{app:roberta_middle} \label{app:roberta_cost}
As mentioned in \cref{sec:finetune} and \cref{sec:adapter}, the mean values and standard errors of finetuning and adapter-tuning RoBERTa with different strategies are listed in \cref{tab:roberta_finetune} and \cref{tab:roberta_adapter}. 
The standard error of our strategies' performance is not significantly higher or lower than the baselines. 
So our strategies can't help solve the stability issue in fine-tuning PLMs. %

\begin{table*}[!htp]\centering
\scriptsize
\begin{tabular}{lrrrrrrrrrrrrr}\toprule
Strategy $\|$ \# of tuned layers &\multicolumn{2}{c}{CoLA} &\multicolumn{2}{c}{MNLI} &\multicolumn{2}{c}{MRPC} &\multicolumn{2}{c}{QNLI} &\multicolumn{2}{c}{QQP} &\multicolumn{2}{c}{SST-2} \\\cmidrule{1-13}
&mean &std &mean &std &mean &std &mean &std &mean &std &mean &std \\\midrule
$(\ell^*, \ell^*, \ell^*) \| 1$ &0.565 &0.0198 &0.845 &0.0024 &0.909 &0.0054 &0.896 &0.0031 &0.869 &0.0007 &0.952 &0.0021 \\
$(\ell^*, \ell^*, L) \| 1$ &0.619 &0.0173 &0.891 &0.0011 &0.925 &0.0072 &0.934 &0.0031 &0.885 &0.0013 &0.960 &0.0031 \\
$(L, L, L) \| 1$ &0.598 &0.0201 &0.820 &0.0021 &0.858 &0.0086 &0.866 &0.0041 &0.859 &0.0008 &0.938 &0.0032 \\
$(\ell^*-1, \ell^*, \ell^*) \| 2$ &0.594 &0.0176 &0.861 &0.0018 &0.920 &0.0091 &0.917 &0.0019 &0.885 &0.0011 &0.954 &0.0057 \\
$(\ell^*-1,\ell^*,L) \| 2$ &0.612 &0.0123 &0.892 &0.0021 &0.928 &0.0040 &0.926 &0.0009 &0.889 &0.0018 &0.957 &0.0031 \\
$(L-1,L,L) \| 2$ &0.616 &0.0070 &0.841 &0.0007 &0.881 &0.0052 &0.902 &0.0014 &0.875 &0.0012 &0.945 &0.0034 \\
$(\ell^*-2, \ell^*, \ell^*) \| 3$ &0.615 &0.0166 &0.870 &0.0018 &0.927 &0.0074 &0.927 &0.0020 &0.887 &0.0006 &0.953 &0.0027 \\
$(\ell^*-2,\ell^*,L) \| 3$ &0.611 &0.0138 &0.892 &0.0019 &0.930 &0.0035 &0.928 &0.0014 &0.888 &0.0017 &0.954 &0.0017 \\
$(L-2,L,L) \| 3$ &0.625 &0.0068 &0.847 &0.0042 &0.895 &0.0049 &0.908 &0.0014 &0.878 &0.0019 &0.944 &0.0025 \\
$(\ell^*+1,L,L) \| L - \ell^*$ &0.623 &0.0080 &0.897 &0.0015 &0.931 &0.0024 &0.942 &0.0014 &0.890 &0.0018 &0.959 &0.0022 \\
$(1,\ell^*,\ell^*) \| \ell^*$ &0.674 &0.0096 &0.886 &0.0018 &0.929 &0.0033 &0.933 &0.0009 &0.895 &0.0018 &0.957 &0.0017 \\
$(1,L,L) \| L$ &0.699 &0.0131 &0.906 &0.0007 &0.934 &0.0017 &0.948 &0.0014 &0.896 &0.0017 &0.965 &0.0012 \\
\bottomrule
\end{tabular}
\caption{Results of finetuning RoBERTa with different stratigies}\label{tab:roberta_finetune}
\end{table*}

\begin{table*}[!htp]\centering
\scriptsize
\begin{tabular}{lrrrrrrrrrrrrr}\toprule
Strategy $\|$ \# of tuned layers &\multicolumn{2}{c}{CoLA} &\multicolumn{2}{c}{MNLI} &\multicolumn{2}{c}{MRPC} &\multicolumn{2}{c}{QNLI} &\multicolumn{2}{c}{QQP} &\multicolumn{2}{c}{SST-2} \\\cmidrule{1-13}
&mean &std &mean &std &mean &std &mean &std &mean &std &mean &std \\\midrule
$(\ell^*, \ell^*, \ell^*) \| 1$ &0.420 &0.0086 &0.652 &0.0016 &0.848 &0.0016 &0.788 &0.0013 &0.758 &0.0022 &0.914 &0.0015 \\
$(\ell^*, \ell^*, L) \| 1$ &0.600 &0.0105 &0.890 &0.0012 &0.911 &0.0059 &0.937 &0.0020 &0.881 &0.0008 &0.959 &0.0012 \\
$(L, L, L) \| 1$ &0.415 &0.0069 &0.570 &0.0054 &0.839 &0.0013 &0.700 &0.0048 &0.787 &0.0045 &0.913 &0.0031 \\
$(\ell^*-1, \ell^*, \ell^*) \| 2$ &0.578 &0.0047 &0.850 &0.0015 &0.909 &0.0024 &0.902 &0.0016 &0.874 &0.0011 &0.956 &0.0019 \\
$(\ell^*-1,\ell^*,L) \| 2$ &0.597 &0.0119 &0.895 &0.0013 &0.922 &0.0054 &0.938 &0.0006 &0.885 &0.0014 &0.954 &0.0026 \\
$(L-1,L,L) \| 2$ &0.583 &0.0194 &0.816 &0.0029 &0.860 &0.0035 &0.857 &0.0022 &0.859 &0.0025 &0.938 &0.0047 \\
$(\ell^*-2, \ell^*, \ell^*) \| 3$ &0.576 &0.0168 &0.866 &0.0007 &0.918 &0.0058 &0.920 &0.0004 &0.881 &0.0009 &0.956 &0.0029 \\
$(\ell^*-2,\ell^*,L) \| 3$ &0.588 &0.0108 &0.898 &0.0006 &0.922 &0.0068 &0.934 &0.0038 &0.887 &0.0018 &0.954 &0.0025 \\
$(L-2,L,L) \| 3$ &0.620 &0.0229 &0.839 &0.0019 &0.867 &0.0069 &0.886 &0.0020 &0.867 &0.0030 &0.945 &0.0037 \\
$(\ell^*+1,L,L) \| L - \ell^*$ &0.619 &0.0198 &0.894 &0.0017 &0.931 &0.0081 &0.942 &0.0011 &0.886 &0.0032 &0.962 &0.0020 \\
$(1,\ell^*,\ell^*) \| \ell^*$ &0.671 &0.0093 &0.887 &0.0019 &0.932 &0.0080 &0.934 &0.0011 &0.892 &0.0023 &0.960 &0.0015 \\
$(1,L,L) \| L$ &0.653 &0.0510 &0.908 &0.0017 &0.930 &0.0015 &0.948 &0.0021 &0.897 &0.0010 &0.964 &0.0006 \\
\bottomrule
\end{tabular}
\caption{Results of adapter-tuning RoBERTa with different strategies}\label{tab:roberta_adapter}
\end{table*}

As discussed in \cref{sec:finetune} and \cref{sec:adapter}, we also fine-tuned and adapter-tuned PLM with the middle layer baseline on CoLA and SST-2 and listed the results in \cref{tab:middle_finetune_roberta} and \cref{tab:middle_adapter_roberta}. 

\begin{table}[!htp]\centering
\scriptsize
\begin{tabular}{lrrrrr}\toprule
Strategy $\|$ \# of tuned layers &\multicolumn{2}{c}{CoLA } &\multicolumn{2}{c}{SST-2 } \\\cmidrule{1-5}
&mean &std &mean &std \\\midrule
$(\ell^*, \ell^*, \ell^*) \| 1$ &0.565 &0.0198 &0.952 &0.0021 \\
$(\ell^*, \ell^*, L) \| 1$ &\textbf{0.619} &0.0173 &\textbf{0.960} &0.0031 \\
$(\ell_{\text{mid}}, \ell_{\text{mid}},\ell_{\text{mid}}) \| 1$ &0.578 &0.0051 &0.936 &0.0017 \\
$(\ell_{\text{mid}}, \ell_{\text{mid}},L \| 1$ &0.607 &0.0087 &0.957 &0.0043 \\
$(\ell^*-1, \ell^*, \ell^*) \| 2$ &0.594 &0.0176 &0.954 &0.0057 \\
$(\ell^*-1,\ell^*,L) \| 2$ &\textbf{0.612} &0.0123 &\textbf{0.957} &0.0031 \\
$(\ell_{\text{mid}}-1, \ell_{\text{mid}},\ell_{\text{mid}}) \| 2$ &0.599 &0.0075 &0.948 &0.0024 \\
$(\ell_{\text{mid}}-1, \ell_{\text{mid}},L) \| 2$ &0.611 &0.0193 &0.953 &0.0014 \\
$(\ell^*-2, \ell^*, \ell^*) \| 3$ &\textbf{0.615} &0.0166 &0.953 &0.0027 \\
$(\ell^*-2,\ell^*,L) \| 3$ &0.611 &0.0138 &\textbf{0.954} &0.0017 \\
$(\ell_{\text{mid}}-2, \ell_{\text{mid}},\ell_{\text{mid}}) \| 3$ &0.612 &0.0122 &0.949 &0.0010 \\
$(\ell_{\text{mid}}-2, \ell_{\text{mid}},L) \| 3$ &0.611 &0.0170 &0.952 &0.0021 \\
\bottomrule
\end{tabular}
\caption{Comparison of fine-tuning with middle layer baseline on CoLA and SST-2}\label{tab:middle_finetune_roberta}
\end{table}

\begin{table}[!htp]\centering
\scriptsize
\begin{tabular}{lrrrrr}\toprule
Strategy $\|$ \# of adapted layers &\multicolumn{2}{c}{CoLA} &\multicolumn{2}{c}{SST-2} \\\cmidrule{1-5}
&mean &std &mean &std \\\midrule
$(\ell^*, \ell^*, \ell^*) \| 1$ &0.420 &0.0086 &0.914 &0.0015 \\
$(\ell^*, \ell^*, L) \| 1$ &\textbf{0.600} &0.0105 &\textbf{0.959} &0.0012 \\
$(\ell_{\text{mid}}, \ell_{\text{mid}},\ell_{\text{mid}}) \| 1$ &0.383 &0.0170 &0.884 &0.0035 \\
$(\ell_{\text{mid}}, \ell_{\text{mid}},L \| 1$ &0.594 &0.0028 &0.956 &0.0022 \\
$(\ell^*-1, \ell^*, \ell^*) \| 2$ &0.578 &0.0047 &\textbf{0.956} &0.0019 \\
$(\ell^*-1,\ell^*,L) \| 2$ &0.597 &0.0119 &0.954 &0.0026 \\
$(\ell_{\text{mid}}-1, \ell_{\text{mid}},\ell_{\text{mid}}) \| 2$ &0.586 &0.0095 &0.946 &0.0026 \\
$(\ell_{\text{mid}}-1, \ell_{\text{mid}},L) \| 2$ &\textbf{0.607} &0.0108 &0.955 &0.0040 \\
$(\ell^*-2, \ell^*, \ell^*) \| 3$ &0.576 &0.0168 &0.956 &0.0029 \\
$(\ell^*-2,\ell^*,L) \| 3$ &0.588 &0.0108 &0.954 &0.0025 \\
$(\ell_{\text{mid}}-2, \ell_{\text{mid}},\ell_{\text{mid}}) \| 3$ &0.601 &0.0061 &0.950 &0.0045 \\
$(\ell_{\text{mid}}-2, \ell_{\text{mid}},L) \| 3$ &\textbf{0.621} &0.0085 &\textbf{0.957} &0.0019 \\
\bottomrule
\end{tabular}
\caption{Comparison of adapter-tuning with middle layer baseline on CoLA and SST-2}\label{tab:middle_adapter_roberta}
\end{table}

 As discussed in \cref{sec:finetune}, we compare the computation cost and storage cost of some strategies on MNLI in \cref{tab:compare_finetune_cost}. 
When only keeping $\ell^* = 14$ layers in the PLM, it reduces inference cost and number of parameters by 40\%.
In general, using our method will reduce the computation though the actual saving depends on the implementation and the devices; see \cref{tab:compare_finetune_cost} for details of our experiments. 
For example, when using the $\ell*$-up strategies, the most optimized implementation would cache the output of the bottom layers and reuse them, which will further reduce the training and inference cost. But we haven't implemented it yet. So there is still plenty of room to improve the efficiency over \cref{tab:compare_finetune_cost} guided by our experimental insights.
\begin{table*}[ht]\centering
\begin{tabular}{lccccc}\hline
Strategy &Training time &Inference time &Total params &Trainable params \\\hline
full fine-tuning &2h30min &50s &355362819 &355362819 \\
$(1, \ell^*, \ell^*)$ &1h40min &31s &229400579 &177399811 \\
$(\ell^*+1, L, L)$ &1h30min &50s &355362819 &127014915 \\
\hline
\end{tabular}
\caption{Computation cost per epoch for RoBERTa-large fine-tuning experiments on MNLI.}\label{tab:compare_finetune_cost}
\end{table*}

\subsection{Experiments on DeBERTa}\label{app:deberta}
As discussed in \cref{sec:exp}, we also conducted experiements on DeBERTa-base. 

\begin{figure}[t]
	\begin{center}
	    \begin{subfigure}[t]{0.48\linewidth}
			\includegraphics[width=0.99\linewidth]{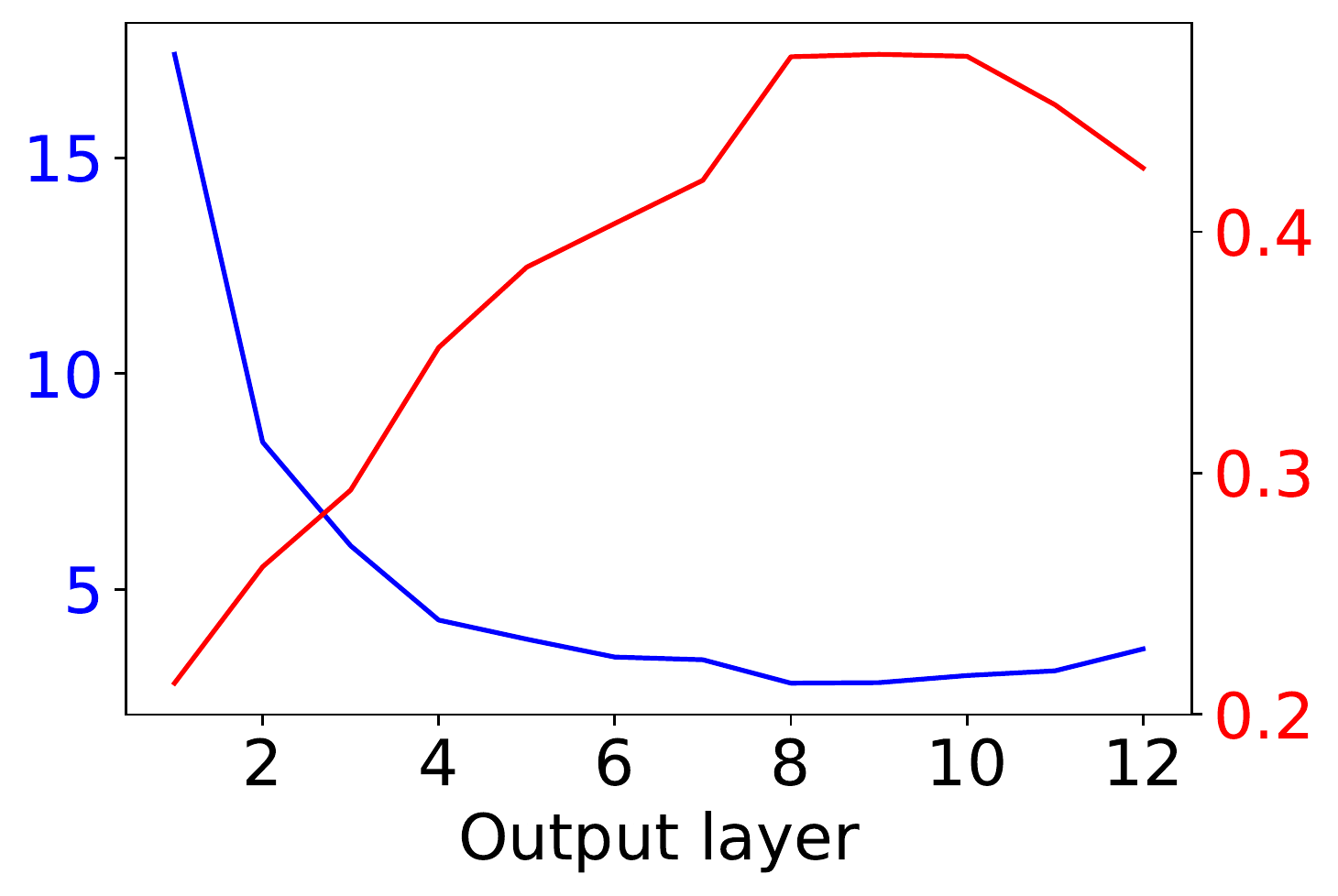}
			\caption{CoLA}
		\end{subfigure}
		~
		\begin{subfigure}[t]{0.48\linewidth}
			\includegraphics[width=0.99\linewidth]{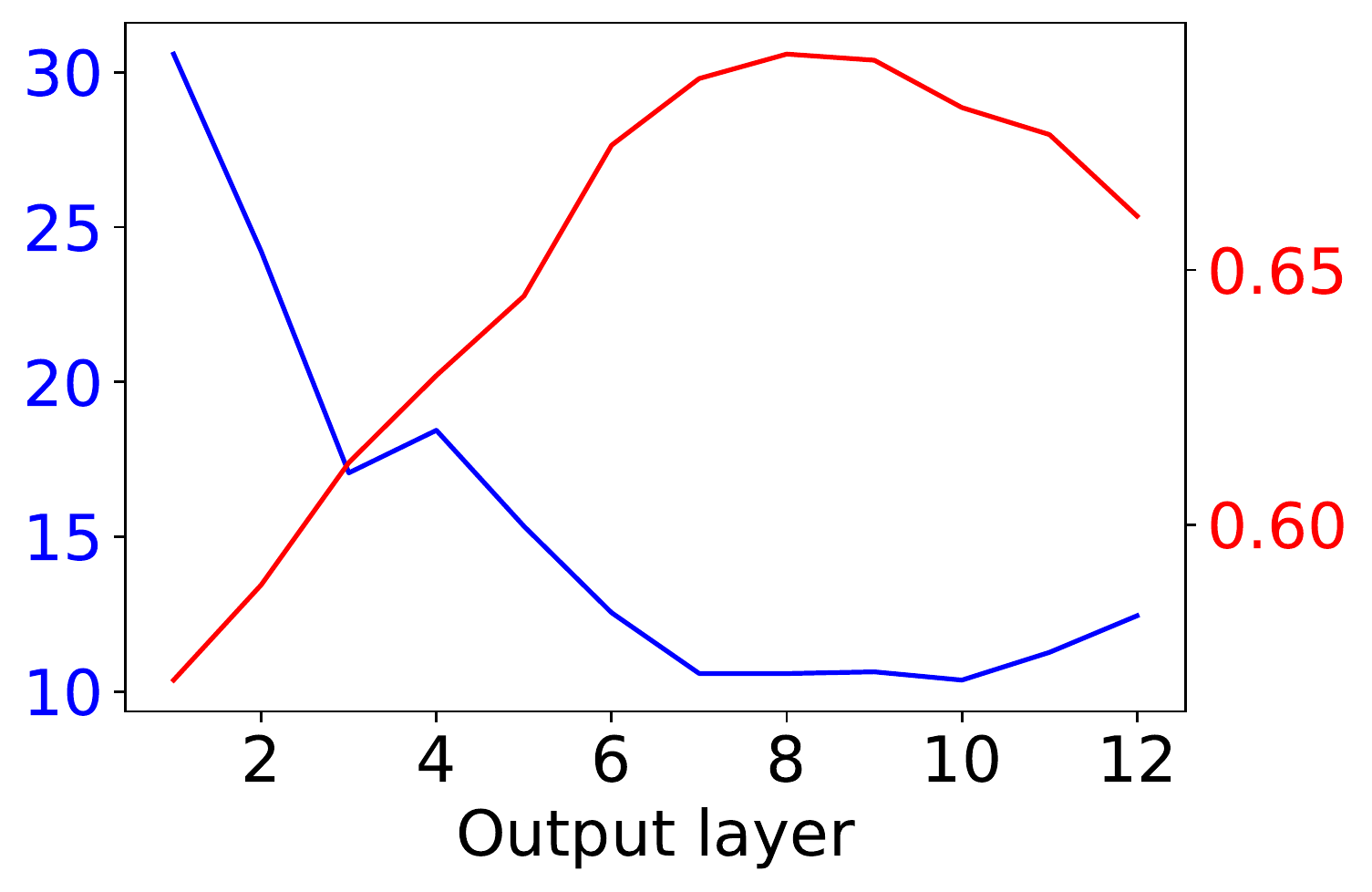}
			\caption{MNLI}
		\end{subfigure}
		
		\begin{subfigure}[t]{0.48\linewidth}
			\includegraphics[width=0.99\linewidth]{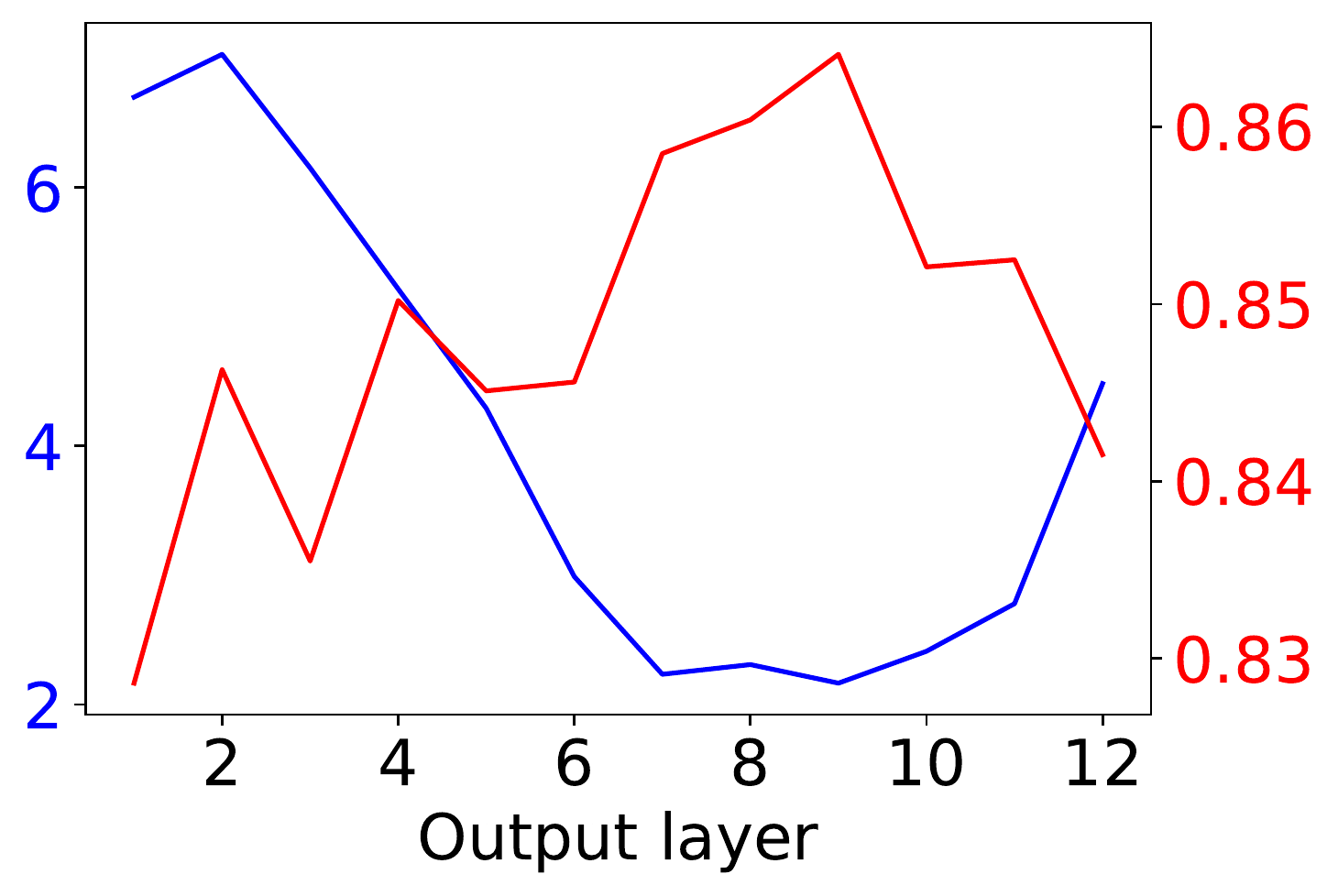}
			\caption{MRPC}
		\end{subfigure}
		~
		\begin{subfigure}[t]{0.48\linewidth}
			\includegraphics[width=0.99\linewidth]{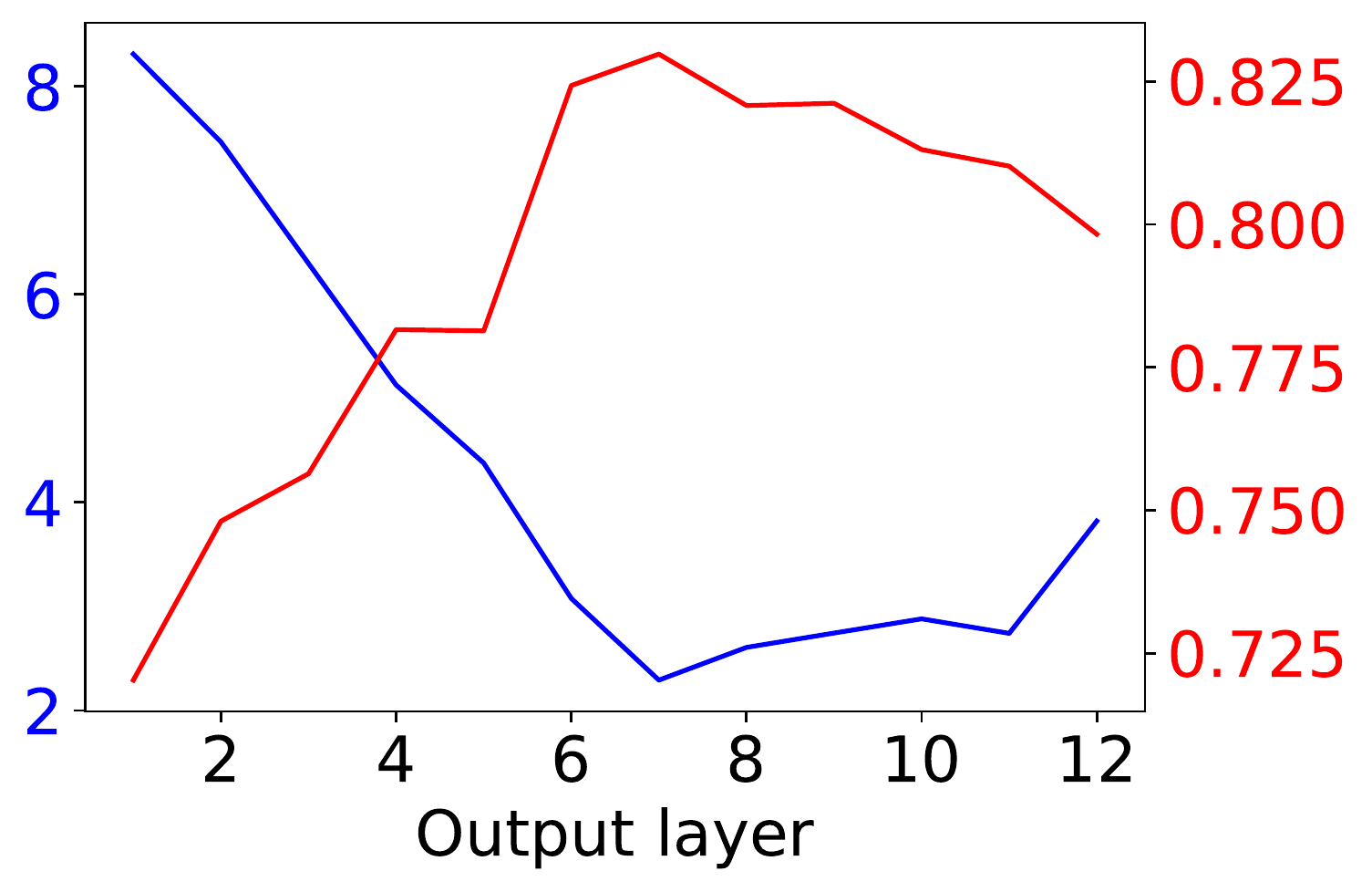}
			\caption{QNLI}
		\end{subfigure}
		
		\begin{subfigure}[t]{0.48\linewidth}
			\includegraphics[width=0.99\linewidth]{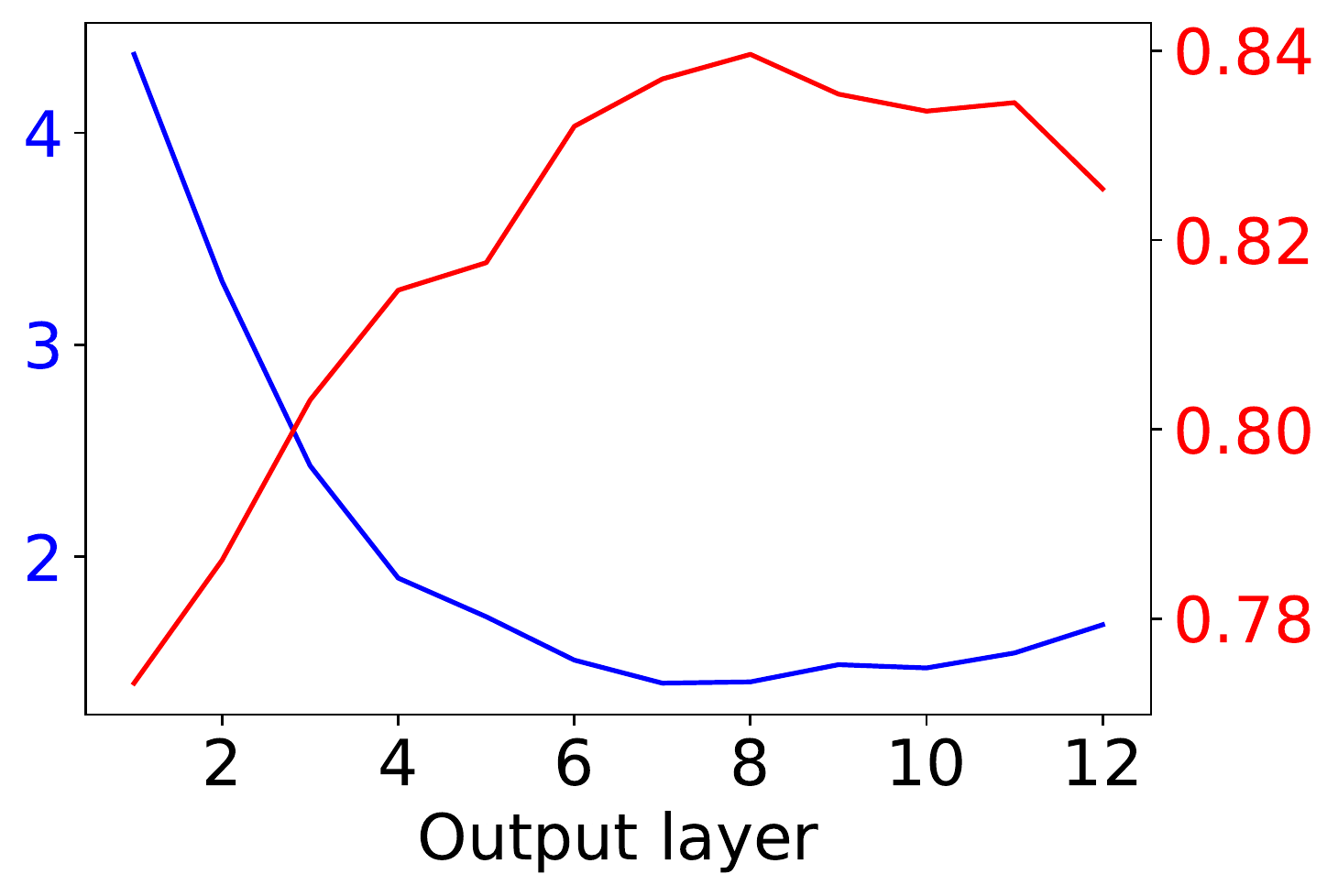}
			\caption{QQP}
		\end{subfigure}
		~
		\begin{subfigure}[t]{0.48\linewidth}
			\includegraphics[width=0.99\linewidth]{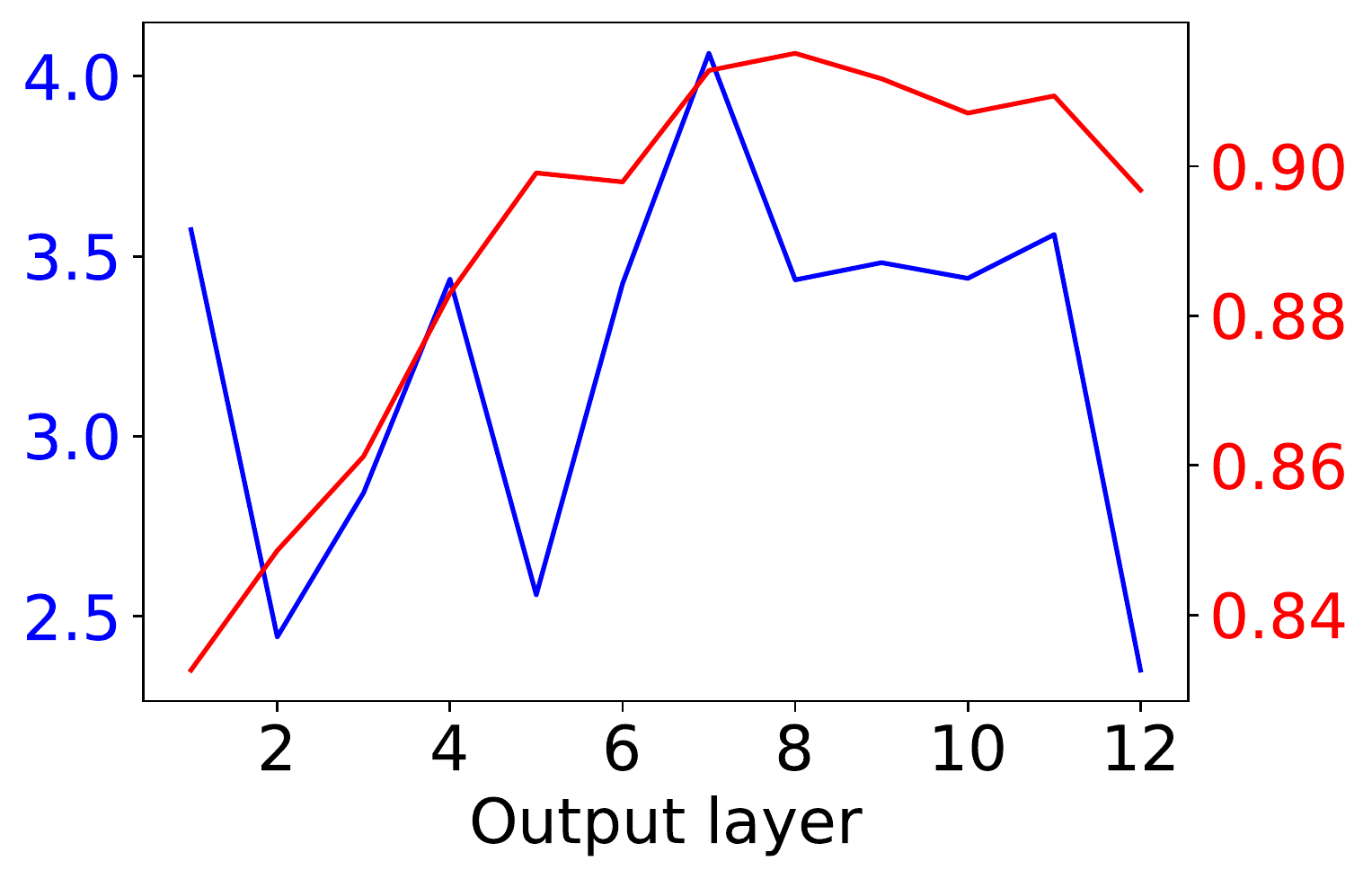}
			\caption{SST-2}
		\end{subfigure}
		\vspace{-8pt}
		\caption{The task-specialty metric (blue) and probing performance (red) of each layer of a pretrained DeBERTa model. Each figure is a GLUE task.}\label{fig:nc_pretrain_deberta}
	\end{center}
	\vspace{-6pt}
\end{figure}

\begin{figure}[t]
	\begin{center}
	    \begin{subfigure}[t]{0.48\linewidth}
		\includegraphics[width=0.95\linewidth]{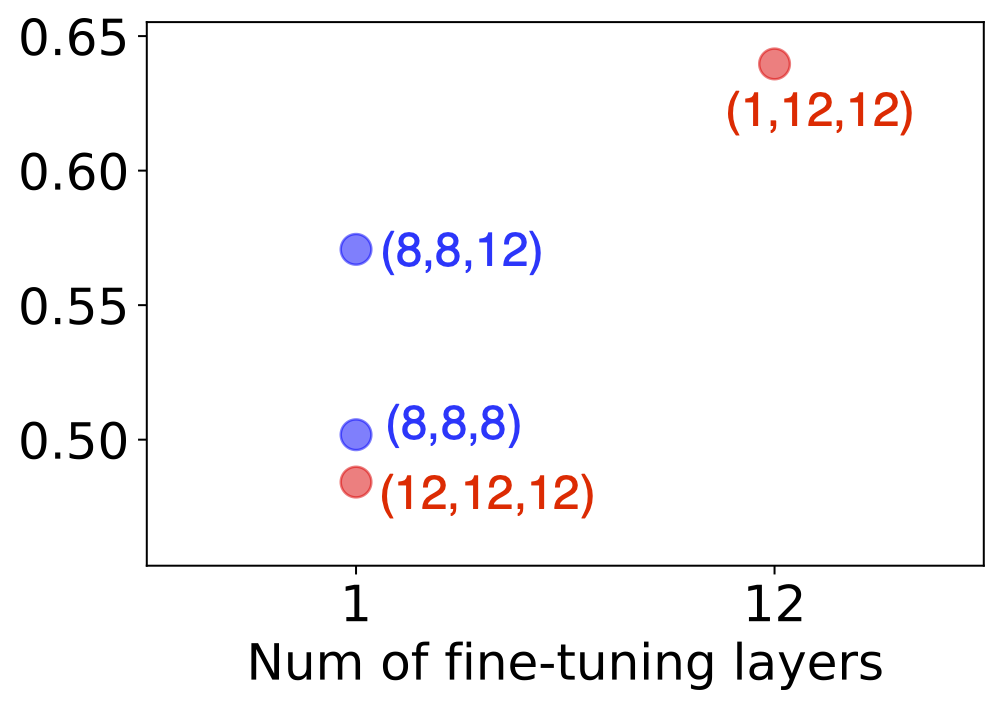}
			\vspace{-4pt}
			\caption{CoLA}
		\end{subfigure}
		~
		\begin{subfigure}[t]{0.48\linewidth}
			\includegraphics[width=0.95\linewidth]{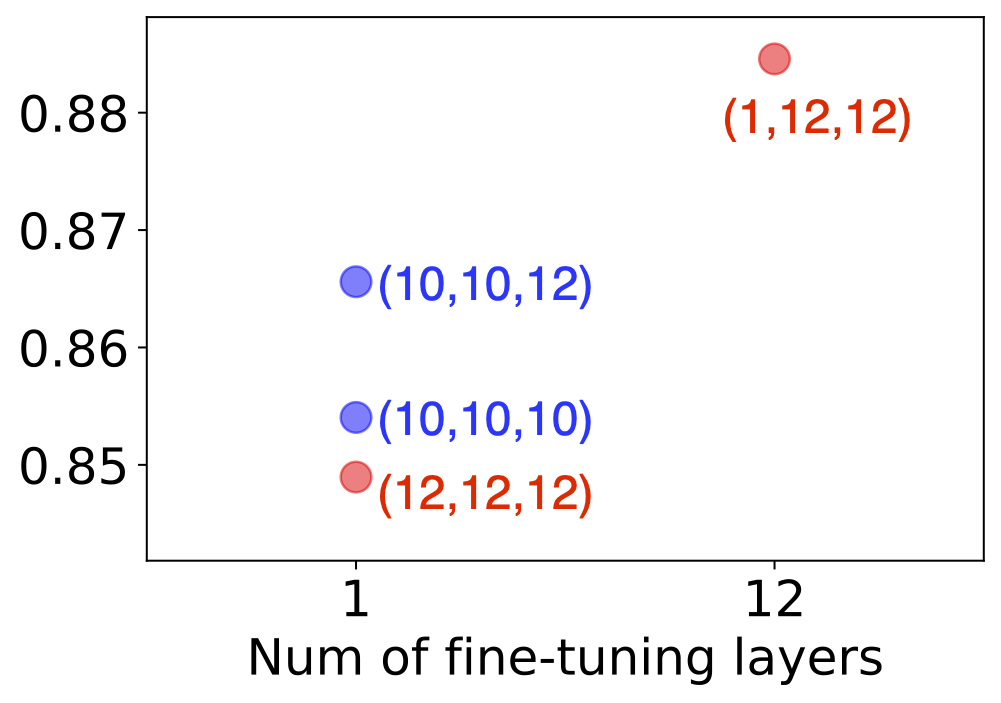}
			\vspace{-4pt}
			\caption{MNLI}
		\end{subfigure}
		\begin{subfigure}[t]{0.48\linewidth}
			\includegraphics[width=0.95\linewidth]{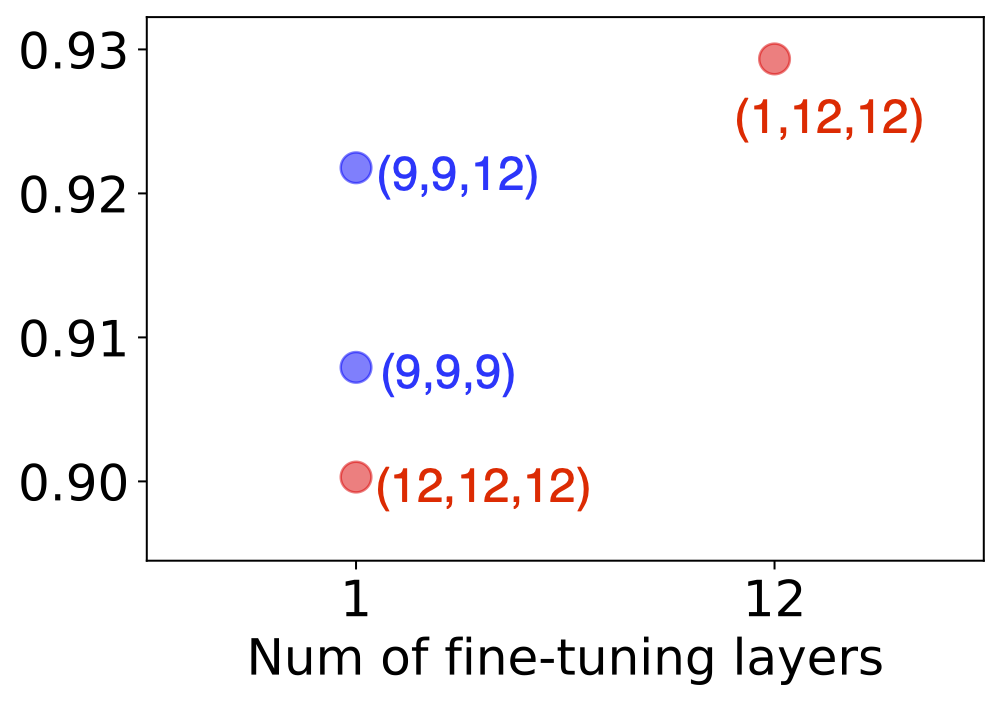}
			\vspace{-4pt}
			\caption{MRPC}
		\end{subfigure}
		~
		\begin{subfigure}[t]{0.48\linewidth}
			\includegraphics[width=0.95\linewidth]{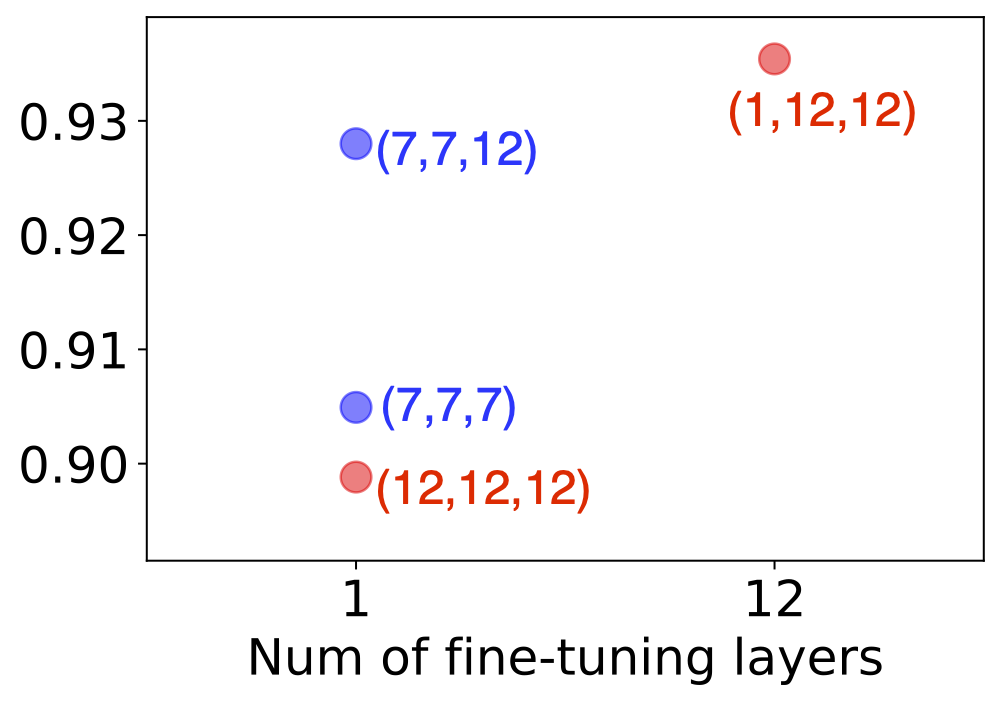}
			\vspace{-4pt}
			\caption{QNLI}
		\end{subfigure}
		\begin{subfigure}[t]{0.48\linewidth}
			\includegraphics[width=0.95\linewidth]{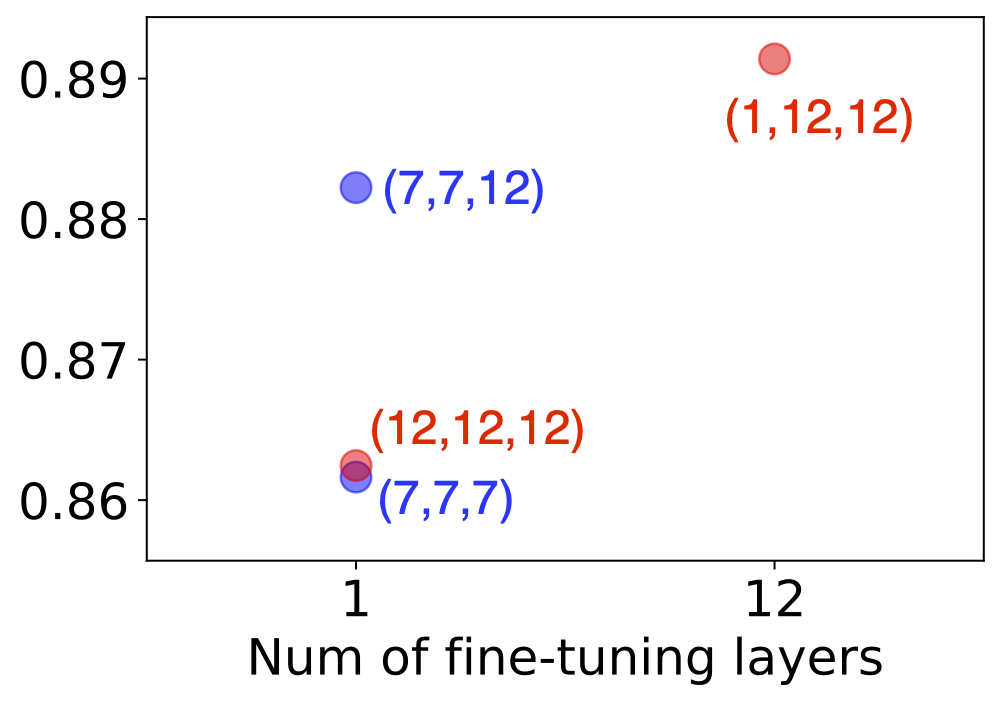}
			\vspace{-4pt}
			\caption{QQP}
		\end{subfigure}
		~
		\vspace{-6pt}
		\caption{%
		Task performance vs.\@ the number of selected layers for fine-tuning DeBERTa-base. 
		The annotation of each dot is its strategy identifier $(\ell_{\text{bottom}},\ell_{\text{top}},\ell_{\text{head}})$.
		}\label{fig:progressive_finetune_deberta}
	\end{center}
	\vspace{-6pt}
\end{figure}

We computed the task-specialty metric $\nu$ for all the 12 layers and plotted with each layer's probing performance in \cref{fig:nc_pretrain_deberta} as we did in \cref{fig:nc_pretrain}. 
For all the tasks except SST-2, we can observe the same pattern as in RoBERTa: the layers with low $\nu$ tend to have high probing performance. 
On SST-2, the task-specialty metric isn't negatively correlated with the probing performance. 
This might be an example mentioned in \cref{sec:limit} that the layers with high $\nu$ can also achieve good performance.
Because the best layer $\ell^*$ selected by the metric is already the last layer for SST-2, we implemented our strategies $(\ell^*, \ell^*, \ell^*)$, $(\ell^*, \ell^*, L)$ on all the tasks except SST-2. We compared them with baseline $(L,L,L)$ and full fine-tuning to see whether the metric can help make fine-tuning more efficiently. 
The results are plotted in \cref{fig:progressive_finetune_deberta} and listed in \cref{tab:deberta_finetune}. 
When only fine-tuning $1$ layer, our strategy $(\ell^*, \ell^*, L)$ always achieves the best performance and the baseline $(L,L,L)$ is always the worst. 
On MRPC, QNLI and QQP, the performance of $(\ell^*, \ell^*, L)$ is even close to the performance of full fine-tuning with fewer than 10\% tuning parameters. 

\begin{table*}[!htp]\centering
\scriptsize
\begin{tabular}{lrrrrrrrrrrr}\toprule
Strategy $\|$ \# tuned layers &\multicolumn{2}{c}{CoLA ($\ell^* = 8$)} &\multicolumn{2}{c}{MNLI ($\ell^* = 10$)} &\multicolumn{2}{c}{MRPC ($\ell^* = 9$)} &\multicolumn{2}{c}{QNLI ($\ell^* = 7$)} &\multicolumn{2}{c}{QQP ($\ell^* = 7$)} \\\cmidrule{1-11}
&mean &std &mean &std &mean &std &mean &std &mean &std \\\midrule
$(\ell^*, \ell^*, \ell^*) \| 1$ &0.502 &0.0137 &0.854 &0.0008 &0.908 &0.0083 &0.905 &0.0023 &0.862 &0.0004 \\
$(\ell^*, \ell^*, L) \| 1$ &0.571 &0.0111 &0.866 &0.0027 &0.922 &0.0065 &0.928 &0.0011 &0.882 &0.0014 \\
$(L, L, L) \| 1$ &0.484 &0.0122 &0.849 &0.0008 &0.900 &0.0057 &0.899 &0.0019 &0.862 &0.0015 \\
$(1,L,L) \| L$ &0.640 &0.0096 &0.885 &0.0013 &0.929 &0.0038 &0.935 &0.0023 &0.891 &0.0007 \\
\bottomrule
\end{tabular}
\caption{Results of fine-tuning DeBERTa with different strategies}\label{tab:deberta_finetune}
\end{table*}

We also compared with middle layer baseline on MNLI. 
The results are listed in \cref{tab:middle_finetune_deberta}. 
$\ell_{\text{mid}}$ works better than $\ell^*$ this time. 

\begin{table}[!htp]\centering
\scriptsize
\begin{tabular}{lrrr}\toprule
Strategy $\|$ \# of tuned layers &\multicolumn{2}{c}{MNLI} \\\cmidrule{1-3}
&mean &std \\\midrule
$(\ell^*, \ell^*, \ell^*) \| 1$ &0.854 &0.0008 \\
$(\ell^*, \ell^*, L) \| 1$ &0.866 &0.0027 \\
$(\ell_{\text{mid}}, \ell_{\text{mid}},\ell_{\text{mid}}) \| 1$ &0.849 &0.0009 \\
$(\ell_{\text{mid}}, \ell_{\text{mid}},L \| 1$ &\textbf{0.872} &0.0022 \\
\bottomrule
\end{tabular}
\caption{Comparison of fine-tuning DeBERTa-base with middle layer baseline on MNLI}\label{tab:middle_finetune_deberta}
\end{table}

\subsection{Task-Specialty vs.\@ Probing Performance for Pretrained Models}\label{app:nc_pretrain}
\begin{figure}[t]
	\begin{center}
	    \begin{subfigure}[t]{0.48\linewidth}
			\includegraphics[width=0.99\linewidth]{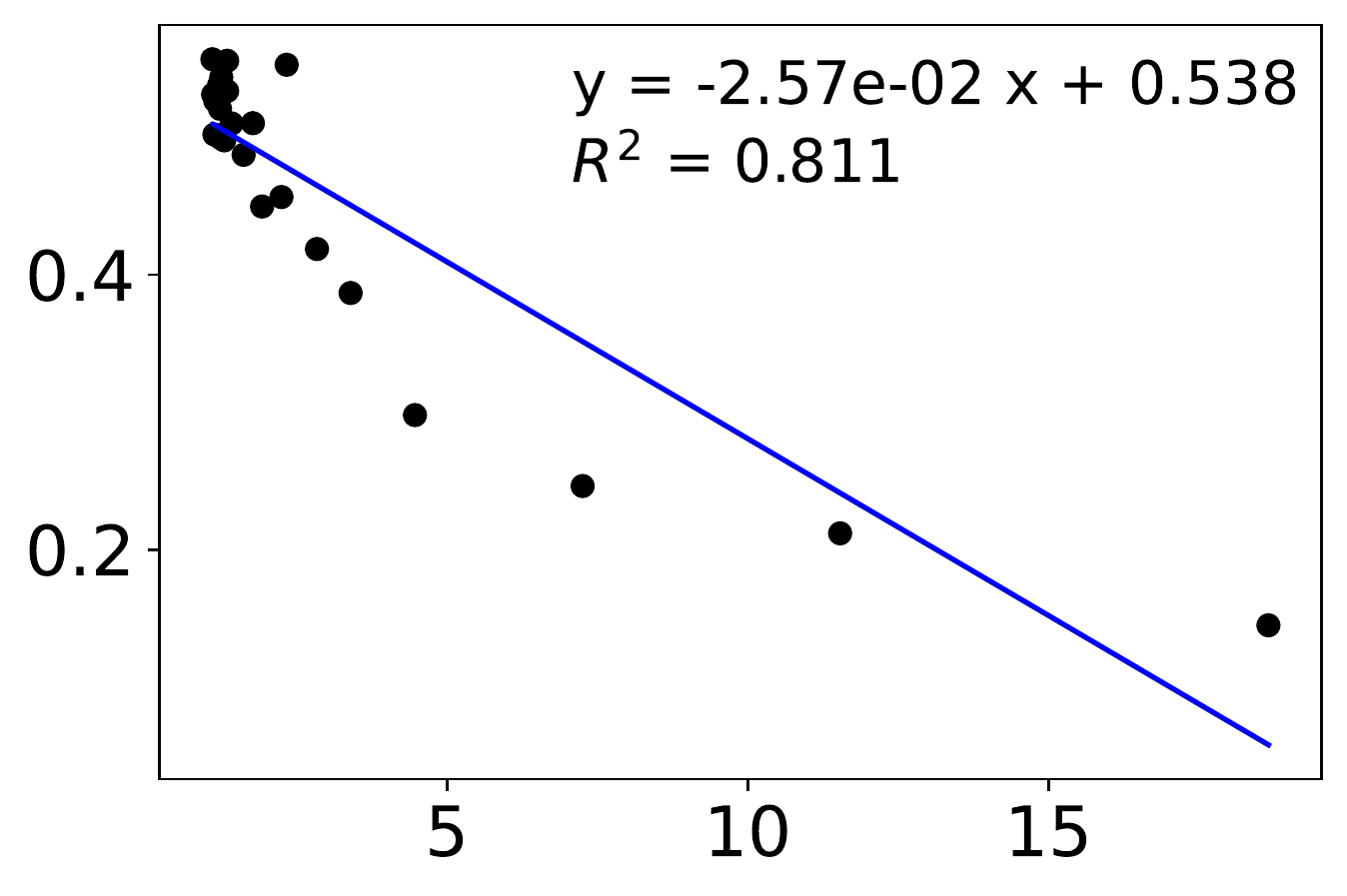}
			\caption{CoLA}
		\end{subfigure}
		~
		\begin{subfigure}[t]{0.48\linewidth}
			\includegraphics[width=0.99\linewidth]{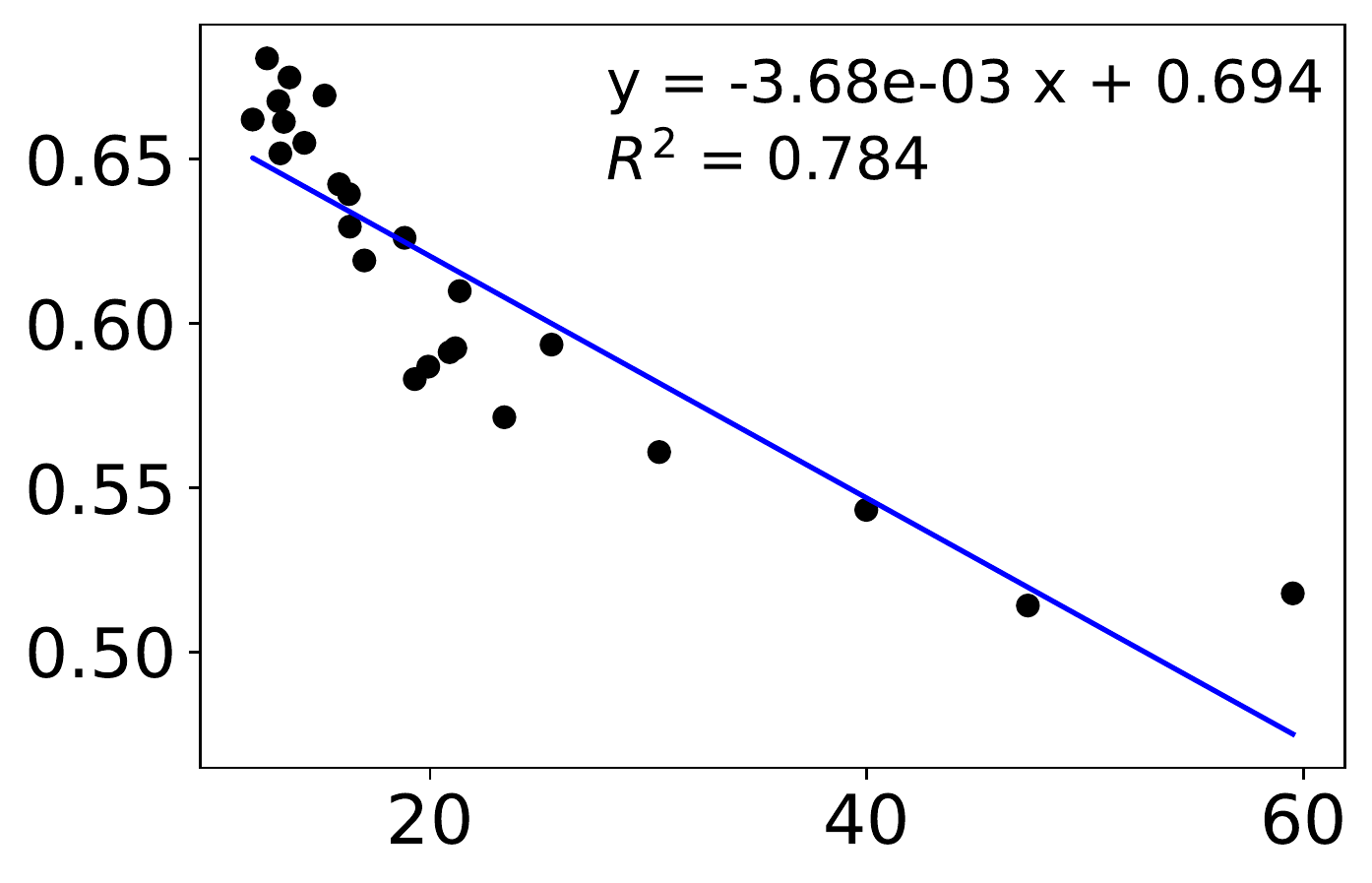}
			\caption{MNLI}
		\end{subfigure}
		
		\begin{subfigure}[t]{0.48\linewidth}
			\includegraphics[width=0.99\linewidth]{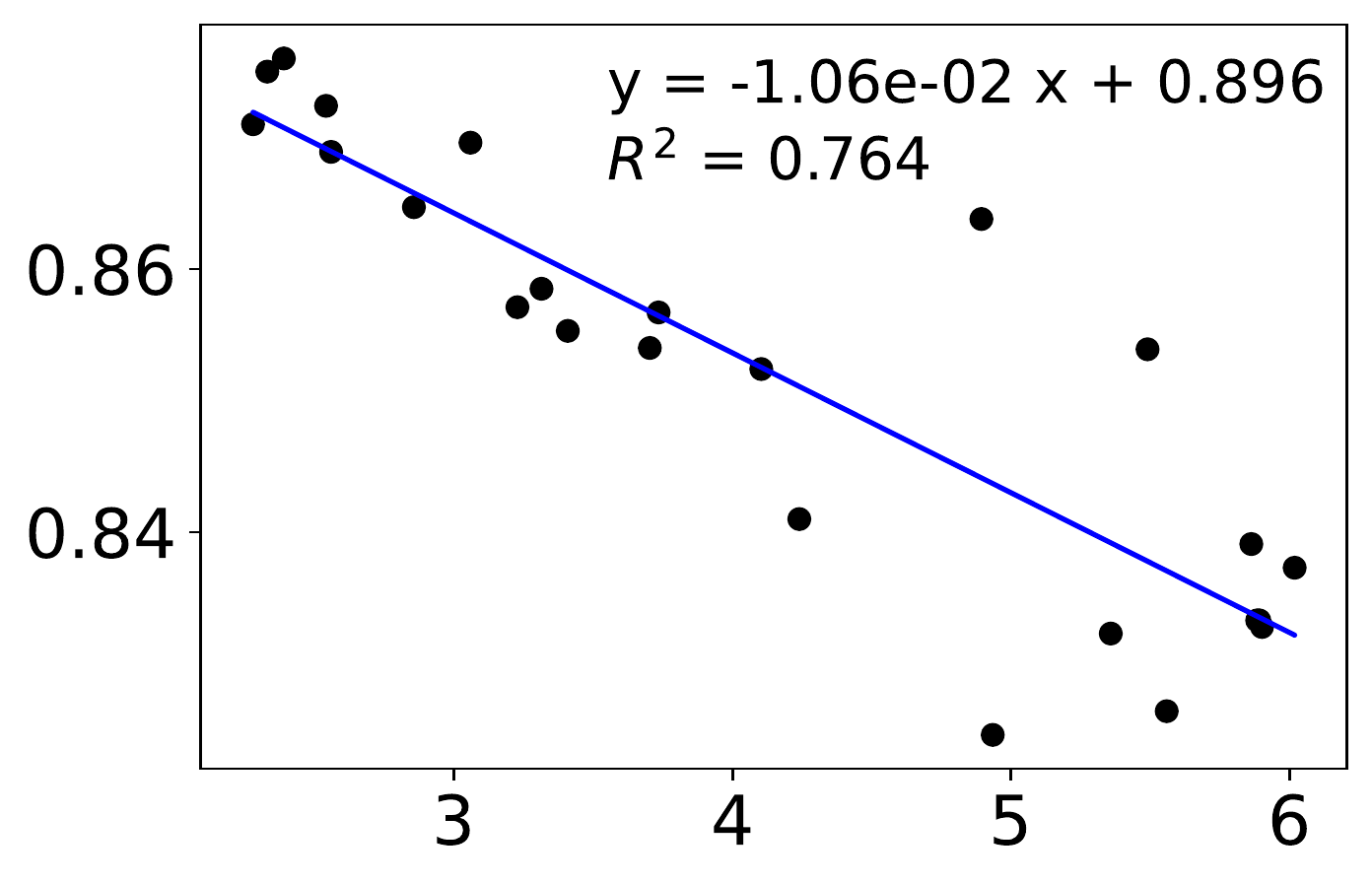}
			\caption{MRPC}
		\end{subfigure}
		~
		\begin{subfigure}[t]{0.48\linewidth}
			\includegraphics[width=0.99\linewidth]{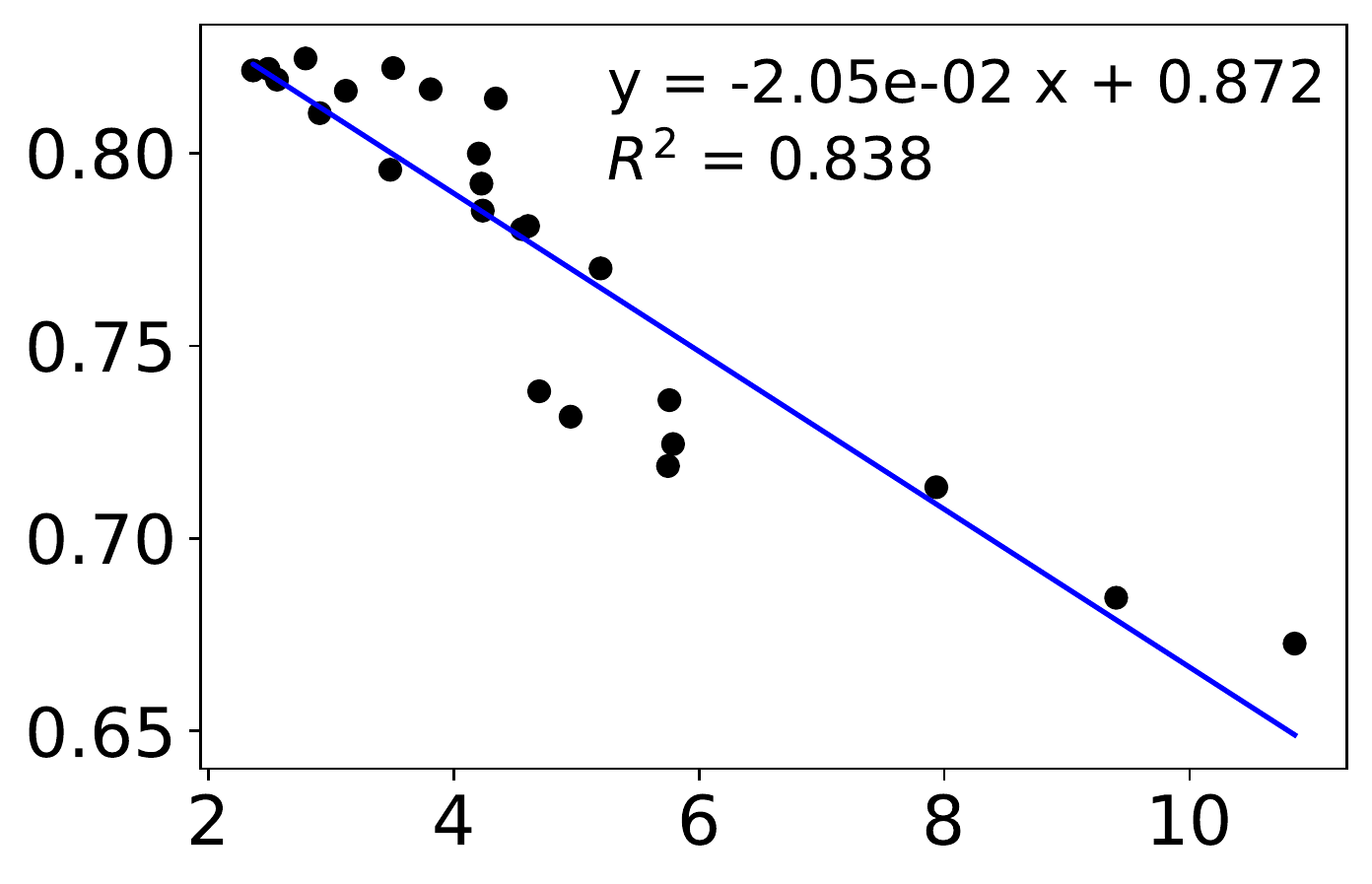}
			\caption{QNLI}
		\end{subfigure}
		
		\begin{subfigure}[t]{0.48\linewidth}
			\includegraphics[width=0.99\linewidth]{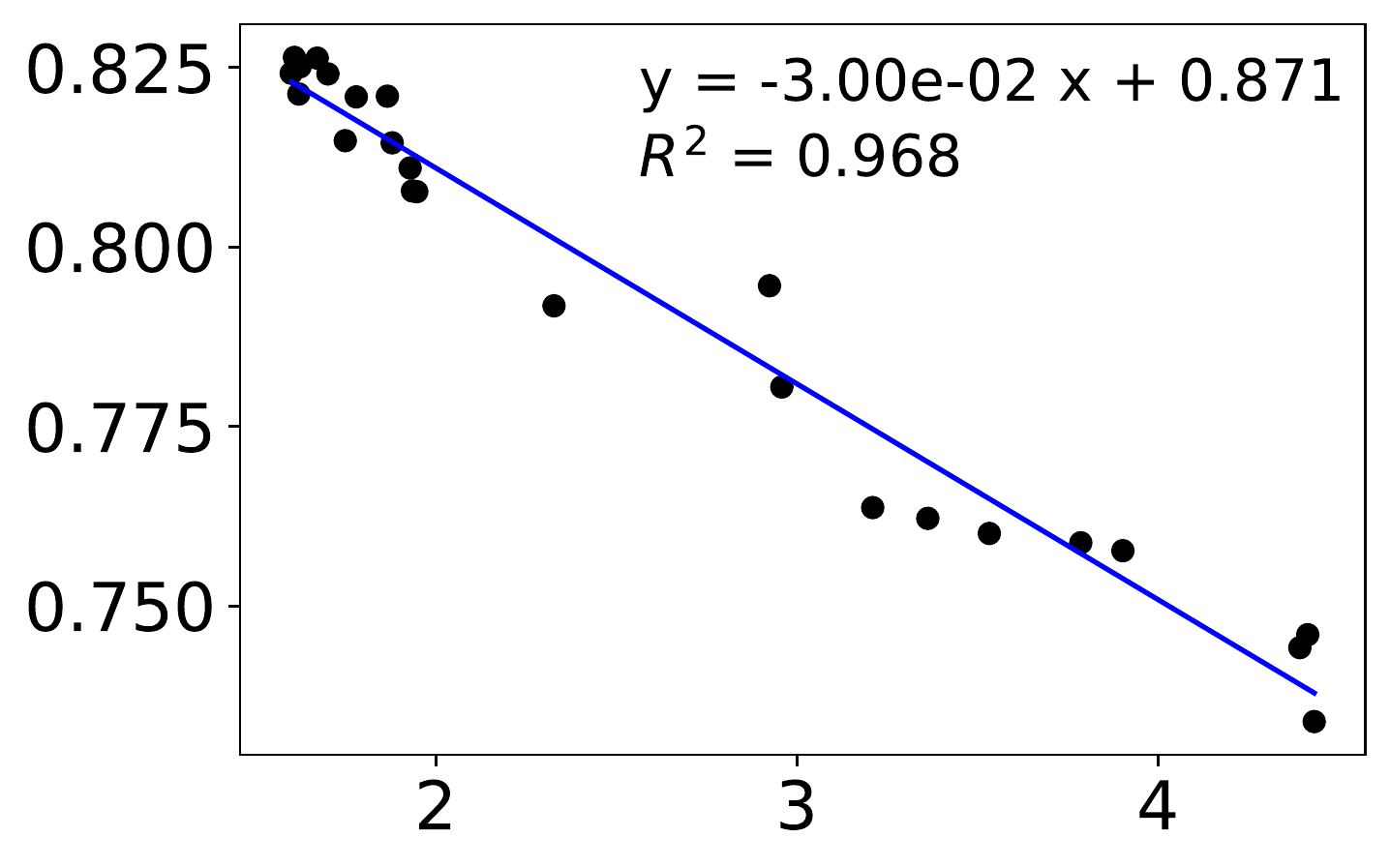}
			\caption{QQP}
		\end{subfigure}
		~
		\begin{subfigure}[t]{0.48\linewidth}
			\includegraphics[width=0.99\linewidth]{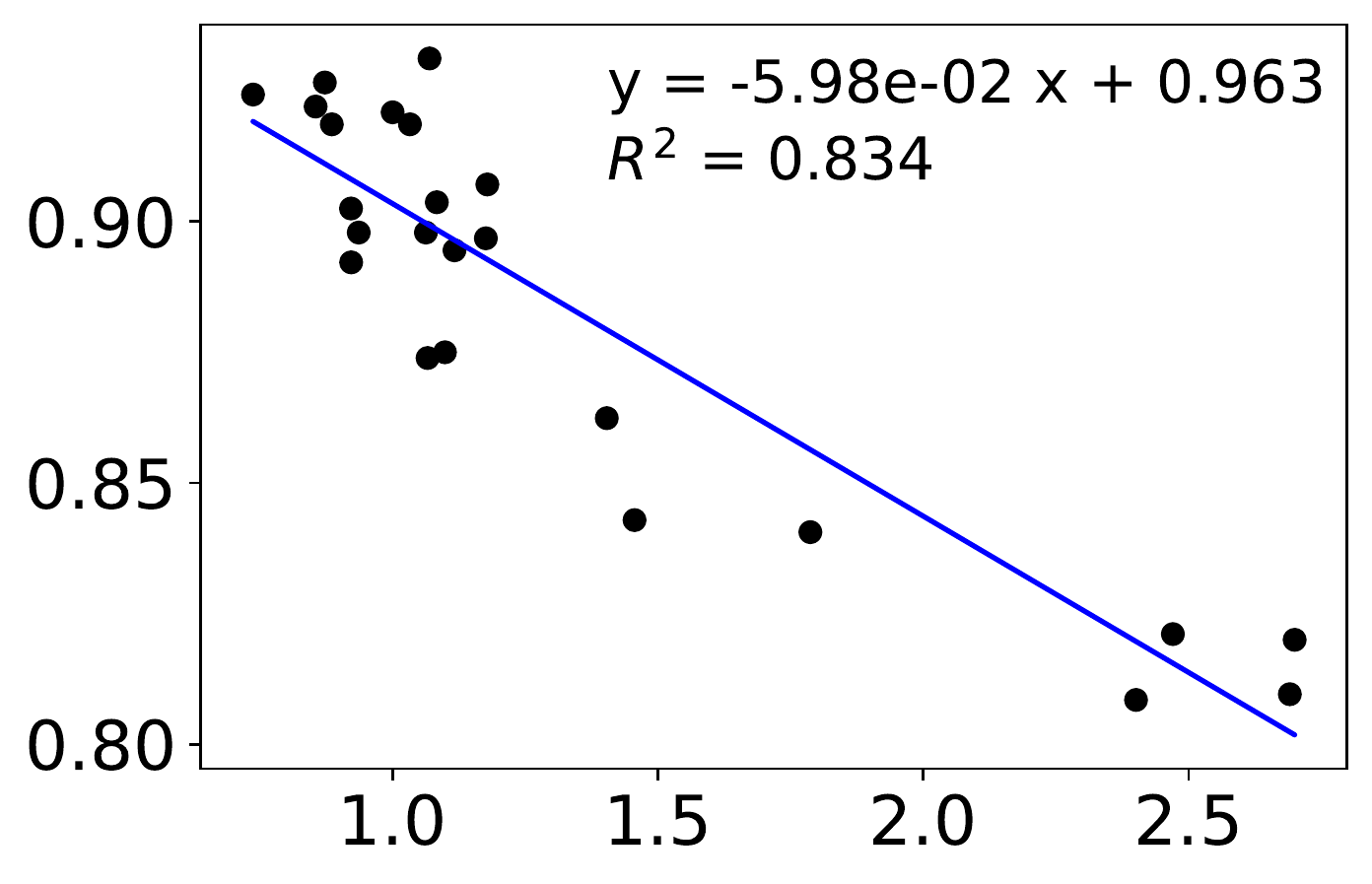}
			\caption{SST-2}
		\end{subfigure}
		\caption{Regressing the per-layer probing performance onto the per-layer task-specialty metric.}\label{fig:nc_regress_pretrain}
	\end{center}
\end{figure}

As discussed in \cref{sec:nc_pretrain}, we regressed the probing performance on $\nu$. 
The regression results are in \cref{fig:nc_regress_pretrain}. 

\subsection{Task-Specialty vs.\@ Probing Performance After Full Fine-Tuning}\label{app:prob}
\begin{figure}[t]
	\begin{center}
	    \begin{subfigure}[t]{0.48\linewidth}
			\includegraphics[width=0.99\linewidth]{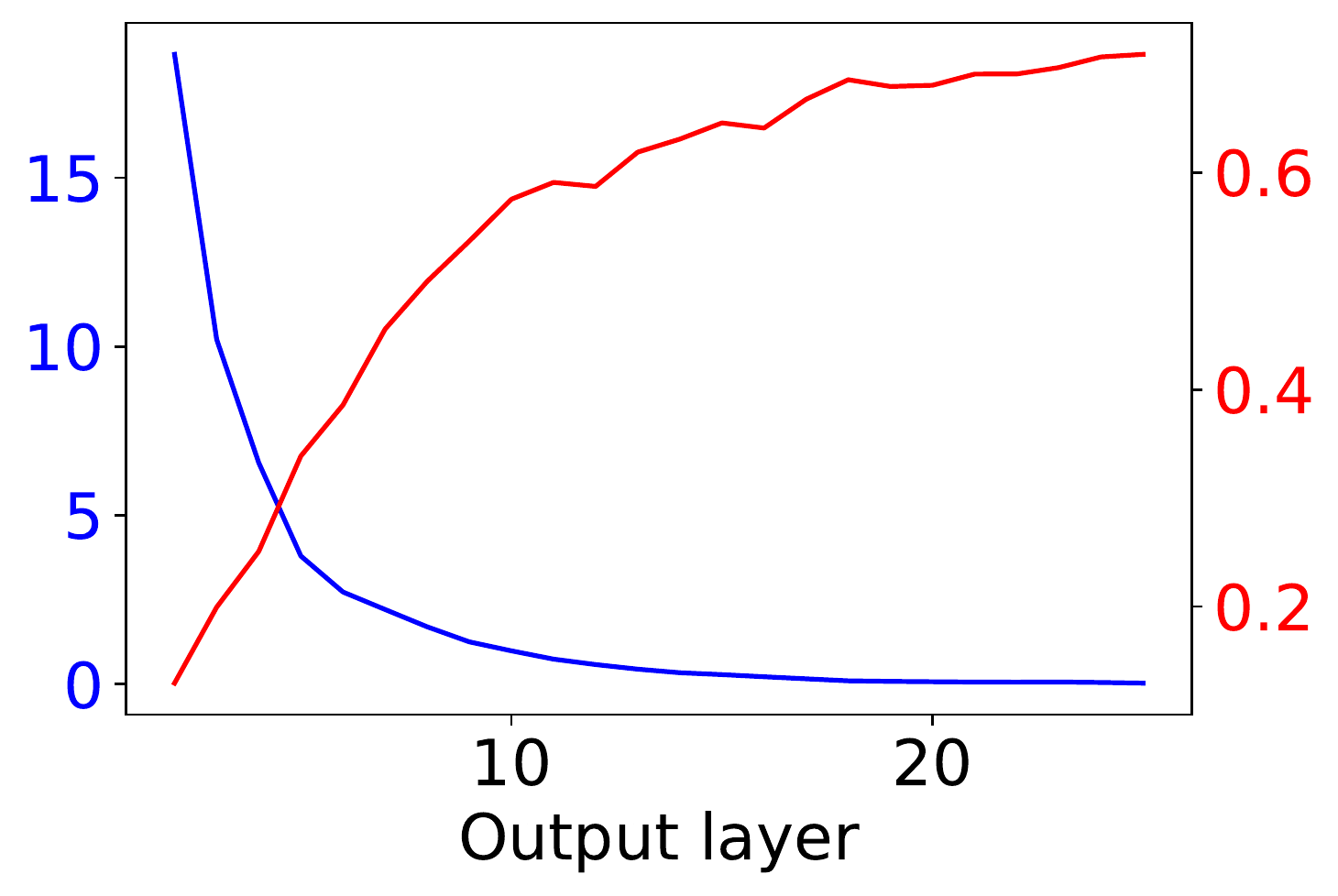}
			\caption{CoLA}
		\end{subfigure}
		~
		\begin{subfigure}[t]{0.48\linewidth}
			\includegraphics[width=0.99\linewidth]{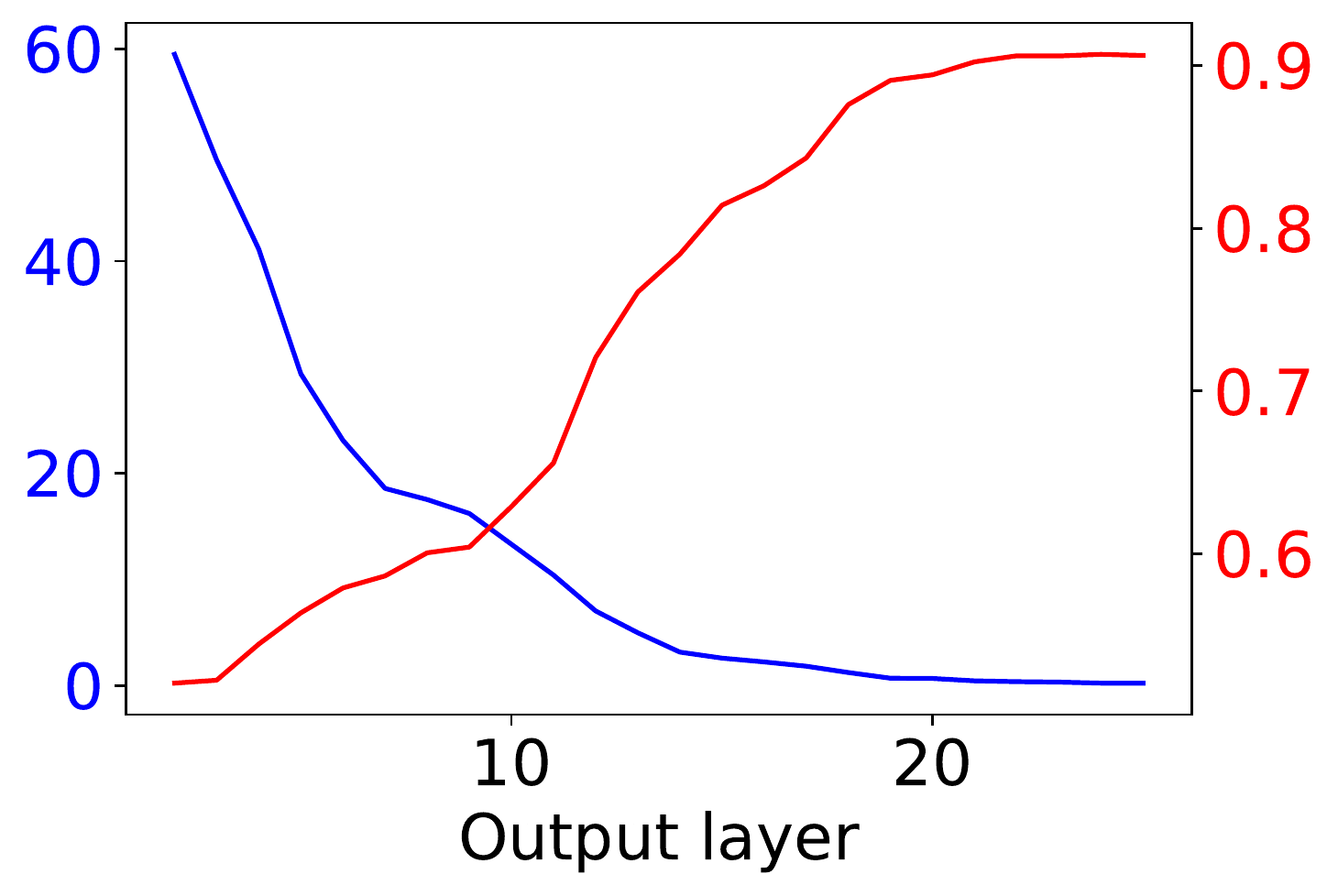}
			\caption{MNLI}
		\end{subfigure}
		
		\begin{subfigure}[t]{0.48\linewidth}
			\includegraphics[width=0.99\linewidth]{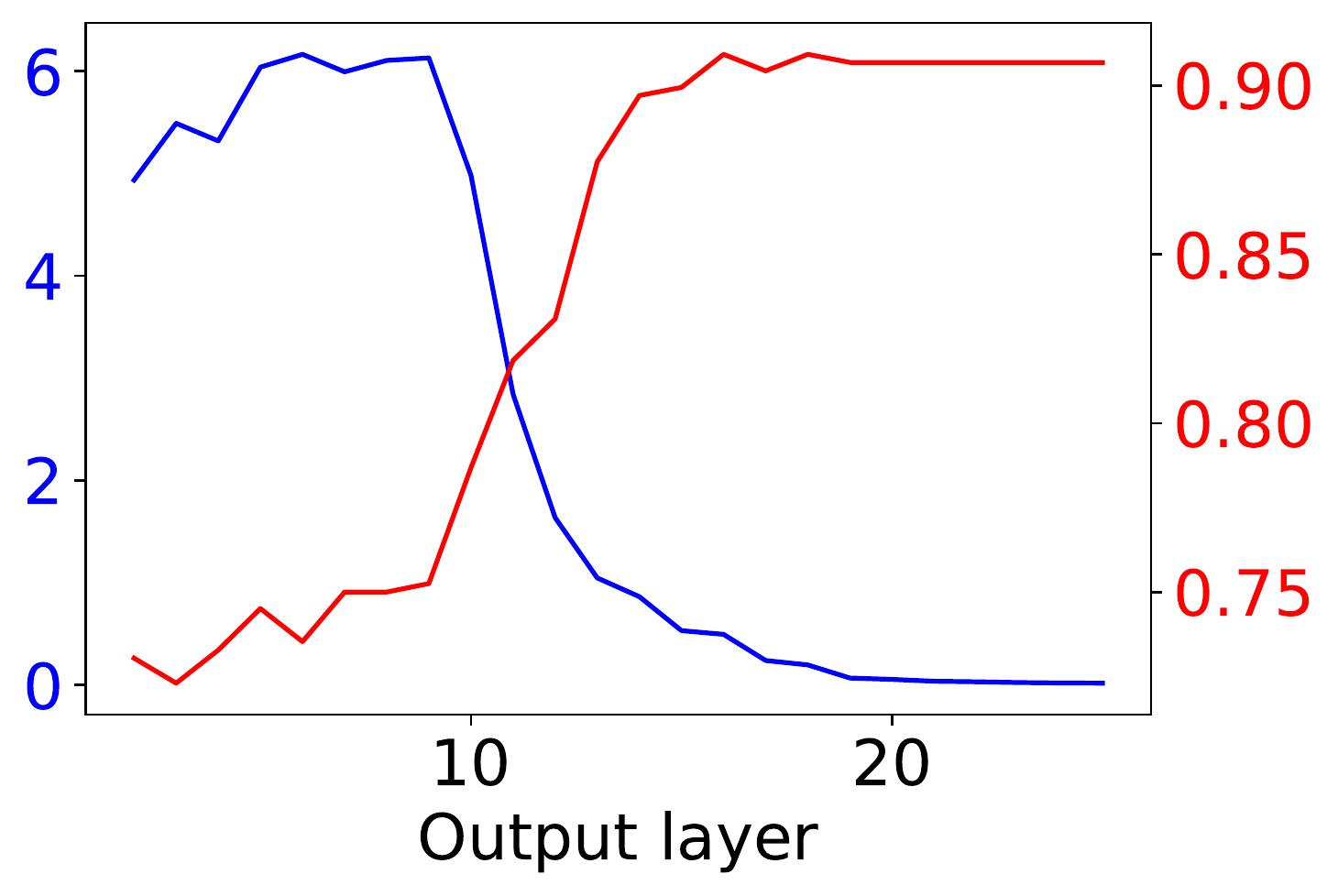}
			\caption{MRPC}
		\end{subfigure}
		~
		\begin{subfigure}[t]{0.48\linewidth}
			\includegraphics[width=0.99\linewidth]{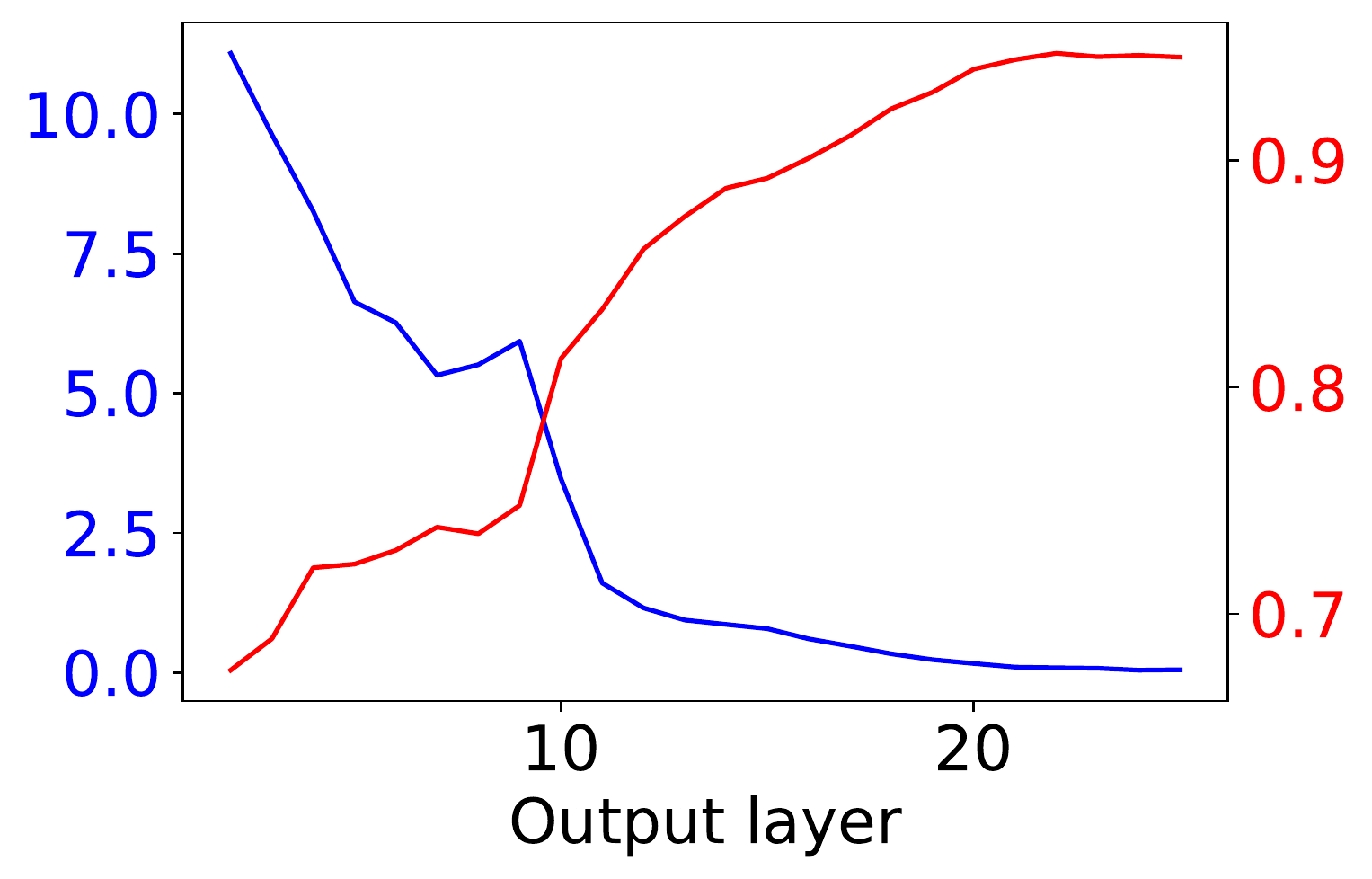}
			\caption{QNLI}
		\end{subfigure}
		
		\begin{subfigure}[t]{0.48\linewidth}
			\includegraphics[width=0.99\linewidth]{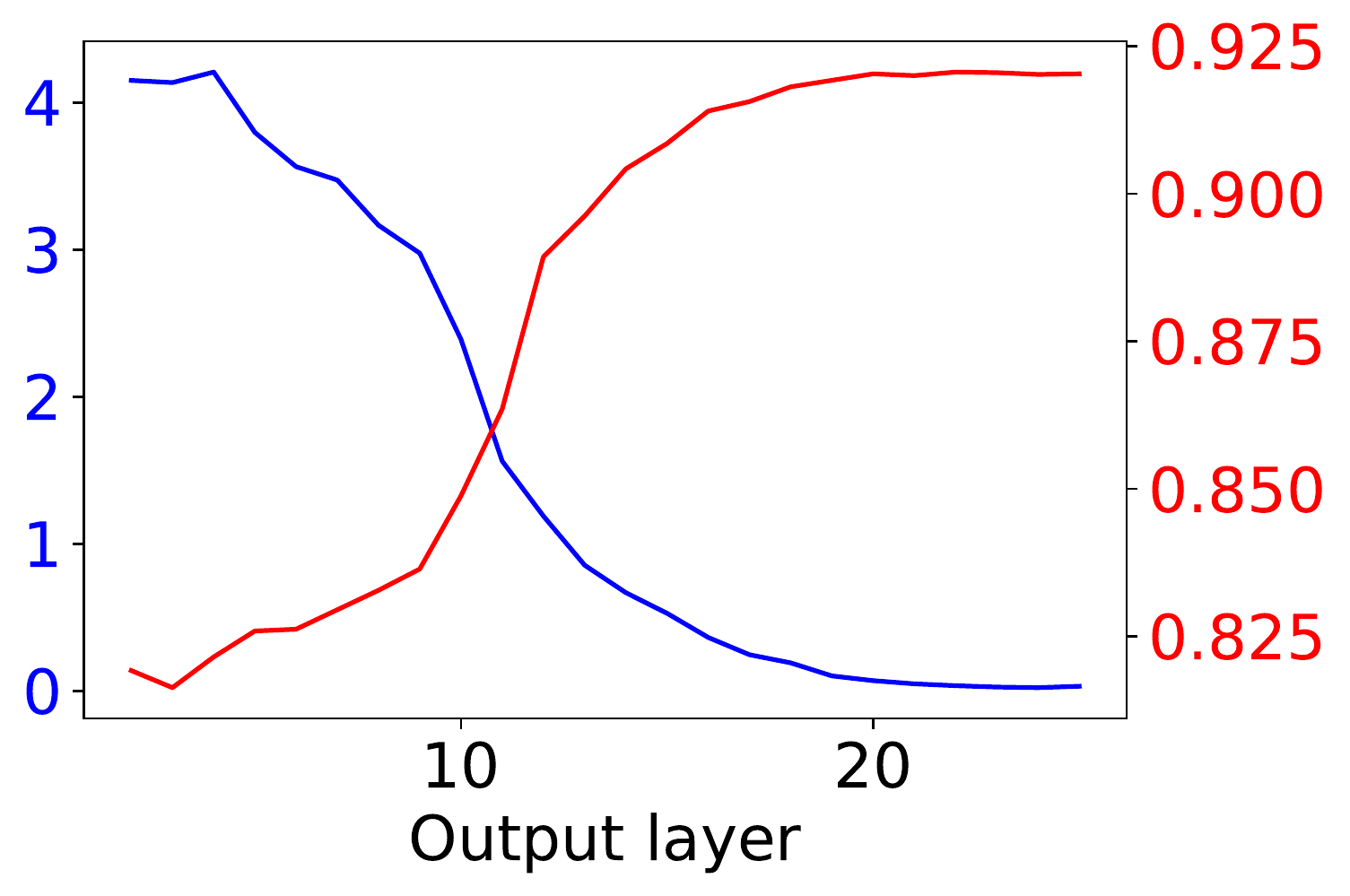}
			\caption{QQP}
		\end{subfigure}
		~
		\begin{subfigure}[t]{0.48\linewidth}
			\includegraphics[width=0.99\linewidth]{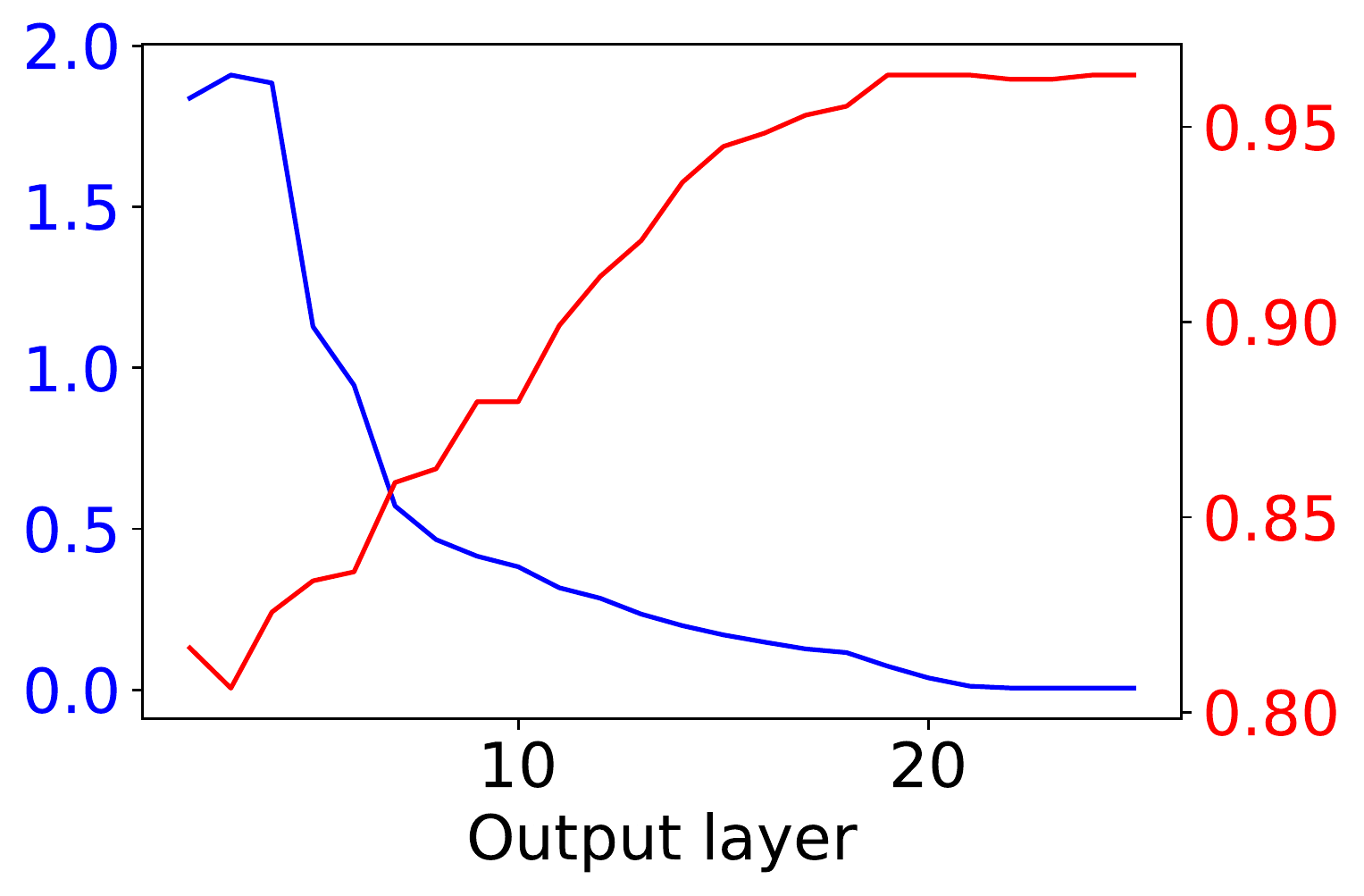}
			\caption{SST-2}
		\end{subfigure}
		\caption{The task-specialty metric (blue) and probing performance (red) of each layer of a RoBERTa model finetuned on each task. Each figure is a GLUE task.}\label{fig:nc_finetune}
	\end{center}
\end{figure}

\begin{figure}[t]
	\begin{center}
	    \begin{subfigure}[t]{0.48\linewidth}
			\includegraphics[width=0.99\linewidth]{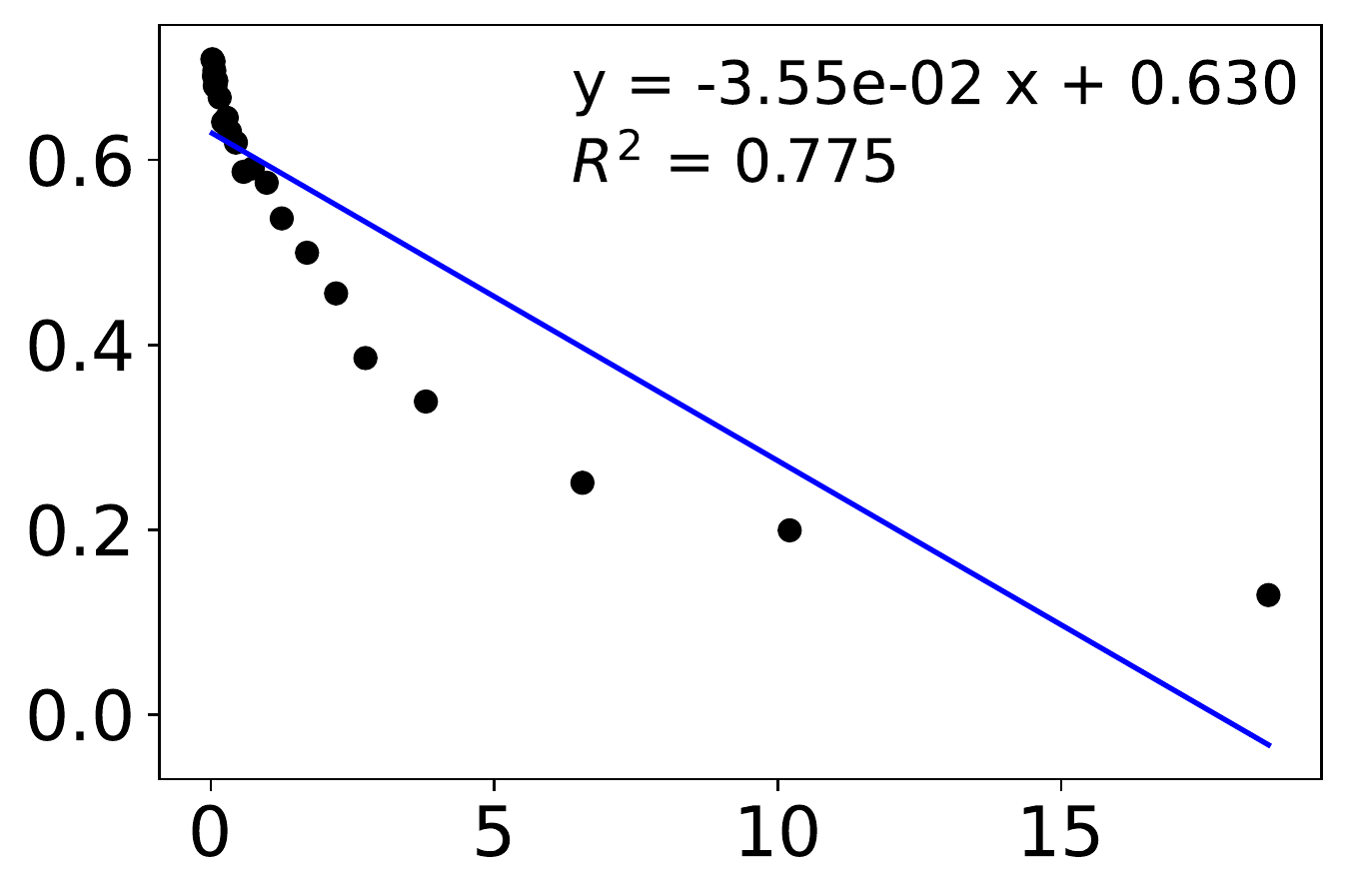}
			\caption{CoLA}
		\end{subfigure}
		~
		\begin{subfigure}[t]{0.48\linewidth}
			\includegraphics[width=0.99\linewidth]{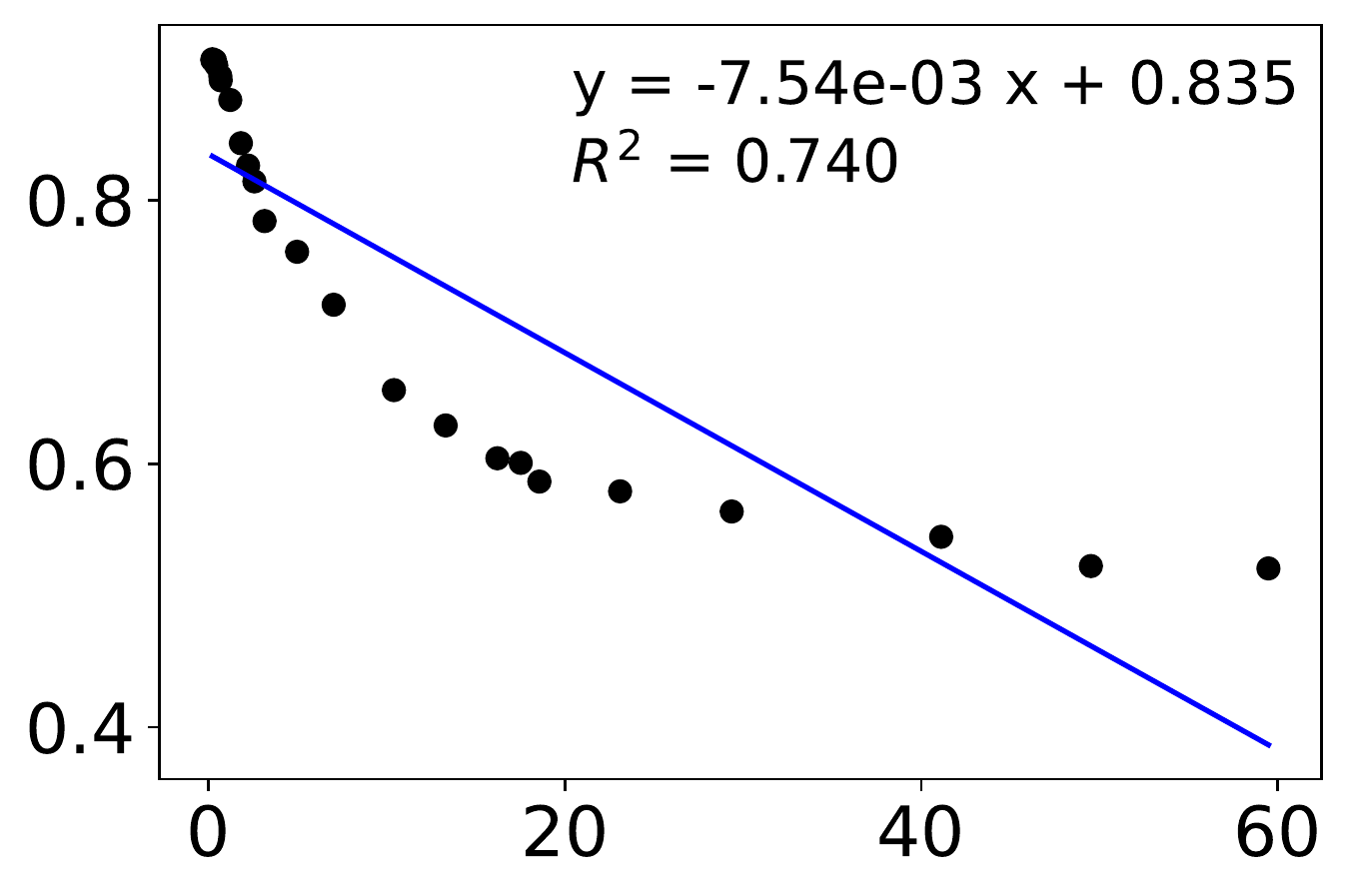}
			\caption{MNLI}
		\end{subfigure}
		
		\begin{subfigure}[t]{0.48\linewidth}
			\includegraphics[width=0.99\linewidth]{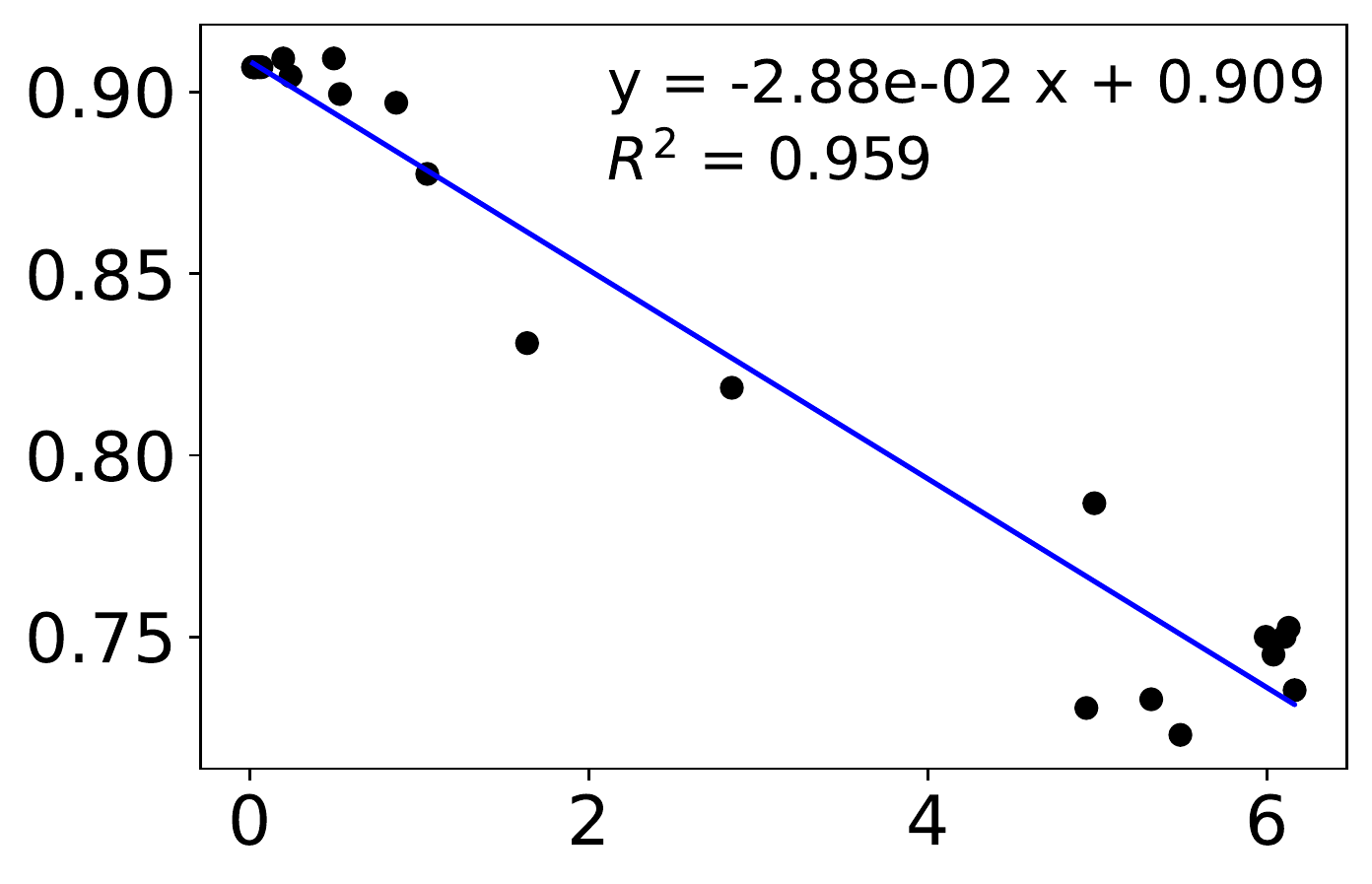}
			\caption{MRPC}
		\end{subfigure}
		~
		\begin{subfigure}[t]{0.48\linewidth}
			\includegraphics[width=0.99\linewidth]{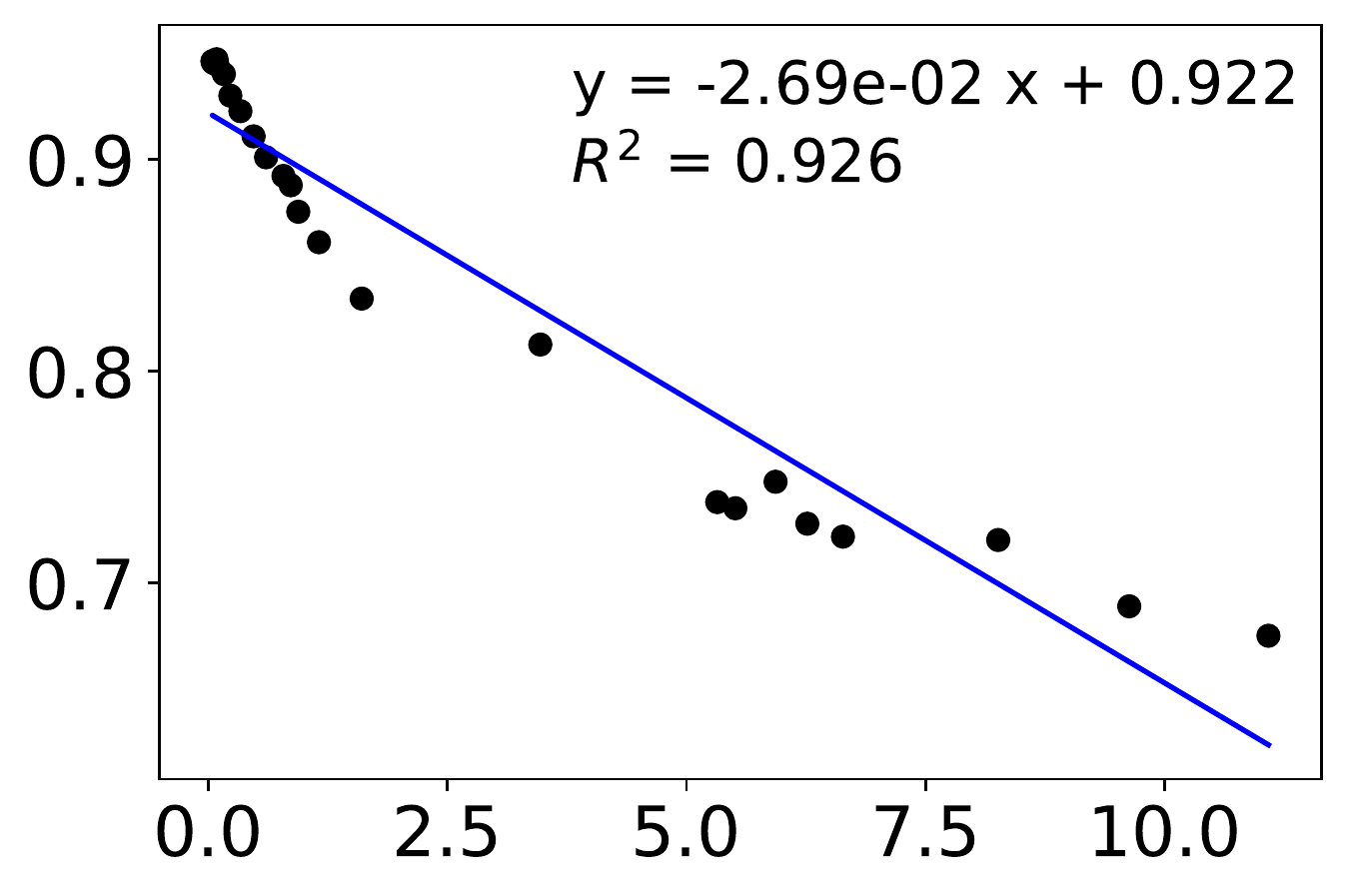}
			\caption{QNLI}
		\end{subfigure}
		
		\begin{subfigure}[t]{0.48\linewidth}
			\includegraphics[width=0.99\linewidth]{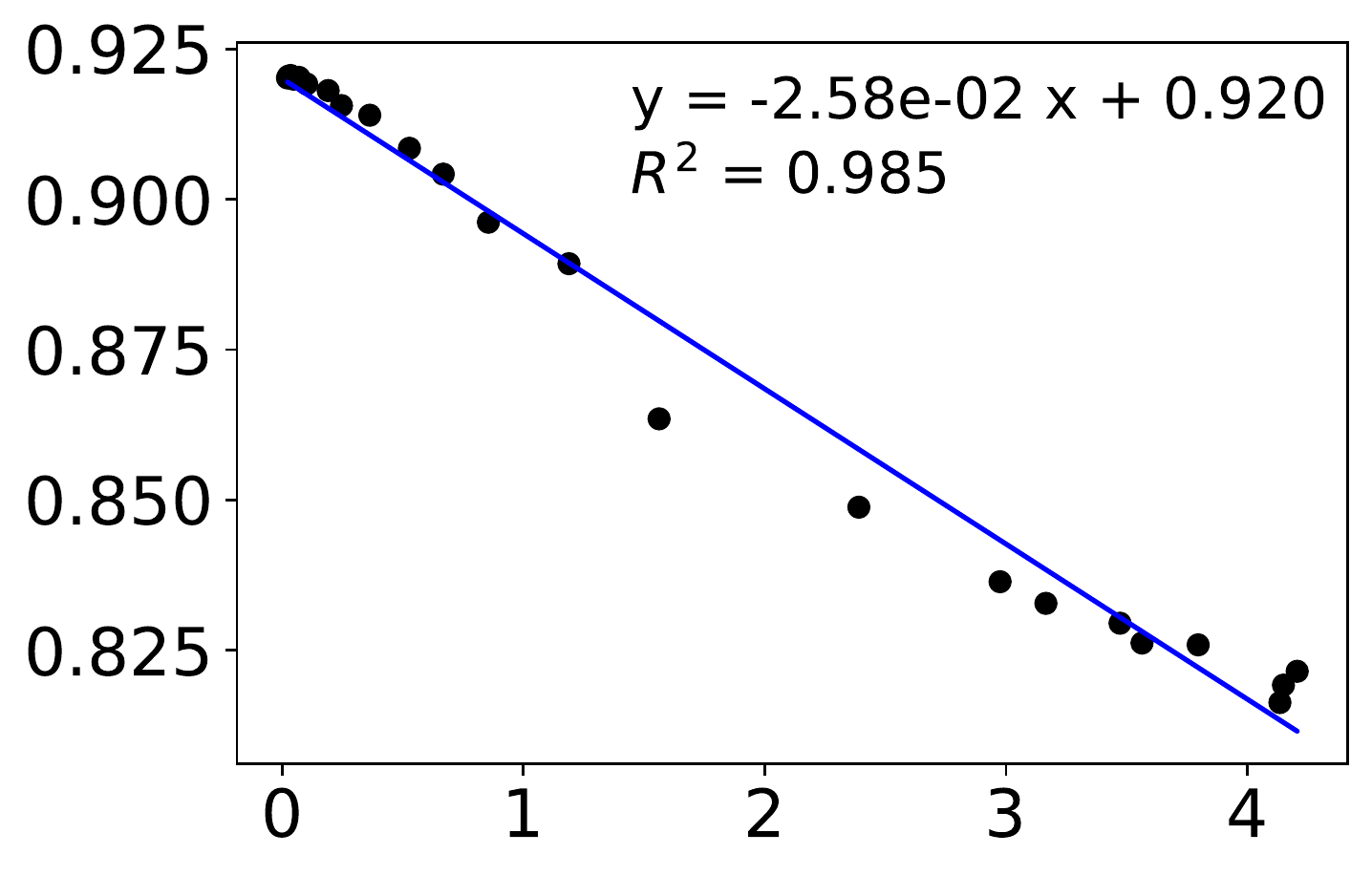}
			\caption{QQP}
		\end{subfigure}
		~
		\begin{subfigure}[t]{0.48\linewidth}
			\includegraphics[width=0.99\linewidth]{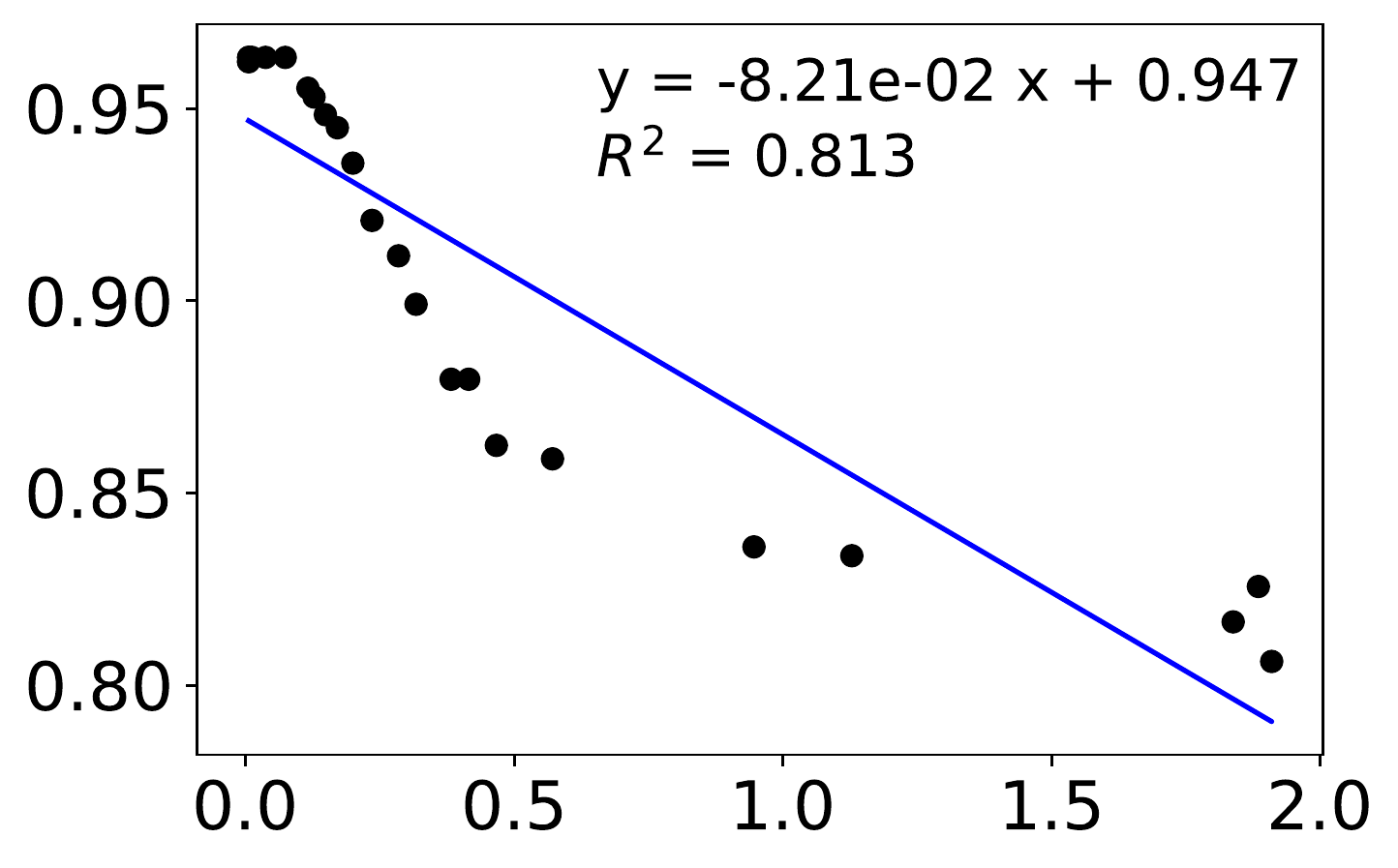}
			\caption{SST-2}
		\end{subfigure}
		\caption{Regressing the per-layer probing performance onto the per-layer task-specialty metric for a RoBERTa model finetuned on each task.}\label{fig:nc_regress_finetune}
	\end{center}
\end{figure}

As discussed in \cref{sec:finetune}, we fully fine-tuned a RoBERTa on each task and obtained the $\nu$ and probing performance of the fine-tuned models. 
The results are presented in \cref{fig:nc_finetune,fig:nc_regress_finetune}.

\subsection{About Computing Task-Specialty Using the CLS Token States}\label{app:othernu}
\begin{figure}[t]
	\begin{center}
	    \begin{subfigure}[t]{0.48\linewidth}
			\includegraphics[width=0.99\linewidth]{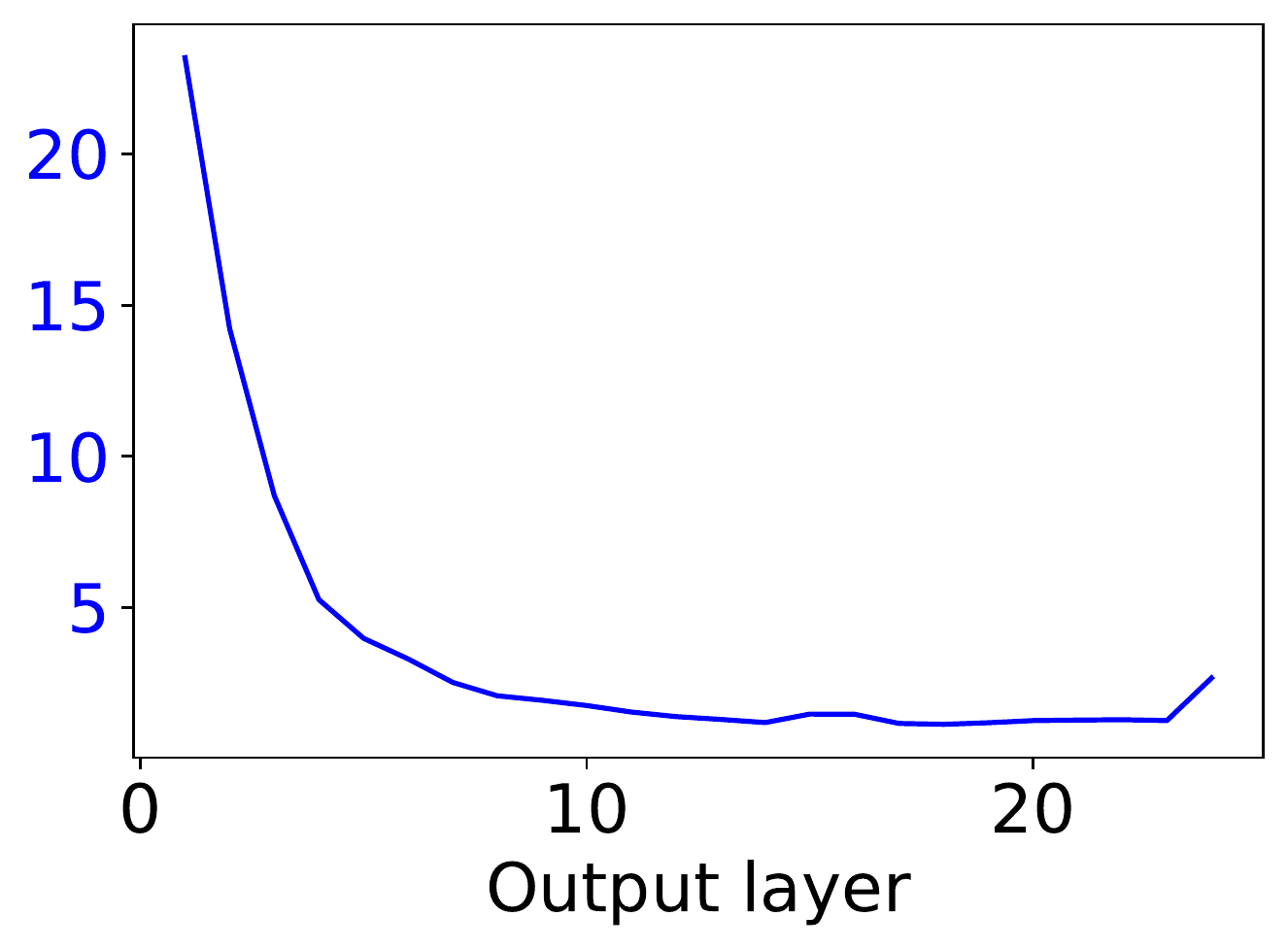}
			\caption{CoLA}
		\end{subfigure}
		~
		\begin{subfigure}[t]{0.48\linewidth}
			\includegraphics[width=0.99\linewidth]{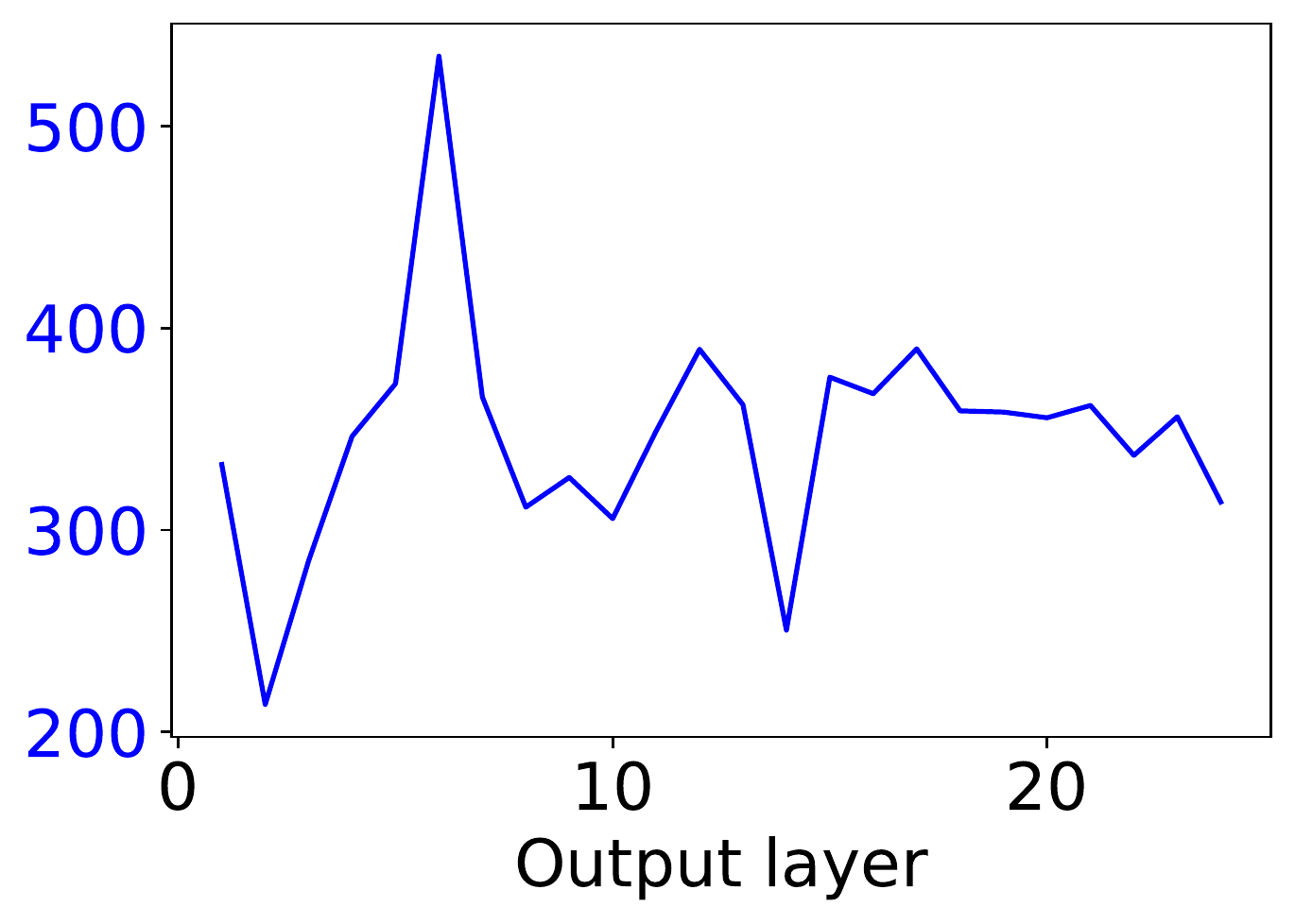}
			\caption{MNLI}
		\end{subfigure}
		
		\begin{subfigure}[t]{0.48\linewidth}
			\includegraphics[width=0.99\linewidth]{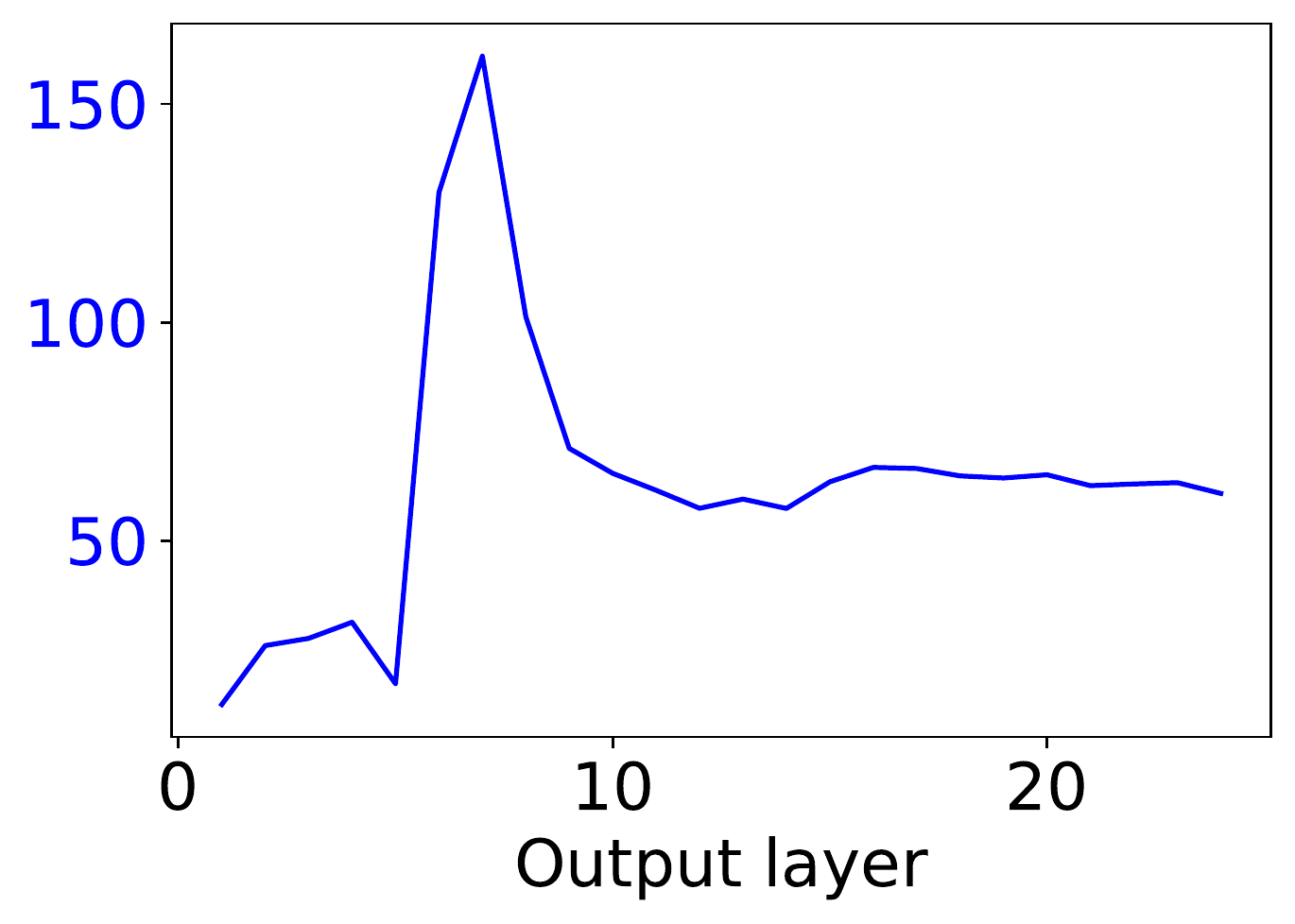}
			\caption{MRPC}
		\end{subfigure}
		~
		\begin{subfigure}[t]{0.48\linewidth}
			\includegraphics[width=0.99\linewidth]{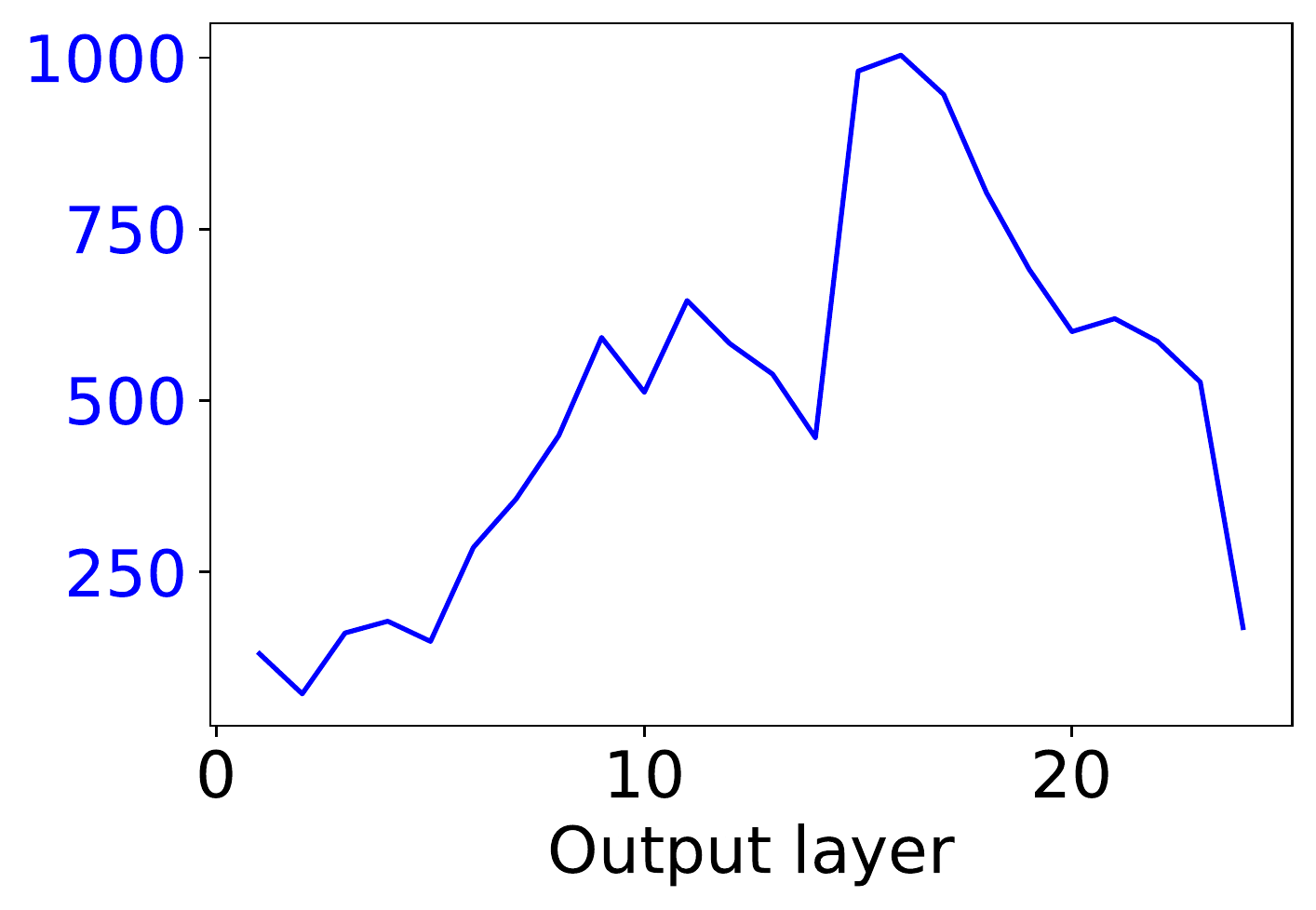}
			\caption{QNLI}
		\end{subfigure}
		
		\begin{subfigure}[t]{0.48\linewidth}
			\includegraphics[width=0.99\linewidth]{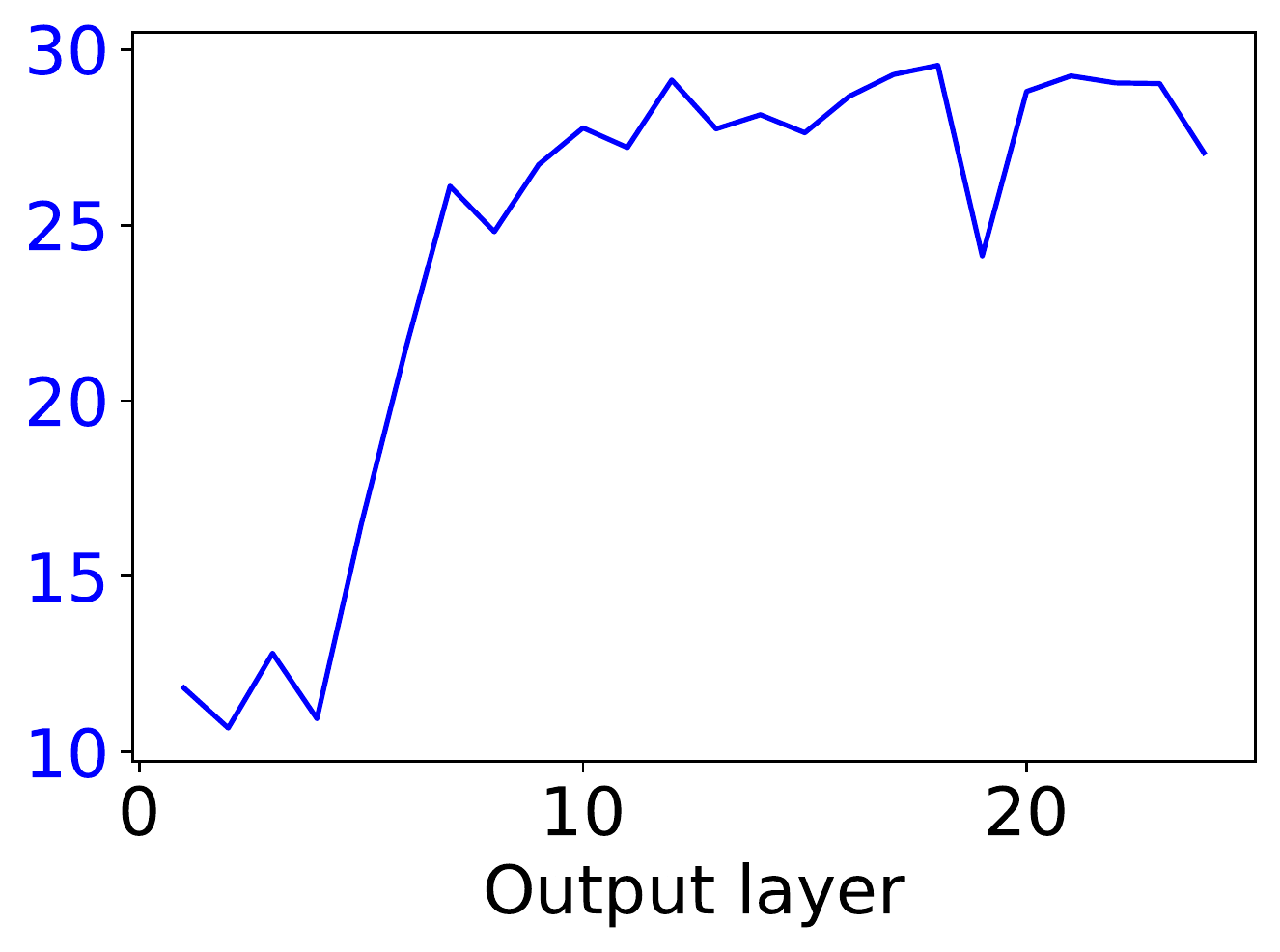}
			\caption{QQP}
		\end{subfigure}
		~
		\begin{subfigure}[t]{0.48\linewidth}
			\includegraphics[width=0.99\linewidth]{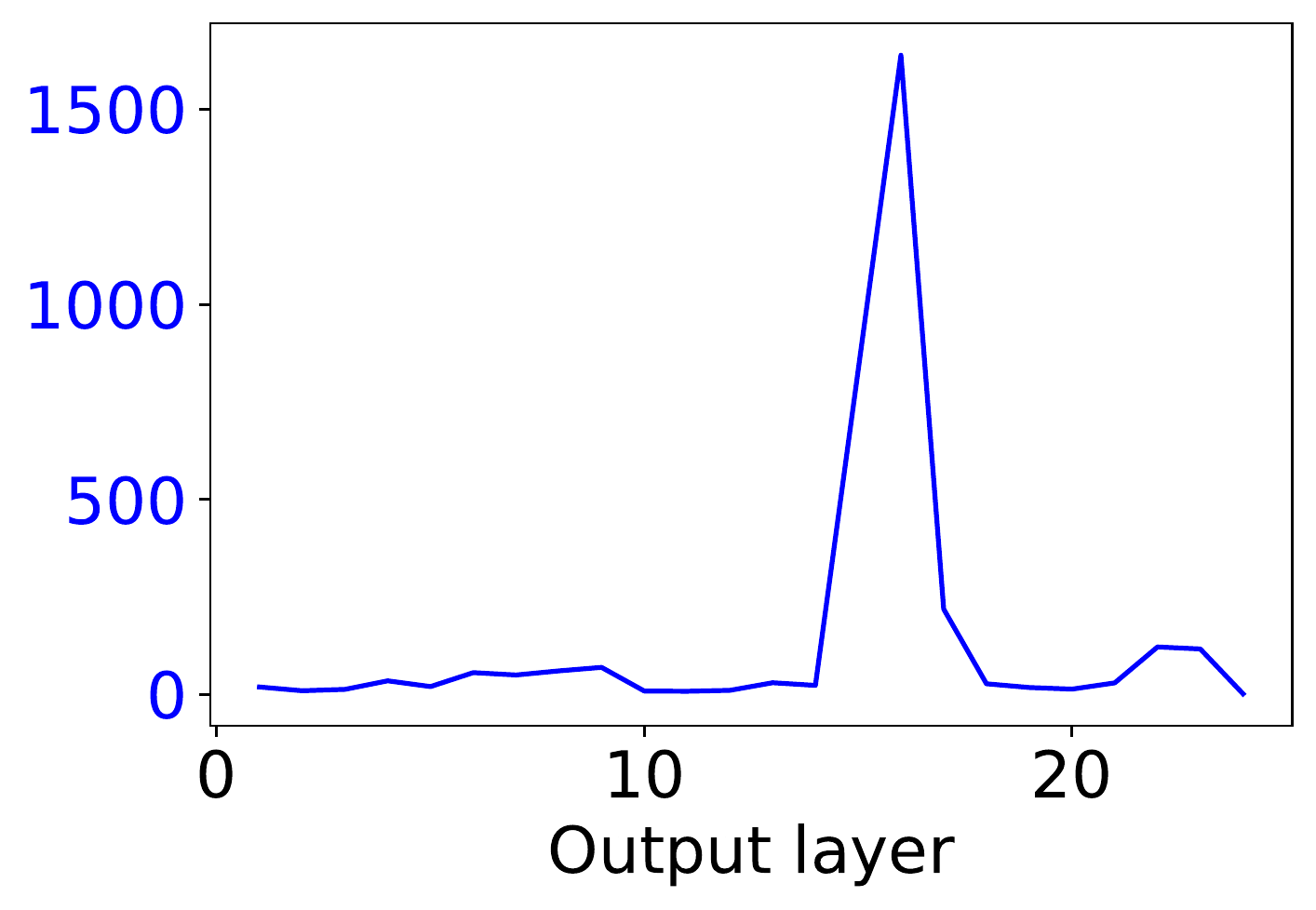}
			\caption{SST-2}
		\end{subfigure}
		\caption{The task-specialty metric computed using the CLS token hidden states. Each figure is a GLUE task.}\label{fig:nc_pretrain_cls}
	\end{center}
\end{figure}

As discussed in \cref{sec:ablation}, we computed $\nu^{(\ell)}$ using the CLS token hidden states and found that the $\nu$ curves would become less trustable. 
The curves are in \cref{fig:nc_pretrain_cls}. 

\subsection{Robustness to Data Imbalance and Data Scarcity}\label{app:nu_sensitivity}\label{app:nu_imbalance}\label{app:mnli_sensitivity}

As discussed in \cref{sec:ablation}, we conducted a series of experiments with SST-2 and MNLI to verify how sensitive our metric is to data imbalance and data scarcity. 
For each experiment on SST-2, we had to decide on two key quantities: the number of training examples $N$ and the portion $p$ that are drawn from the negative group. 
In other words, we built a dataset $\{ (\vec{x}_n, y_n)\}_{n=1}^{N}$ by sampling $pN$ examples from the negative group and $(1-p)N$ examples from the positive group. %
For the data-imbalanced experiments, we fixed $N=20000$ and used $p \in \{0.5, 0.25, 0.1, 0.05 \}$. 
For the data-scarce experiments, we fixed $p=0.5$ and used $N \in \{40000, 20000, 5000, 200 \}$.

\begin{figure}[t]
	\begin{center}
		\begin{subfigure}[t]{0.48\linewidth}
			\includegraphics[width=0.95\linewidth]{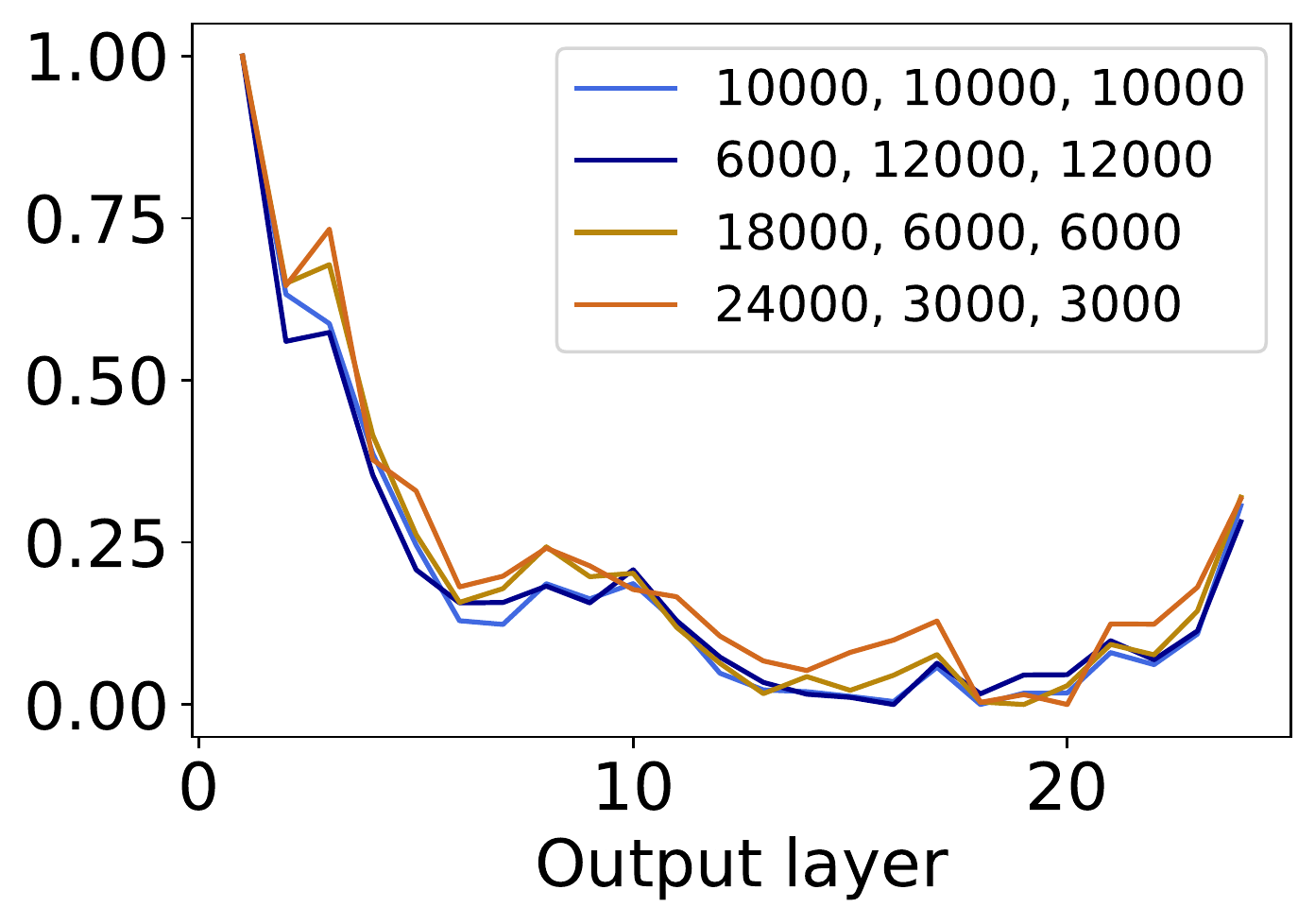}
			\vspace{-4pt}
			\caption{$\nu$ curves}
			\label{fig:layerwise_imbalance_mnli}
		\end{subfigure}
		\hfill
		\begin{subfigure}[t]{0.48\linewidth}
			\includegraphics[width=0.95\linewidth]{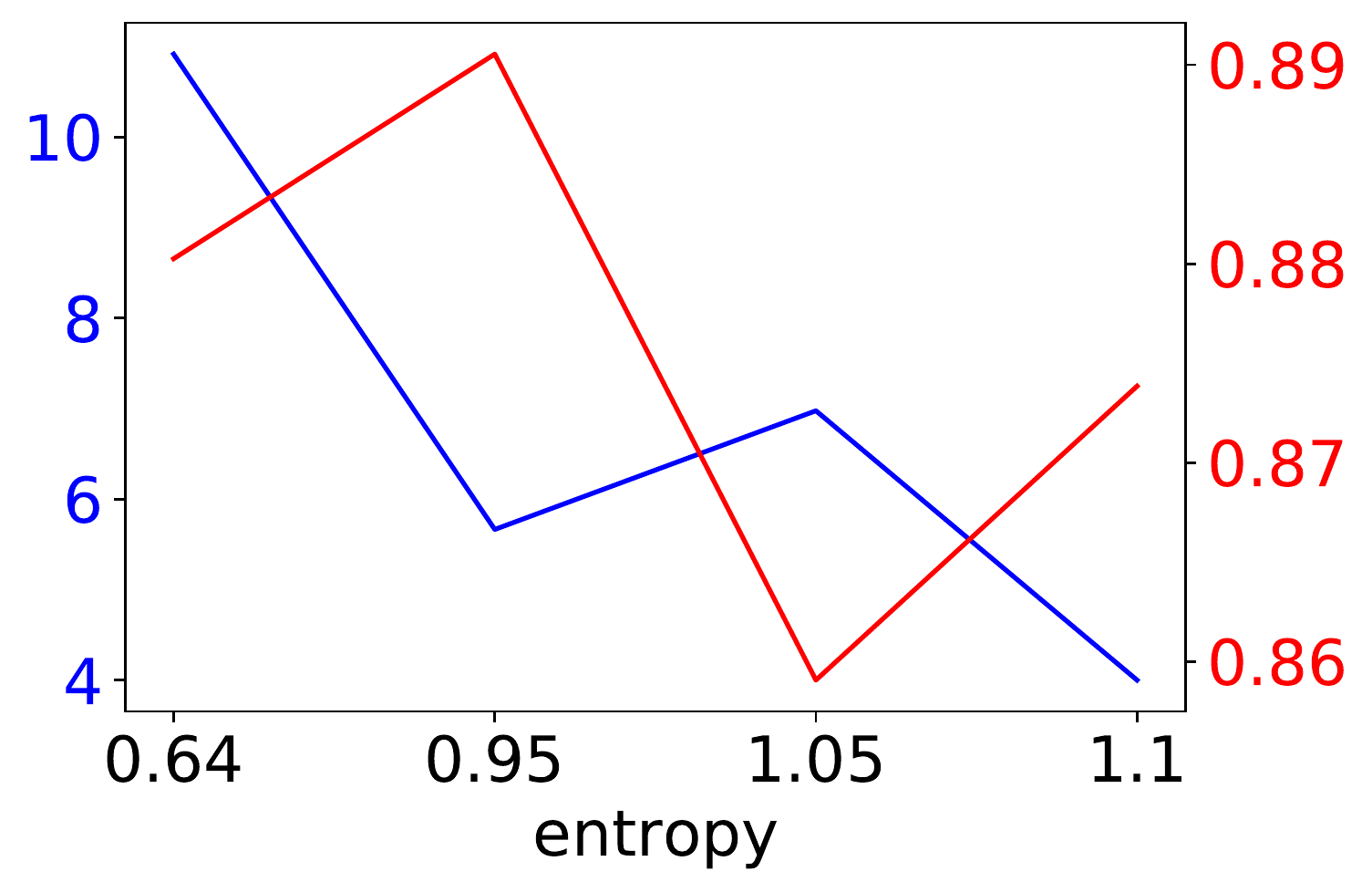}
			\vspace{-4pt}
            \caption{$\zeta$ (\textcolor{blue}{blue}) and $|\rho|$ (\textcolor{red}{red})}
			\label{fig:metric_imbalance_mnli}
		\end{subfigure}
	\vspace{-8pt}
	\caption{Results of data-imbalanced experiments on MNLI.}\label{fig:sensitivity_imbalance_mnli}
	\end{center}
	\vspace{-4pt}
\end{figure}

For MNLI, we need to decide the sample size of each class because it is a classification task with three classes. 
We use $(n_0, n_1, n_2)$ to denote the sample size of the three classes.
For the data-imbalanced experiments, $(n_0, n_1, n_2)$ is chosen from $\{(10000, 10000, 10000), (6000, 12000, 12000),\\ (18000, 6000, 6000), (24000, 3000, 3000)\}$. The total number is always $30000$ to avoid the effect of data amount. 
Similarly as in \cref{sec:imbalance}, we plotted $\nu$ and $\zeta$ and $|\rho|$ in \cref{fig:sensitivity_imbalance_mnli}. 
Our metric is robust to data imbalance on MNLI.

\begin{figure}[t]
	\begin{center}
		\begin{subfigure}[t]{0.48\linewidth}
			\includegraphics[width=0.95\linewidth]{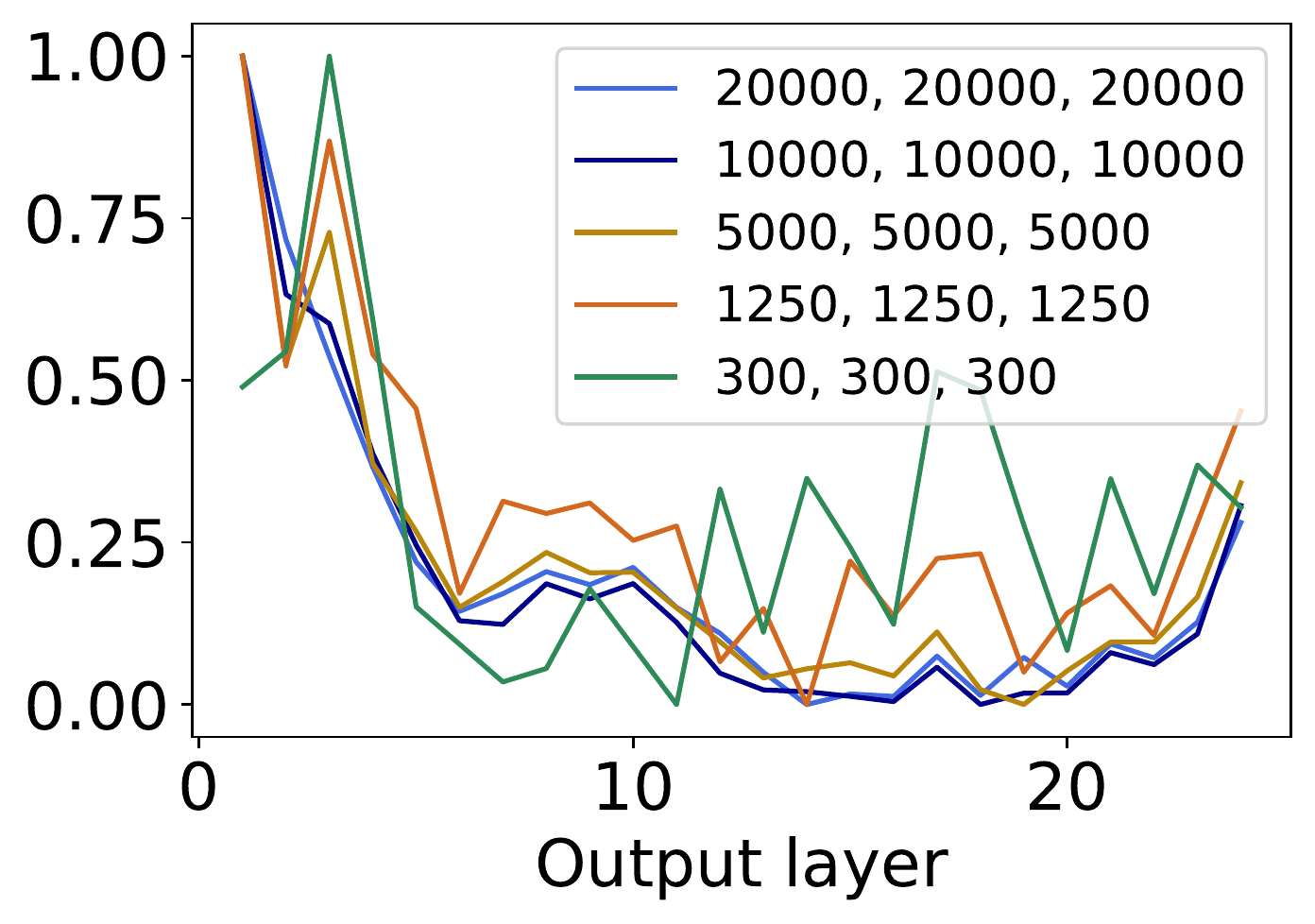}
			\vspace{-4pt}
            \caption{$\nu$ curves}
			\label{fig:layerwise_scarcity_mnli}
		\end{subfigure}
		\hfill
		\begin{subfigure}[t]{0.48\linewidth}
			\includegraphics[width=0.95\linewidth]{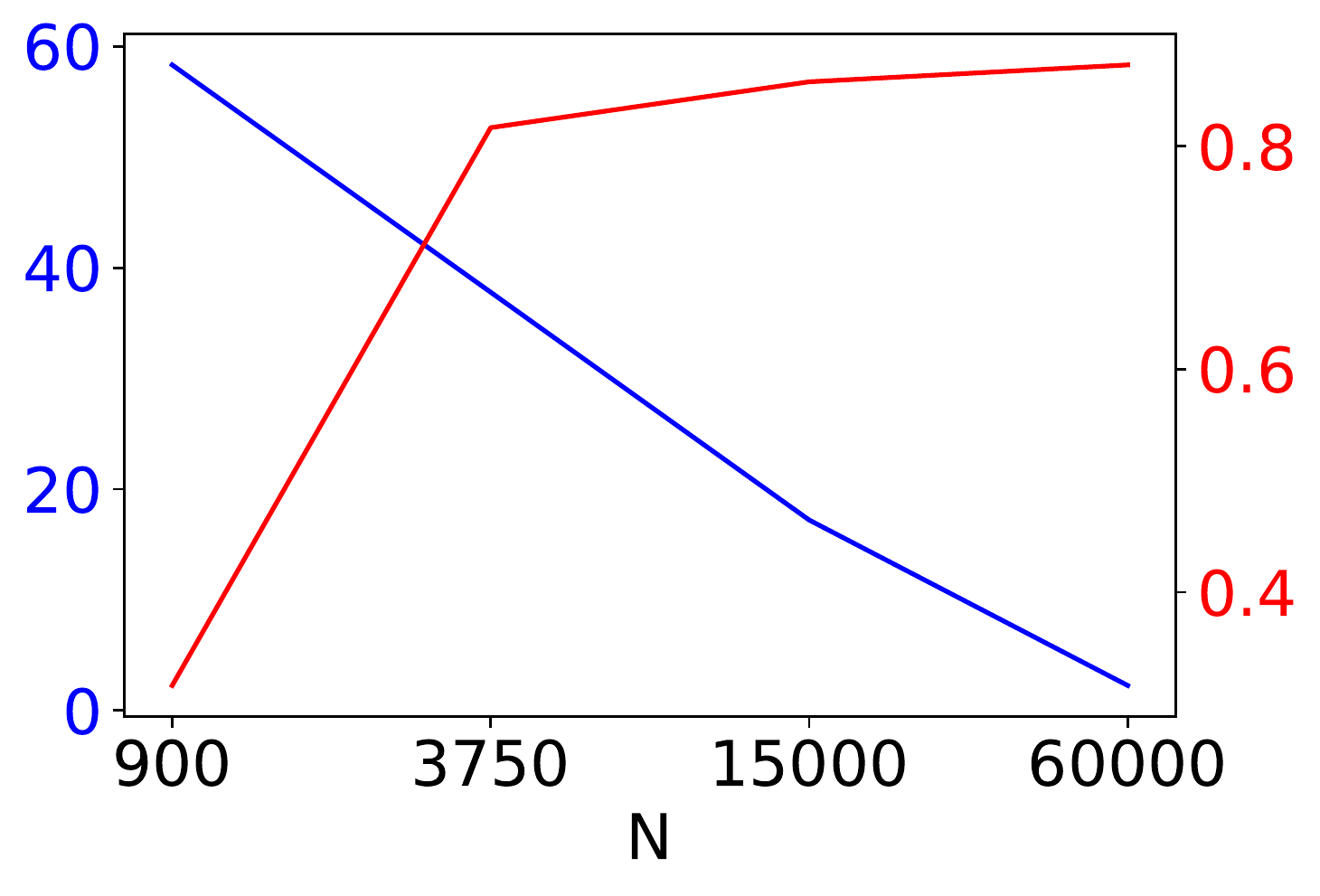}
			\vspace{-4pt}
			\caption{$\zeta$ (\textcolor{blue}{blue}) and $|\rho|$ (\textcolor{red}{red})}
			\label{fig:metric_scarcity_mnli}
		\end{subfigure}
	\vspace{-4pt}
	\caption{Results of data-scarce experiments on MNLI.}\label{fig:sensitivity_scarcity_mnli}
	\end{center}
	\vspace{-2pt}
\end{figure}

For the data-scarce experiments, $(n_0, n_1, n_2)$ is chosen from $\{(20000, 20000, 20000), (10000, \\10000, 10000), (5000, 5000, 5000), (1250, 1250, \\1250), (300, 300, 300)\}$. The sampled dataset is always balanced.
We plotted $\nu$ and $\zeta$ and $|\rho|$ in \cref{fig:sensitivity_scarcity_mnli}. 
Our metric has the same trend on MNLI with more than a few thousand samples.
This conclusion is consistent with the conclusion on SST-2. 

\subsection{About Alternatives to Our Task-Specialty Metric}\label{app:other_metric}\label{app:only}
\paragraph{Canonical correlation analysis.}\label{app:cca}
As discussed in \cref{sec:cca}, a potential alternative to our proposed metric is the canonical correlation between the hidden states $\vec{h}_n^{(\ell)}$ and the class labels $y_n$. 
Hidden state vectors $\vec{h}_n^{(\ell)}$ and one-hot vectors $\vec{y}_n$ can be viewed as i.i.d. samples from random vectors $\vec{h}^{(\ell)}$ and $\vec{y}$ respectively, whose relationship can be quantified by canonical correlation analysis.
It maximizes the correlations between linear projections of paired samples from these random vectors (or ``views"): $\vec{v}_1^{(\ell)}, \vec{w}_1^{(\ell)} = \argmax_{\vec{v}, \vec{w}} \text{corr}(\vec{v}^{\top} \vec{h}^{(\ell)}, \vec{w}^\top \vec{y})$. 
The subsequent directions $\vec{v}_j, \vec{w}_j$, maximize the same correlation subject to each new projection being uncorrelated with others in the same view for $2 \leq i \leq J=\min\{ D, |\mathcal{Y}|\}$,  where $D$ is the dimension of hidden states.
The algorithm thus provides $J$ correlation values. The CCA score is measured as the average over all but the last correlation. This is based on the assumption that the last direction measures noise correlations.
This assumption is confirmed by our empirical observation that the last correlation values are always close to zero (of the order 1e-2).

\begin{figure}[t]
	\begin{center}
	    \begin{subfigure}[t]{0.48\linewidth}
			\includegraphics[width=0.99\linewidth]{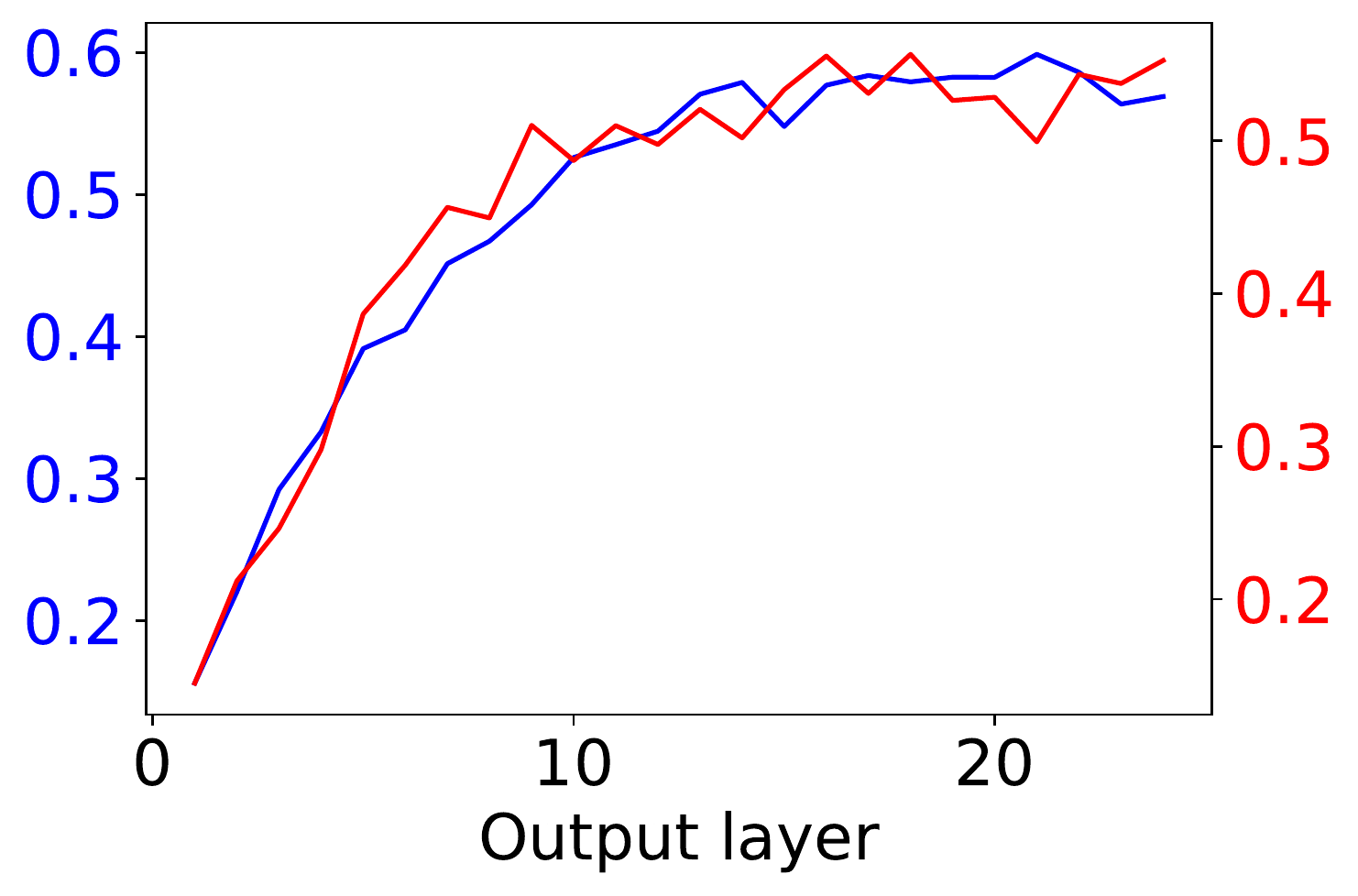}
			\caption{CoLA}
		\end{subfigure}
		~
		\begin{subfigure}[t]{0.48\linewidth}
			\includegraphics[width=0.99\linewidth]{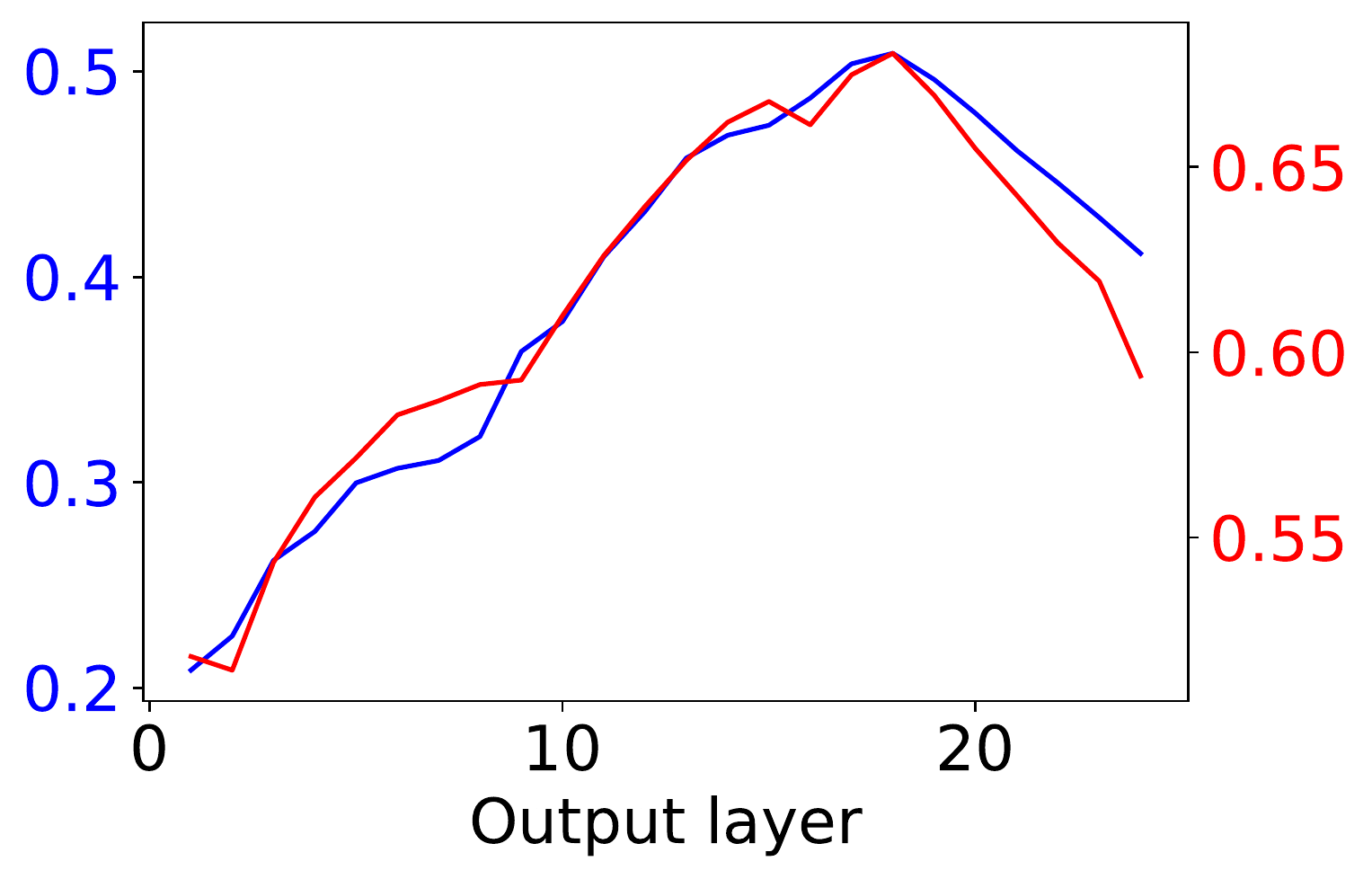}
			\caption{MNLI}
		\end{subfigure}
		
		\begin{subfigure}[t]{0.48\linewidth}
			\includegraphics[width=0.99\linewidth]{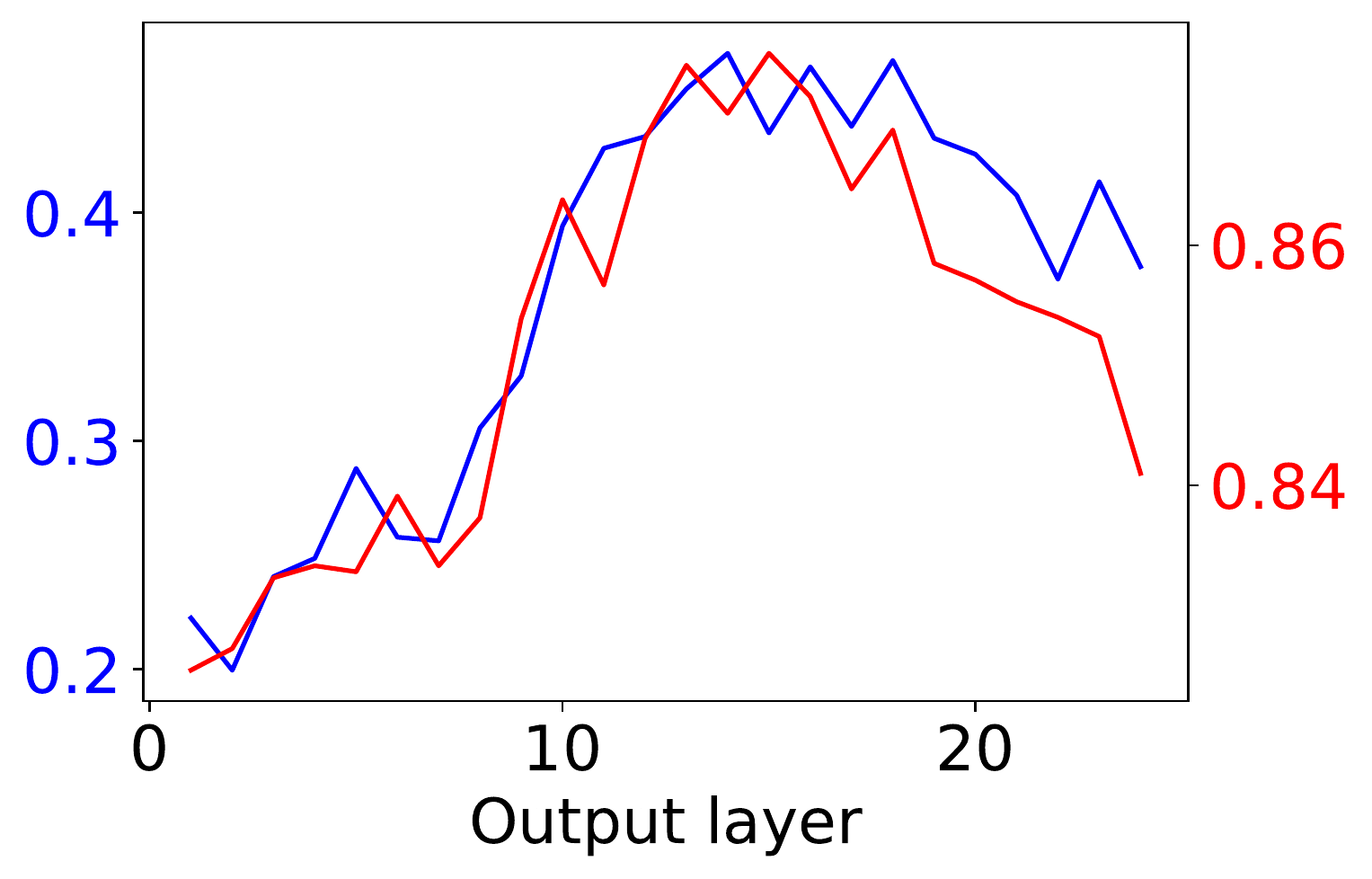}
			\caption{MRPC}
		\end{subfigure}
		~
		\begin{subfigure}[t]{0.48\linewidth}
			\includegraphics[width=0.99\linewidth]{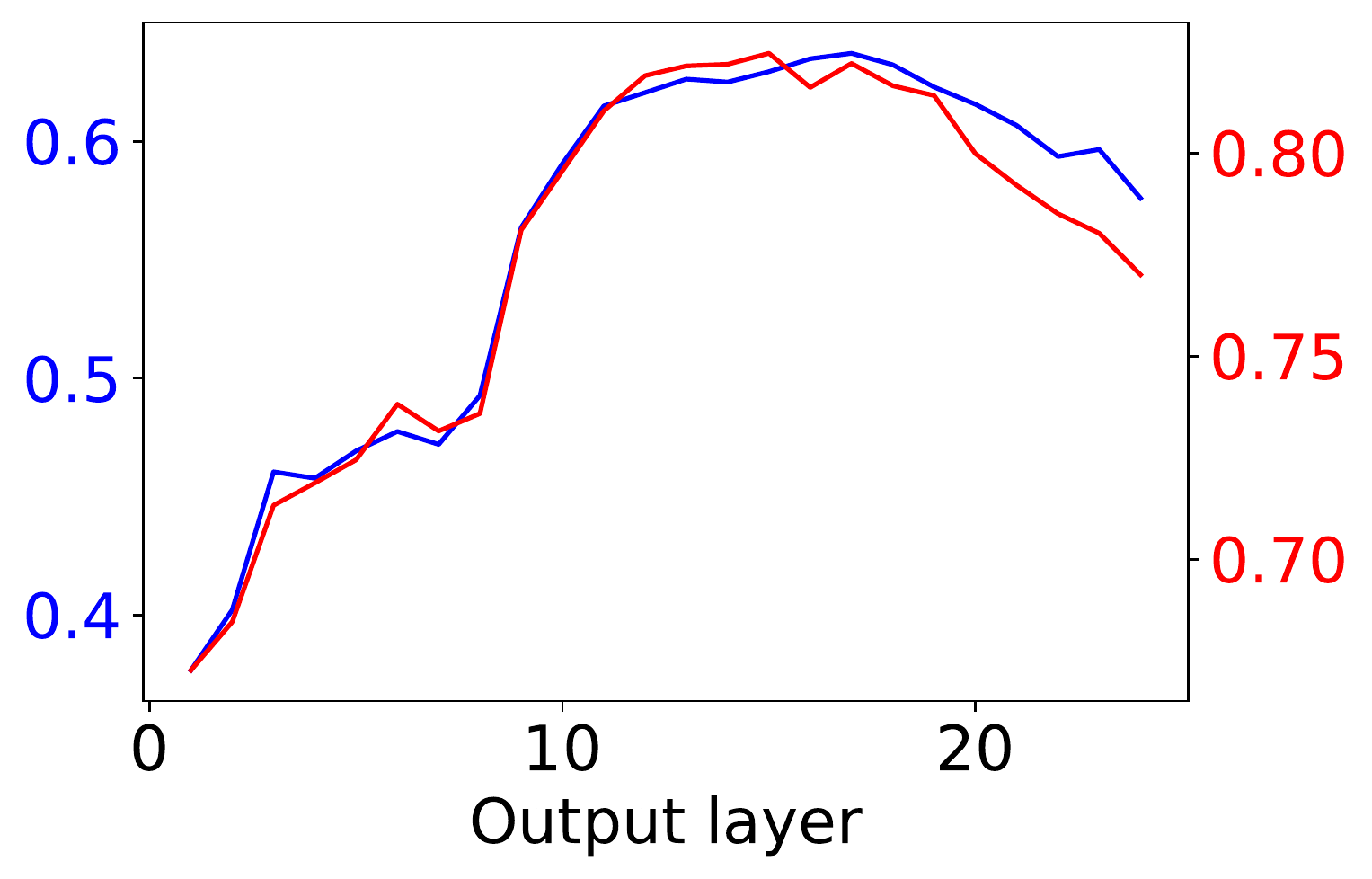}
			\caption{QNLI}
		\end{subfigure}
		
		\begin{subfigure}[t]{0.48\linewidth}
			\includegraphics[width=0.99\linewidth]{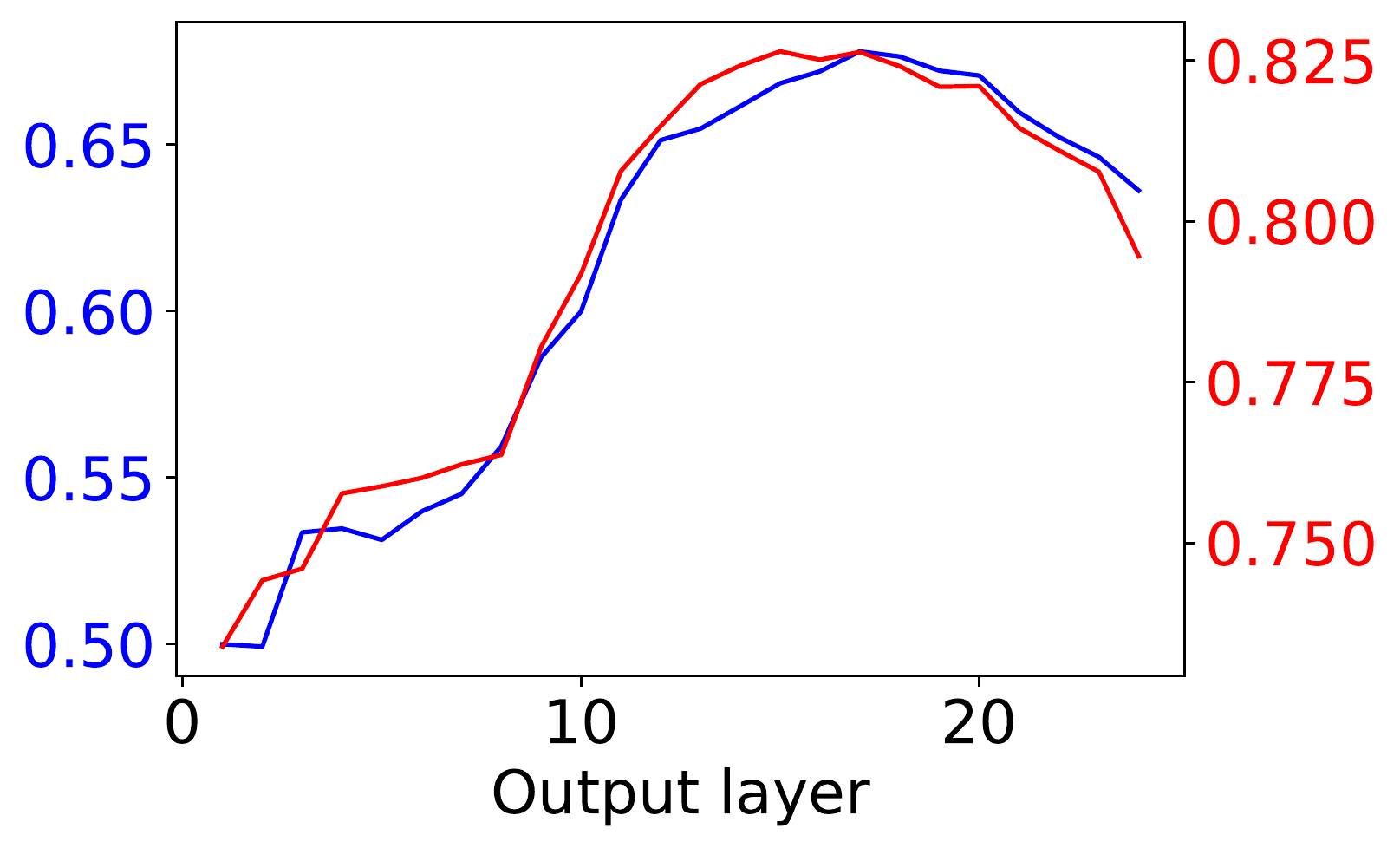}
			\caption{QQP}
		\end{subfigure}
		~
		\begin{subfigure}[t]{0.48\linewidth}
			\includegraphics[width=0.99\linewidth]{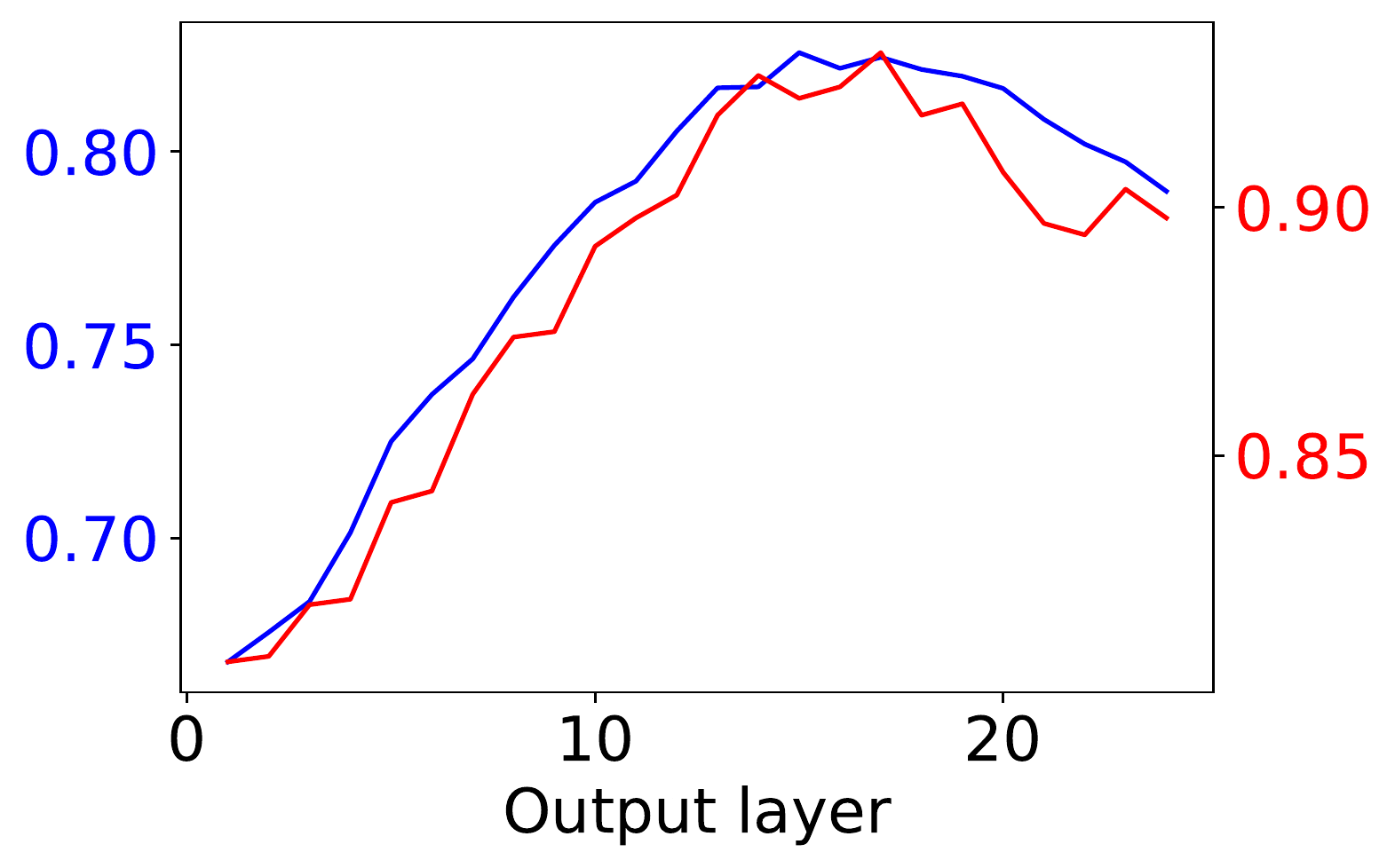}
			\caption{SST-2}
		\end{subfigure}
		\caption{The CCA score (blue) and probing performance (red) of each layer of a pretrained RoBERTa model. Each figure is a GLUE task.}\label{fig:cca_pretrain}
	\end{center}
\end{figure}

The CCA score are plotted in blue curves and the probing performance of the pretrained model are plotted in red curves in \cref{fig:cca_pretrain}. 
They almost overlap under different y-axes.

In order to ensure computational stability, the sampled auto-covariance matrices of $\vec{h}^{(\ell)}$ and $\vec{y}$ are perturbed by small constants,  $\epsilon_{\vec{h}}$ and $\epsilon_{\vec{y}}$, along the diagonal~\citep{de2003regularization}.
For each GLUE task we sample $N$ class-balanced data-points from the train set. In order to choose the regularization parameters and to avoid overfitting, we perform 10-fold cross validation by using eight of the ten splits to learn the linear projection matrices for different values of $\epsilon_\vec{h}$ and $\epsilon_\vec{y}$. We use one of the two remaining splits as development set and the other one as test set. The correlation for the development set is evaluated using the learned projection matrices and the best performing pair is then used to evaluate the test set score. We repeat this procedure thrice for each of the two samples of $N$ data points. We experiment with different values of $N$. Some of the previous work, although using CCA for representation learning, follows the same scoring procedure to choose the regularization parameters and the corresponding projection matrices~\citep{wang2015deep}.

CCA has been previously used to measure similarity of layer-wise representations within and across neural network models~\citep{raghu2017svcca}, and to measure similarity of the layer-wise word-level representations with off-the-shelf embedding maps~\citep{pasad2021layer}. \citet{williams2019quantifying} use CCA to correlate grammatical gender and lexical semantics by representing the discrete gender class as a one-hot vector.

\paragraph{Numerical Rank.}\label{app:rank}
As discussed in \cref{sec:rank}, the rank-based metric is $\varrho^{(\ell)} \defeq \frac{1}{|\mathcal{Y}|} \sum_{y\in\mathcal{Y}} \varrho_y^{(\ell)}$ where $\varrho_y^{(\ell)} \defeq\frac{\|\vec{H}_y^{(\ell)} \|_{*}^2}{\| \vec{H}_y^{(\ell)}\|_{F}^2}$. 
Each $\varrho_y^{(\ell)}$ can be viewed as measuring the sparsity of the singular values $\{ \sigma_i\}$ of $\vec{H}_y^{(\ell)}$ because $\varrho_y^{(\ell)} = \frac{\| \vec{\sigma}\|_1^2}{\| \vec{\sigma}\|_2^2}$.
It is an approximation of $\| \vec{\sigma}\|_0$, i.e., the rank of matrix $\vec{H}_y^{(\ell)}$.

\paragraph{Mutual information.}\label{app:mi}
Discrete mutual information (MI) gives a measure of mutual dependence between two discrete random variables. We use MI dependence between the sentence representations and the corresponding GLUE task labels as a measure of layer-wise task specificity. In order to discretize continuous-valued sentence representations, we run k-means clustering to obtain discrete clusters, as in \citet{voita2019bottom}. This measure has been previously used to measure the phone and word content in layer-wise representations of a pre-trained speech model~\citep{pasad2021layer}.

In our experiments, we sampled $N$ class-balanced data-points. We held a tenth of these samples out and ran the k-means clustering algorithm with $C$ clusters on the sampled data. Then the categorical ID of each held-out sample is defined to be the ID of the learned cluster that it was assigned to.
However, after extensive tuning, we still could not obtain any mutual information numbers that look reasonably high: actually, all the numbers were close to zero and they didn't differ much. 
We believe that the difficulty stems from the fact that learning clusters is \emph{unsupervised} and unsupervised learning is known to be difficult. 
Indeed, if we just use the class labels $y_n$ as the cluster IDs, we can observe a neat clustering---that is why our proposed metric $\nu$ is effective. 
However, it seems extremely difficult to learn that clustering in an unsupervised fashion. 

\end{document}